\documentclass{article}

\usepackage[preprint]{neurips_2026}

\usepackage[utf8]{inputenc} %
\usepackage[T1]{fontenc}    %
\usepackage[greek.polutoniko, english]{babel}
\usepackage{microtype}      %

\usepackage{amsmath, amssymb, mathtools, amsthm}
\usepackage{amsfonts}       %
\usepackage{nicefrac}       %
\usepackage{bbm}            %

\theoremstyle{plain}

\theoremstyle{definition}

\theoremstyle{remark}

\usepackage{graphicx}
\usepackage[export]{adjustbox} %
\usepackage{wrapfig}
\usepackage{subcaption}
\usepackage[font=small]{caption}

\usepackage{booktabs}       %
\usepackage{multirow}
\usepackage{array}          %
\usepackage{rotating}
\usepackage{multicol}

\usepackage{tcolorbox}
\usepackage{framed}
\usepackage[framemethod=TikZ]{mdframed}

\usepackage{xcolor}         %
\usepackage{paralist}
\usepackage[inline]{enumitem}
\usepackage{pifont}
\usepackage{fancyvrb}

\usepackage{tikz}
\usetikzlibrary{arrows.meta, positioning, fit, backgrounds, intersections, shapes.geometric}

\usepackage{algorithm}
\usepackage{algpseudocode}

\usepackage{listings}

\definecolor{codegreen}{rgb}{0,0.6,0}
\definecolor{codegray}{rgb}{0.5,0.5,0.5}
\definecolor{codepurple}{rgb}{0.58,0,0.82}
\definecolor{backcolour}{rgb}{0.95,0.95,0.92}

\lstdefinestyle{gridstyle}{
    basicstyle=\tiny\ttfamily,
    breaklines=true,           
    frame=single,              
    columns=fullflexible,
    keepspaces=true,
    aboveskip=0pt,
    belowskip=0pt
}

\lstdefinestyle{mystyle}{
    backgroundcolor=\color{backcolour},   
    commentstyle=\color{codegreen},
    keywordstyle=\color{magenta},
    numberstyle=\tiny\color{codegray},
    stringstyle=\color{codepurple},
    basicstyle=\ttfamily\footnotesize,
    breakatwhitespace=false,         
    breaklines=true,                 
    captionpos=b,                    
    keepspaces=true,                 
    numbers=left,                    
    numbersep=5pt,                  
    showspaces=false,                
    showstringspaces=false,
    showtabs=false,                  
    tabsize=2
}

\usepackage{url}
\usepackage{hyperref}

\hypersetup{
    colorlinks = true,
    citecolor=dartmouthgreen, %
    linkcolor=dartmouthgreen,
    urlcolor=dartmouthgreen
}

\usepackage[capitalize,noabbrev]{cleveref}

\usepackage{xcolor}
\usepackage{colortbl}
\usepackage{textgreek}
\usepackage{calc} %

\definecolor{darkgreen}{rgb}{0.0, 0.5, 0.0}
\definecolor{blue}{rgb}{0.0, 0.47, 0.75}
\definecolor{dartmouthgreen}{rgb}{0.05, 0.5, 0.06}
\definecolor{drab}{rgb}{0.59, 0.44, 0.09}
\definecolor{navyblue}{rgb}{0.0, 0.0, 0.5}

\definecolor{Gray}{gray}{0.5}
\definecolor{LightCyan}{rgb}{0.88,1,1}

\definecolor{eclipseblue}{rgb}{0.16,0.0,1.0}
\definecolor{eclipsegreen}{rgb}{0.25,0.5,0.35}
\definecolor{eclipsepurple}{rgb}{0.5,0.0,0.33}
\definecolor{backcolour}{rgb}{1.0, 1.0, 1.0} %

\DeclareMathOperator*{\argmin}{arg\,min}

\newcommand{\vc}{\mathbf{c}}

\newcommand{\vh}{\mathbf{h}}

\newcommand{\vo}{\mathbf{o}}

\newcommand{\vx}{\mathbf{x}}

\newcommand{\vz}{\mathbf{z}}

\lstdefinestyle{pythonstyle}{
    language=Python,
    backgroundcolor=\color{backcolour},
    commentstyle=\color{dartmouthgreen},
    keywordstyle=\color{eclipsepurple}\bfseries,
    numberstyle=\tiny\color{gray},
    stringstyle=\color{eclipseblue},
    basicstyle=\ttfamily\footnotesize,
    breakatwhitespace=false,
    breaklines=true,
    captionpos=b,
    keepspaces=true,
    numbers=left,
    numbersep=5pt,
    showspaces=false,
    showstringspaces=false,
    showtabs=false,
    tabsize=4,
    frame=lines,                       %
    rulecolor=\color{black}
}

\title{Intrinsic Selection and Particle Resampling for Inference-Time Scaling Beyond Domain Verifiability}

\author{%
  Giorgio Giannone~\footnote{Correspondence to:~\texttt{ggiorgio@mit.edu}} \\
  AI Innovation, Red Hat \\
  \And
  Mustafa Eyceoz \\
  AI Innovation, Red Hat \\
  \And
  Shabana Baig \\
  AI Innovation, Red Hat \\
  \And
  Shivchander Sudalairaj \\
  AI Innovation, Red Hat \\
  \And
  Anna Clare Doris \\
  DeCoDE Lab, MIT \\
  \AND
  Faez Ahmed \\
  DeCoDE Lab, MIT \\
  \And
  Akash Srivastava \\
  Core AI, IBM \\
  \And
  Kai Xu \\
  AI Innovation, Red Hat \\
}

\begin{document}

\maketitle

\begin{abstract}
Inference-Time Scaling (ITS) has largely succeeded in verifiable domains like math and coding, where cheap verification enables scalable output selection. However, extending ITS to tasks prone to systematic failure - driven by faulty initial assumptions or unmet multidimensional constraints - typically relies on costly external solvers or brittle, model-based verifiers. Our key insight is that the intrinsic statistics of parallel sample sets, specifically length-adjusted tail entropy, provide a robust discriminative signal for solution quality without access to ground truth. Crucially, these statistics serve as a difficulty gate for adaptive compute allocation, dynamically routing problems across scaling regimes. First, \emph{Intrinsic Selection} (\texttt{iS}) ranks candidates post-hoc, matching consensus-based algorithms across three domains and improving engineering design selection by 20\% over pass@1 baselines. Second, \emph{Intrinsic Particle Filtering} (\texttt{iPF}) generalizes this to step-level resampling, guiding generation toward high-confidence reasoning trajectories to improve pass@1 by 6.1 points on average on hard math problems. Finally, \emph{Particle Distillation} (\texttt{dPF}) injects privileged guidance via early logit blending and KL-guided resampling, steering generation past systematic reasoning errors to satisfy expert rubrics, yielding up to 26.5\% gains on complex clinical responses. Our pipeline applies seamlessly across broad-purpose, domain-specialized, and multimodal architectures, successfully extending ITS to open-ended domains without requiring trained reward models or exact ground-truth verification.
\end{abstract}

\section{Introduction}
\label{sec:intro}

Recent advances in reasoning and agentic foundation models~\citep{guo2025deepseek,jaech2024openai,team2025kimi} have produced dramatic performance gains. These successes are most prominent in verifiable domains like mathematics and coding, where computationally inexpensive, exact verification, such as symbolic parsing, unit testing, and output matching, enables reinforcement-based tuning at scale~\citep{lambert2024tulu,wen2025reinforcement}. In these structured settings, building reward models and test-driven verification pipelines is highly feasible, creating a clear path to building high-performing workflows through increased inference-time compute~\citep{snell2024scaling,zhang2025agentic}.

Leveraging this compute, Inference-Time Scaling (ITS) techniques combine parallel and sequential sampling to rank and select promising candidate solutions~\citep{brown2024large,wu2025inference,wang2022self,snell2024scaling}. While ITS has proven remarkably effective when proxy verification is readily available, extending it beyond these domains exposes a fundamental challenge. Existing methods rely heavily on extrinsic metrics, such as reward models, output standardization, or step-level scoring, to rank, select, and steer generation. For domains where external verification is expensive or unavailable at scale, progress has been severely limited~\citep{tang2025beyond}.

\begin{figure}[t]
    \centering
    \begin{subfigure}[t]{0.23\textwidth}
        \centering
        \includegraphics[width=\textwidth]{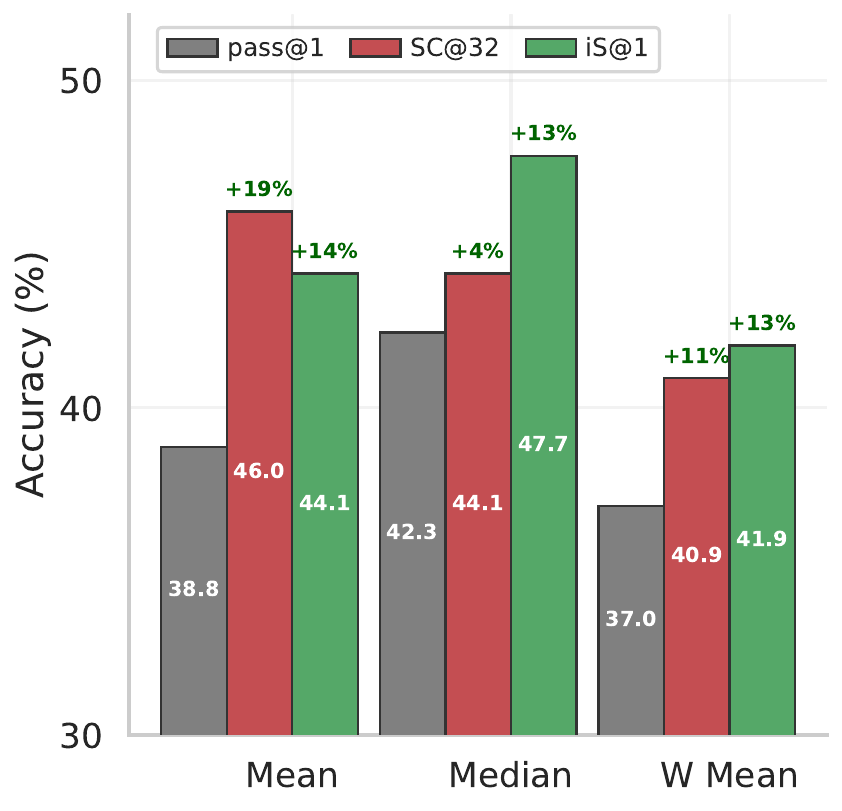}
        \caption{iS vs SC across math, reasoning, and coding domains over seven datasets.}
        \label{fig:performance-phase1}
    \end{subfigure}
    \hfill
    \begin{subfigure}[t]{0.23\textwidth}
        \centering
        \includegraphics[width=\textwidth]{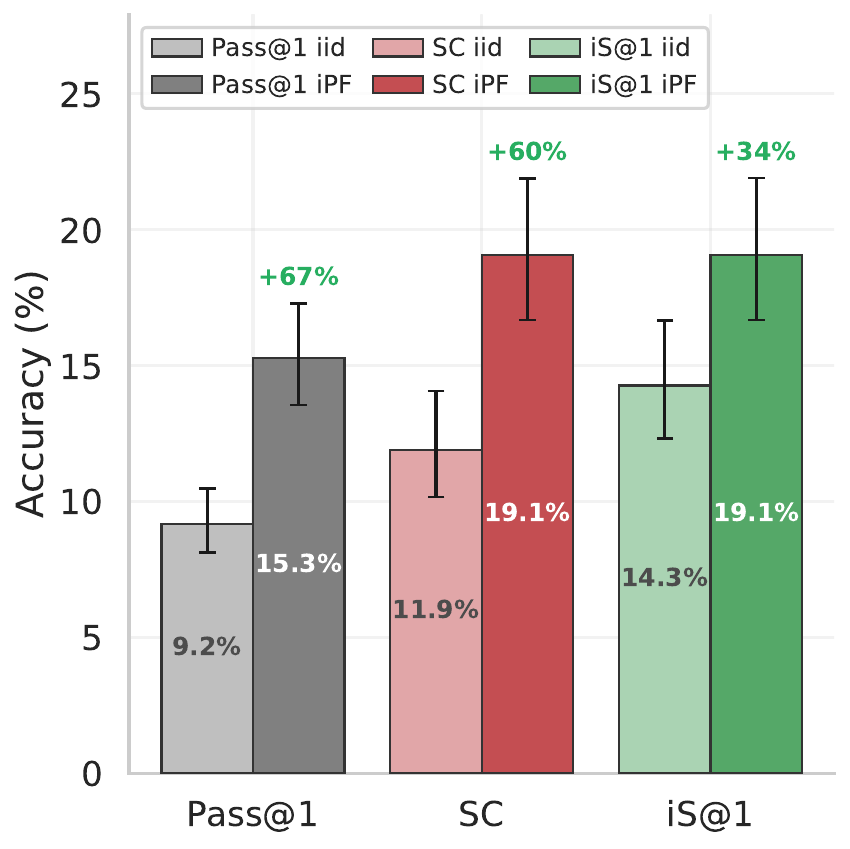}
        \caption{iPF vs iid sampling on hard to solve AIME problems.}
        \label{fig:aime-hard}
    \end{subfigure}
    \hfill
    \begin{subfigure}[t]{0.23\textwidth}
        \centering
        \includegraphics[width=\textwidth]{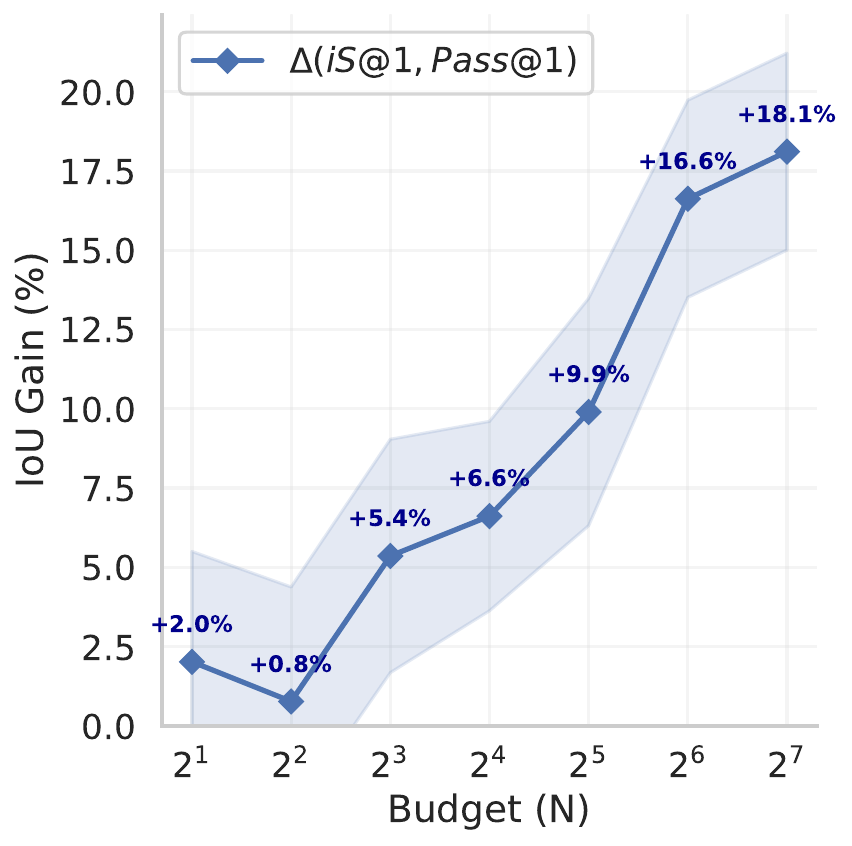}
        \caption{OOD Image-to-CAD selection on Fusion360 (IoU).}
        \label{fig:cad-method}
    \end{subfigure}
    \hfill
    \begin{subfigure}[t]{0.23\textwidth}
        \centering
        \includegraphics[width=\textwidth]{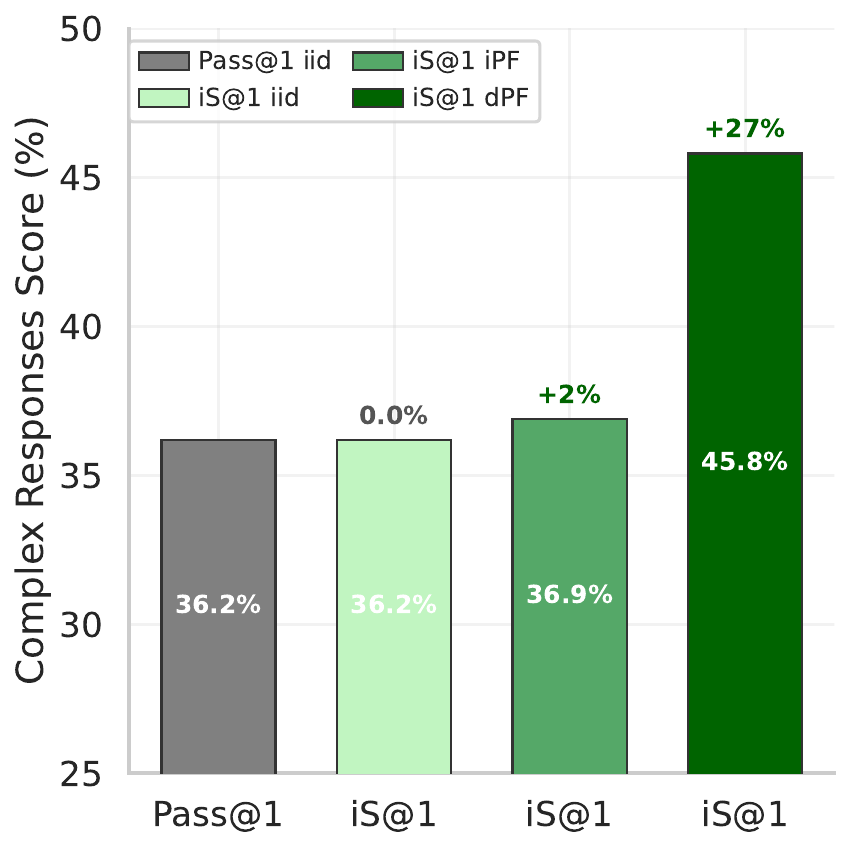}
        \caption{dPF on HealthBench-Hard complex responses.}
        \label{fig:health-method}
    \end{subfigure}
    \caption{Intrinsic inference scaling across domains. \textbf{(a)}~iS matches self-consistency without output verification. \textbf{(b)}~iPF improves pass@1 on hard AIME problems. \textbf{(c)}~iS enables inference scaling for CAD generation without reward models. \textbf{(d)}~dPF improves complex clinical responses via rubric-guided distillation.}
    \label{fig:intro-scaling-improvement}
\end{figure}

\paragraph{The Bottleneck of Extrinsic Verification} Relying on extrinsic metrics creates a severe bottleneck: consensus methods fail in open-ended settings~\citep{wang2022self,chen2023universal}, reward models are expensive and rarely available~\citep{zhang2025lessons,stiennon2020learning}, and heuristic parsers or solvers are often fragile and slow (Appendix~\ref{appx:verifiability}, Figure~\ref{fig:domain_verifiability-appx}). Thus, we ask: \emph{can we guide inference-time scaling using only the intrinsic statistics of generated sample sets, sidestepping external verifiers entirely?}

Our key observation is that parallel sample sets inherently carry strong discriminative signals - specifically, per-token tail entropy and overall length distributions - that encode solution quality and problem difficulty without ground-truth access. By analyzing the proportion of correct solutions $n$ within $N$ samples, we structure our approach around three challenge regimes: identifying the best candidate in \emph{solved-easy} settings ($n/N$ is sizable), concentrating compute on sparse correct trajectories in \emph{solved-hard} tasks ($n/N \ll 1$), and correcting faulty early assumptions to navigate spaces of \emph{systematic failure} ($n/N \approx 0$).

\paragraph{Contributions} 
We propose a framework for inference-time scaling that \emph{bypasses traditional verifiers and process reward models}. It comprises three complementary methods, each addressing a different challenge regime:

\begin{itemize}%
  \item \emph{Intrinsic Selection} (\texttt{iS}):
    ranks candidates using set-level length distributions and adjusted
    per-token tail entropy, requiring no parsers, canonical forms, or
    reward models.

  \item \emph{Intrinsic Particle Filtering} (\texttt{iPF}):
    performs step-level resampling guided solely by intrinsic entropy, steering
    generation toward high-confidence reasoning trajectories without
    process reward models.

  \item \emph{Particle Distillation} (\texttt{dPF}):
    injects privileged information, such as rubric criteria or hints, via logit blending and KL-guided resampling, correcting faulty early assumptions and enforcing constraints without explicit verification.
\end{itemize}

These three methods form a highly complementary pipeline that operates entirely without trained reward models or ground-truth verification. \emph{While iS and iPF are strictly verifier-free - relying solely on internal token distributions - dPF introduces privileged guidance (e.g., rubrics or hints) to steer generation on highly constrained problems where intrinsic signals alone are insufficient.} Crucially, their success across diverse architectures suggests intrinsic entropy is a fundamental property rather than a model-specific artifact.

Evaluated across five domains (math, science, coding, CAD, and clinical reasoning), the framework integrates seamlessly into modern inference engines without complex calibration or post-processing to enable intrinsic inference scaling (Figure~\ref{fig:intro-scaling-improvement}). Notably, the strictly verifier-free iS matched consensus algorithms without output verification across multiple datasets (Figure~\ref{fig:performance-phase1}), and improved pass@1 by 20\% in hard-to-verify CAD tasks (Figure~\ref{fig:cad-method}). On the hardest math problems, iPF increased pass@1 by 6.1 points on average (Figure~\ref{fig:aime-hard}). Finally, utilizing external guidance without reward models, dPF achieved up to 26.5\% gains in clinical responses using just 8 particles instead of 32 independent samples (Figure~\ref{fig:health-method}).

\section{Background and Related Work}
\label{sec:background}

\paragraph{Domain Verifiability}
Appendix~\ref{appx:verifiability} Table~\ref{tab:verifiability-appx} contrasts verifiable domains (e.g., math, coding), which offer cheap automated rewards (\(V(x,y) \in \{0,1\}\)), with hard-to-verify domains (e.g., healthcare, large-scale engineering) that require expensive subjective or model-based evaluation. These high costs bottleneck scalability. Furthermore, even in structured fields, heuristic parsers and solvers remain fragile and computationally expensive, effectively rendering them hard-to-verify at scale~(Appendix~\ref{appx:verifiability}).

\paragraph{Inference-Time Scaling}
Allocating additional compute during generation significantly improves reasoning~\citep{snell2024scaling,brown2024large,wu2025inference}. Scaling strategies generally divide into \emph{parallel sampling} (generating and selecting from independent trajectories) and \emph{sequential search} (navigating reasoning trees toward high-density solution regions).

\paragraph{Consensus-based Selection}
Self-Consistency (SC)~\citep{wang2022self} is a highly successful verification-free parallel strategy. It approximates the conditional answer distribution by marginalizing latent trajectories \(\{\vz\}^K_{k=1} \sim p(\vz \mid \vc)\):
\[%
p(\vx \mid \vc) = \sum_{\vz} p(\vx \mid \vz, \vc) p(\vz \mid \vc).
\]
These systems extract consensus via majority voting over \(M\) choices \(\{a_m\}^{M}_{m=1}\), as defined in Eq.~\ref{eq:sc-vote}:
\begin{equation}
x_a^* = \arg\max_{a} \sum^M_{m=1} \mathbb{I}(a_m = a).
\label{eq:sc-vote}
\end{equation}
While scalable and robust to noise, SC relies on cheap syntactic verification, severely limiting its use in open-ended or multi-part tasks. Universal SC~\citep{chen2023universal} addresses this via LLM-based judging, but incurs new model-based verification costs.

\paragraph{Confidence-based Selection}
To bypass external verifiers, confidence-based methods leverage internal model signals. This includes intrinsic verification in RL-trained models~\citep{guo2025deepseek,jaech2024openai}, Self-Certainty~\citep{kang2025scalable}, trace-level logit weighting~\citep{fu2025deep}, and uncertainty-aware generation~\citep{zhang2025tokur}. For a probabilistic model $p$ defined over vocabulary \(V\) and sequence length \(T\) we define: \(p(\vz) = p(\vz_T) \prod^{T-1}_{t=1} p(\vz_t \mid \vz_{<t})\), and quantify model uncertainty using token entropy~\citep{shannon1948mathematical} in Eq.~\ref{eq:token-entropy}:
\begin{equation}
    \mathbb{H}[\vz_t \mid \vz_{<t}] = - \sum_{v \in \mathcal{V}} p(\vz_t = v \mid \vz_{<t}) \log p(\vz_t = v \mid \vz_{<t}).
\label{eq:token-entropy}
\end{equation}
Eq.~\ref{eq:kl-div} demonstrates that this is equivalent to the negated KL divergence relative to a uniform distribution \(U\):
\begin{equation}
\mathbb{KL}[p, U]_t = \sum_{v \in \mathcal{V}} p(\vz^v_t) \log \dfrac{p(\vz^v_t)}{1/V} = - \mathbb{H}\left[p\right]_t + \log(V).
\label{eq:kl-div}
\end{equation}
Sequences with lower entropy (or higher KL divergence) exhibit relative confidence. Our approach complements RL-trained models~\citep{guo2025deepseek,jaech2024openai}; rather than relying solely on step-level verification within individual trajectories, we extract discriminative signals from the collective statistics of an \(N\)-sample set.

\paragraph{Particle-based Inference and Guidance}
Sequential Monte Carlo (SMC) methods~\citep{doucet2001introduction,zhao2024probabilistic,lew2023sequential} and Particle Filtering (PF)~\citep{puri2025probabilistic} approximate the posterior \(p(\vz \mid \vc, \vo) \propto p(\vo \mid \vz, \vc) p(\vz \mid \vc)\) via sequential importance resampling~\citep{liu2001theoretical} with a proposal \(q(\vz \mid \vc)\), leading to the standard weight update shown in Eq.~\ref{eq:smc-weight}:
\begin{equation}
w_{t} \propto w_{t-1}~\dfrac{p(\vz_{t} \mid \vz_{t-1}, \vc, \vo_{t})}{q(\vz_{t} \mid \vz_{t-1}, \vc)}.
\label{eq:smc-weight}
\end{equation}
Using the model $p(\vz \mid \vc)$ itself as a bootstrap filter proposal~\citep{elfring2021particle} reduces this weight update to \(w_t \propto w_{t-1}\,p(\vo_t \mid \vz_t, \vc)\). Typically, the observation model relies on a Process Reward Model (PRM), i.e., \(r_{\phi}(\vz_t, \vc) = p(\vo_t = 1 \mid \vz_t, \vc)\). However, PRMs are expensive, domain-specific, and prone to overconfidence on out-of-distribution long sequences~\citep{park2025know}, limiting their viability in complex or specialized fields.

See Appendix~\ref{appx:related-work} and~\ref{appx:verifiability} for extended related work.

\section{Method}
\label{sec:method}
\begin{figure}[t]
    \centering
    \includegraphics[width=.9\linewidth]{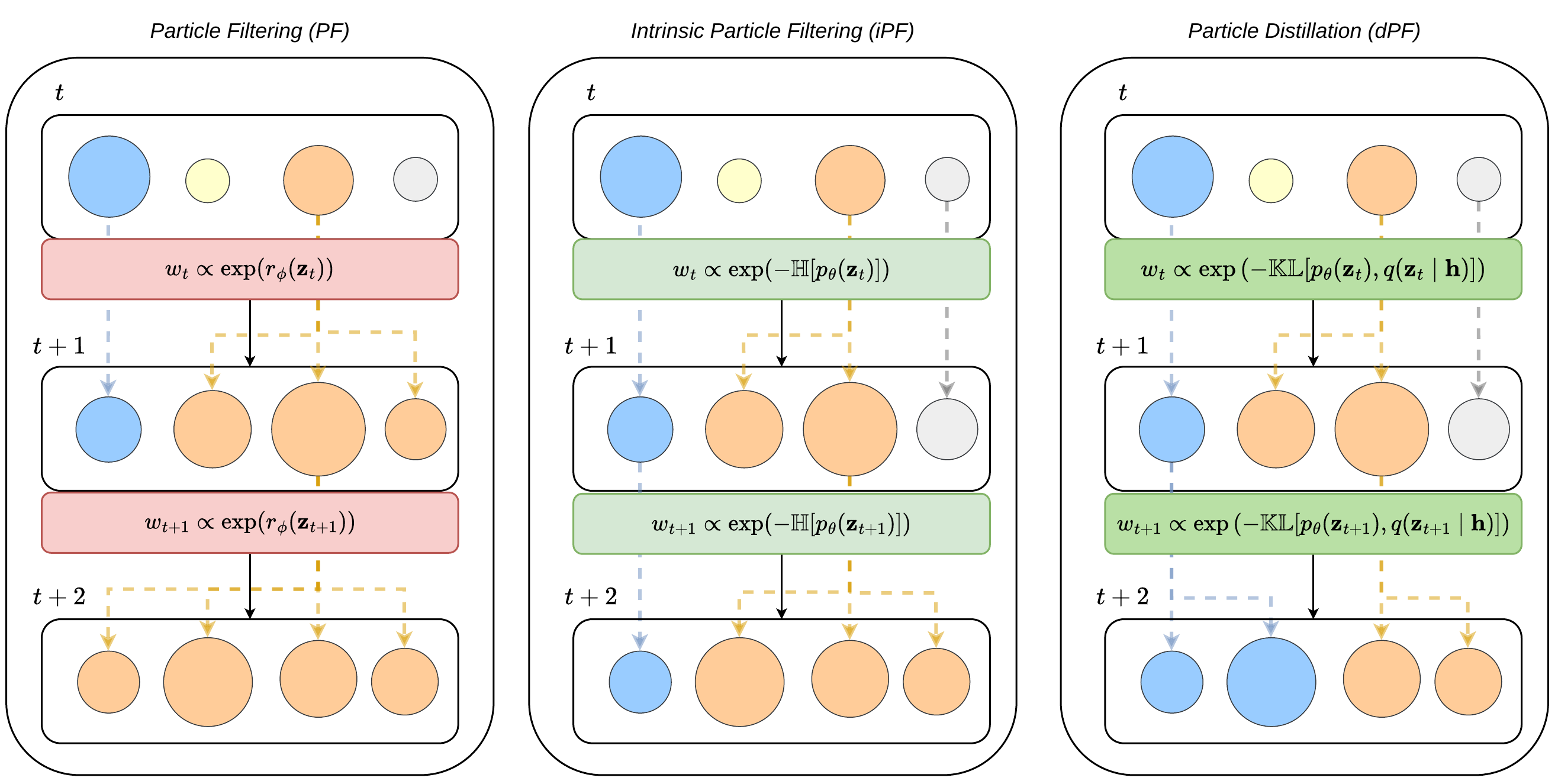}
    \caption{Overview of intrinsic particle filtering (iPF) and particle distillation (dPF). Both extend particle filtering to hard-to-verify domains without requiring calibrated external reward models. iPF resamples particles using intrinsic token-level entropy to concentrate compute on high-confidence trajectories, while dPF uses privileged guidance (e.g., rubrics or hints) through logit blending and KL-guided resampling to correct early errors and enforce task-specific constraints.}
    \label{fig:main-intro}
\end{figure}

Our goal is to enable effective inference-time scaling across a broad range of domains, regardless of whether external verification is available. To do so, we exploit intrinsic statistics of generated samples to derive signals for candidate selection, problem difficulty estimation, entropy-guided resampling, and guided steering, as illustrated in our framework overview (Figure~\ref{fig:main-intro}). See Appendix~\ref{app:derivation} for more details.

\subsection{Intrinsic Selection (iS)}

When the model can generate correct answers among $N$ parallel samples, the main challenge is to identify the best candidate without external verification. Given a problem $\mathbf{c}$, a model $p_{\theta}(\mathbf{z} \mid \mathbf{c})$, and a set of $N$ sampled trajectories $\{\mathbf{z}_n\}_{n=1}^N$, Intrinsic Selection (\texttt{iS}) ranks candidates using adaptive token-level intrinsic statistics:
\begin{equation}
  \texttt{iS}(\mathbf{c}) = g\!\left(\{\mathbf{z}_n\}_{n=1}^{N},\, l(\cdot),\, \mathbb{H}[\cdot]\right).
  \label{eq:is-formalist}
\end{equation}
The scoring function $g$ in Eq.~\ref{eq:is-formalist} aggregates per-token entropy over an adaptively chosen tail region determined by sequence length $l(\cdot)$ and token entropy $\mathbb{H}[\cdot]$.

\paragraph{Adaptive Tail Entropy}
We define an adaptive tail window that balances two competing effects: longer windows capture more of the answer region, while shorter windows reduce noise from the preceding reasoning trace. Using the set-level length distribution $l(\mathbf{z})$ and entropy distribution $\mathbb{H}[\mathbf{z}_n]$, we define the tail cutoff as
\begin{equation}
t_c = \frac{\sqrt{N^{-1}\sum_{n=1}^N l(\mathbf{z}_n)}}{N^{-1}\sum_{n=1}^N \mathbb{H}[\mathbf{z}_n]}.
\label{eq:tail}
\end{equation}
The sublinear scaling $t_c \propto \sqrt{l}$ in Eq.~\ref{eq:tail} reflects the empirical observation that answer spans grow sub-linearly with total sequence length. The inverse dependence on mean entropy ensures that, when the model is uncertain across the sample set, a narrower tail is used to avoid uninformative entropy spikes. Letting the tail index set be $\mathcal{A} = \{t \in \mathbb{Z} \mid t_c \le t \le T\}$, we define the adjusted tail entropy score for sample $n$ as:
\begin{equation}
s_n(\mathbf{c}) = \underset{t \in \mathcal{A}}{\mathrm{agg}}\!\left(\mathbb{H}[\mathbf{z}_t^{(n)} \mid \mathbf{z}_{<t}^{(n)}]\right).
\label{eq:tail-entropy}
\end{equation}
In Eq.~\ref{eq:tail-entropy} we aggregate over tokens using the median, which is robust to outlier spikes caused by punctuation or formatting, and select the sample with the smallest $s_n$. In the following, we refer to this quantity interchangeably as \texttt{iS}@1.

\begin{figure}[t]
    \centering
    \includegraphics[width=.9\linewidth]{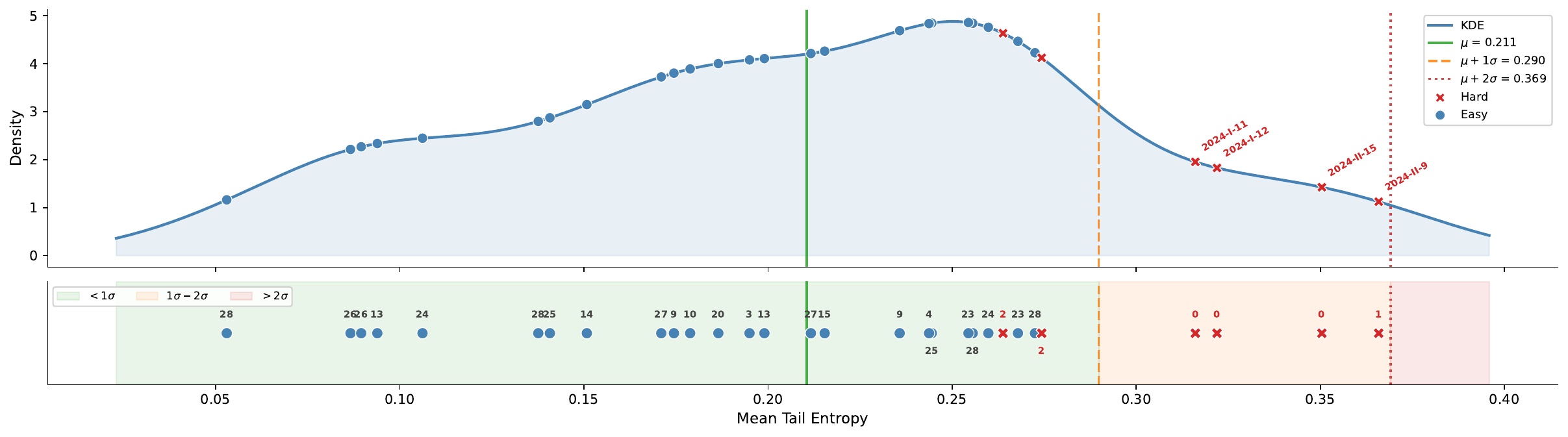}
    \caption{Problem-level entropy and estimated difficulty. Higher adjusted mean per-token tail entropy correlates with harder problems. Instances more than $1\sigma$ above the mean are strongly associated with hardness, defined by $n/N$ with $n \in \{0,1,2\}$.}
    \label{fig:entropy-difficulty}
\end{figure}

\paragraph{Difficulty Estimation and Routing}
Problem difficulty is inherently compute and model-dependent; what is challenging for a base model may be trivial for an advanced one. Therefore, we propose a data-driven, verifier-free approach to estimate relative problem difficulty using small-scale inference statistics (Appendix~\ref{appx:difficulty} Table~\ref{tab:tail_entropy_difficulty_results}). By generating a standard budget of $N$ parallel samples from the base policy $p_{\theta}$, we analyze set-level intrinsic metrics to approximate difficulty. Specifically, after applying length-based trimming to filter outliers, we compute the average token-level tail entropy.

These scores reflect the model's task-specific uncertainty, allowing us to accurately isolate the highest-entropy subset (Figure~\ref{fig:entropy-difficulty}). 
This mechanism naturally serves as a difficulty routing for downstream compute allocation: problems exhibiting low tail entropy are resolved instantly via \texttt{iS}, while tasks triggering the high-entropy gate are dynamically routed to \texttt{iPF} or \texttt{dPF} for step-level intervention or guidance.

\subsection{Intrinsic Particle Filtering (iPF)}

When a correct trajectory exists but lies in a low-density region ($n/N \ll 1$), parallel sampling alone is inefficient. We therefore introduce Intrinsic Particle Filtering (\texttt{iPF}), which concentrates generation on sparse, high-confidence trajectories using intrinsic entropy signals rather than external reward models.

In standard sequential importance resampling for inference scaling, a reward model is used to approximate the target distribution~\citep{puri2025probabilistic,zhang2025lessons}. If we take the model itself as the proposal distribution, i.e.,
\[
q(\vz_t \mid \vz_{t-1}, \vc) = p(\vz_t \mid \vz_{t-1}, \vc),
\]
then by Bayes' rule the unnormalized particle weight update reduces to:
\begin{equation}    
w_t \propto w_{t-1} \, p(\vo_t \mid \vz_t, \vc).
\label{eq:ipf-obs}
\end{equation}

Rather than using an external reward to define the observation likelihood $p(\vo_t \mid \vz_t, \vc)$ in Eq.~\ref{eq:ipf-obs}, we posit that particles with low recent entropy correspond to more confident and promising reasoning directions. We therefore define a pseudo-observation likelihood~\citep{karampela2026variational,gloaguen2022pseudo} as an exponential function of intrinsic entropy (as defined in Eq.~\ref{eq:token-entropy}), which yields the \texttt{iPF} resampling update in Eq.~\ref{eq:ipf}:
\begin{equation}
w_t \propto w_{t-1} \exp(-\mathbb{H}[\vz_t]).
\label{eq:ipf}
\end{equation}

\paragraph{Building the Resampling Distribution}
When intrinsic metrics are used for resampling, the resulting unnormalized weights can vary widely in magnitude. This contrasts with standard PRM-based resampling, where scores are typically bounded in $[0,1]$ and log-weights can be obtained using transformations such as $\log(p/(1-p))$. Without such bounds, intrinsic signals can lead to degenerate resampling, in which a single particle dominates and the particle set collapses. To mitigate this issue, in Eq.~\ref{eq:ipf-norm} we apply a robust normalization step before computing the final weights:
\begin{equation}
\tilde{w}_t \leftarrow
\frac{\mathbb{H}[\mathbf{z}_t] - m(\mathbb{H}[\mathbf{z}_t])}
{s(\mathbb{H}[\mathbf{z}_t]) + \epsilon},
\label{eq:ipf-norm}
\end{equation}
where $\epsilon$ is a small constant for numerical stability, and $m(\cdot)$ and $s(\cdot)$ denote the batch mean and standard deviation, respectively. We then obtain normalized particle weights using the softmax function, $w_t = \sigma(\tilde{w}_t)$, where $\sigma(x) = \exp(x)/\sum_{x \in X} \exp(x)$. We use the same normalization strategy for both $\mathbb{H}$-based and $\mathbb{KL}$-based scoring.

\subsection{Particle Distillation (dPF)}

When a problem exhibits \emph{systematic failure} ($pass@N \approx 0$), simply increasing the sample budget $N$ is computationally wasteful. Because the base policy is consistently misled by incorrect early assumptions, standard inference-time scaling yields diminishing returns. To overcome this, we introduce \emph{Particle Distillation} (\texttt{dPF}), an extension of our resampling framework that operates without trained reward models. \texttt{dPF} actively corrects these faulty assumptions using a guide conditioned in-context on privileged information~\citep{liao2026self}---such as hints, expert rubrics, or environmental constraints~\citep{zhao2026self,agarwal2024policy,hubotter2026reinforcement}. By combining early-trajectory logit blending with subsequent KL-guided resampling, \texttt{dPF} steers generation to satisfy complex multidimensional constraints before the model can commit to a flawed reasoning path.

\paragraph{Logits Blending and Guided Divergence}
When a model systematically fails due to incorrect early assumptions, naively scaling generation is futile. \texttt{dPF} mitigates this by mixing the base model's unconditional generation with a guide conditioned on privileged information $\mathbf{h}$ (e.g., rubrics, constraints, or hints) during the critical first steps:
\begin{equation}
\text{logits}(\vz^{h}_t) = \alpha~\text{logits}(\vz_t, \vc) + (1-\alpha)~\text{logits}(\vz_t, \vc, \mathbf{h}) \quad \alpha \in (0,1),
\label{eq:dpf-logits}
\end{equation}
where $\vz_t \sim p_{\theta}(\vz_t \mid \vz^{h}_{t-1}, \vc)$ in Eq.~\ref{eq:dpf-logits}. Inspired by classifier-free guidance in diffusion models~\citep{ho2022classifier}, we apply a linear annealing schedule to the blending parameter $\alpha$. This directly steers trajectories toward correct assumptions before the model commits to a flawed reasoning path. 

After annealing $\alpha$, we transition to steering the remaining steps via guided step-level resampling~\citep{doucet2001sequential,zhao2024probabilistic}. Instead of raw entropy, we dynamically compute particle weights based on the KL divergence between the base model and the guide:
\begin{equation}
w_t \propto w_{t-1} \exp\left(-\mathbb{KL}[p_{\theta}(\vz_t \mid \vz_{t-1}, \vc), q(\vz_t \mid \vz_{t-1} , \vc, \mathbf{h})]\right).
\label{eq:dpf}
\end{equation}
By penalizing divergence using Eq.~\ref{eq:dpf}, we effectively distill the guided trajectory back into the unconstrained model, navigating highly constrained problems leveraging guidance.

\section{Experiment}
\label{sec:experiment}

\begin{figure}[t]
    \centering
    \begin{subfigure}[b]{0.3\textwidth}
        \centering
        \includegraphics[width=\textwidth]{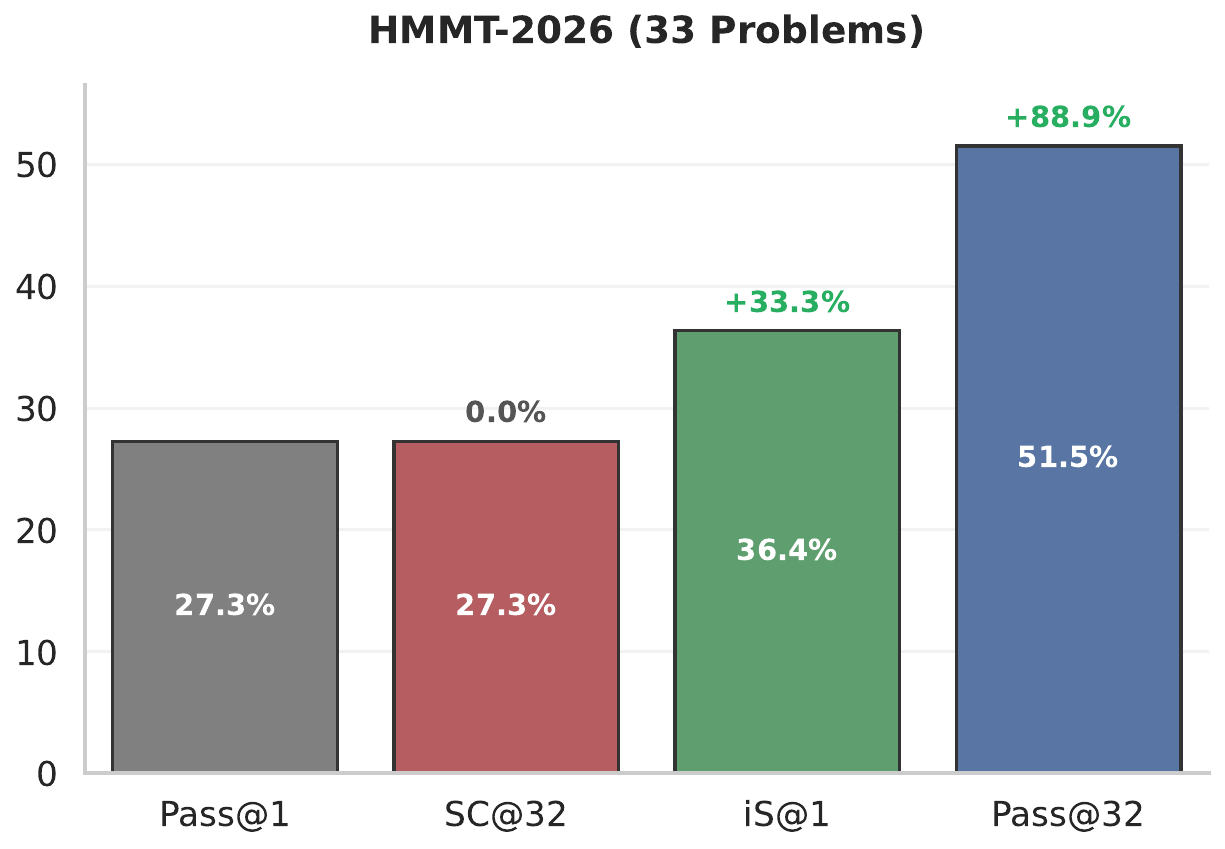}
        \caption{HMMT 2026}
        \label{fig:h26-intro}
    \end{subfigure}
    \hfill
    \begin{subfigure}[b]{0.3\textwidth}
        \centering
        \includegraphics[width=\textwidth]{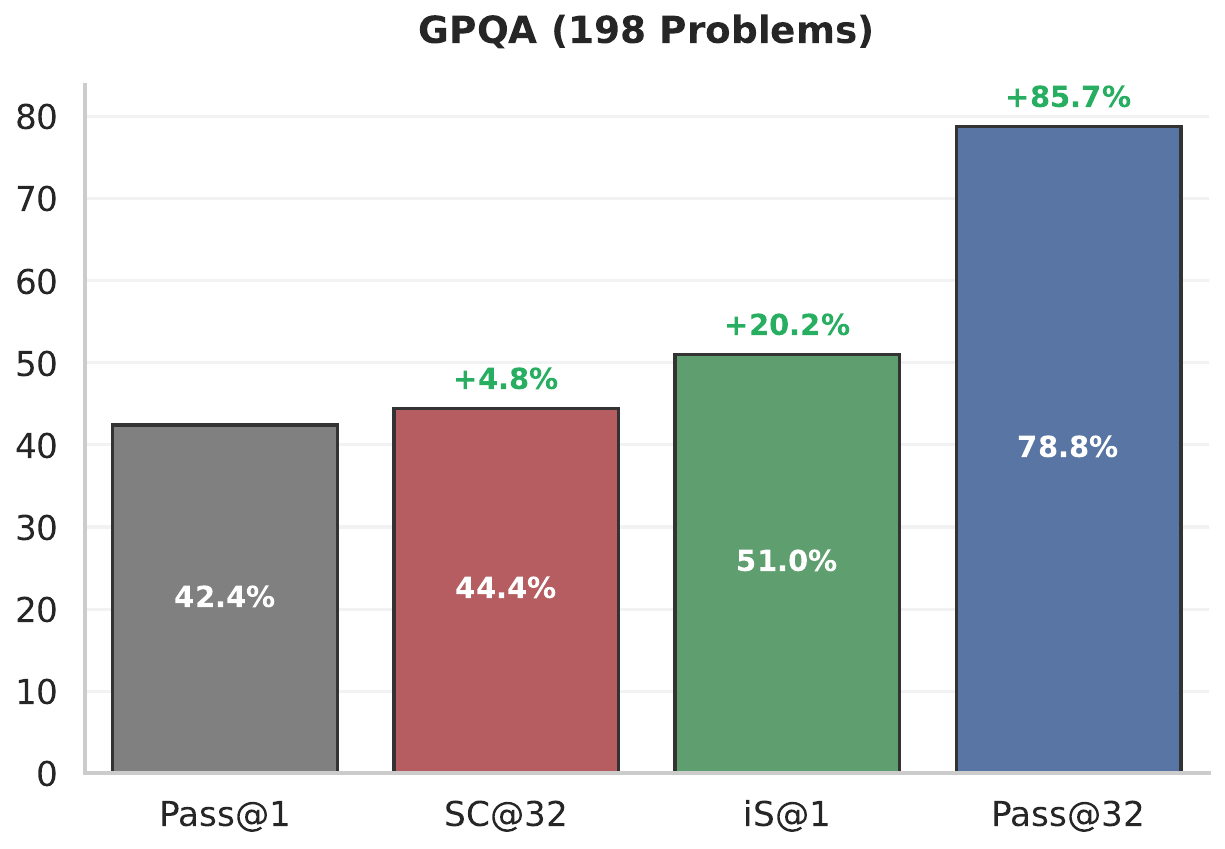}
        \caption{GPQA-Diamond}
        \label{fig:gpqa-intro}
    \end{subfigure}
    \hfill
    \begin{subfigure}[b]{0.3\textwidth}
        \centering
        \includegraphics[width=\textwidth]{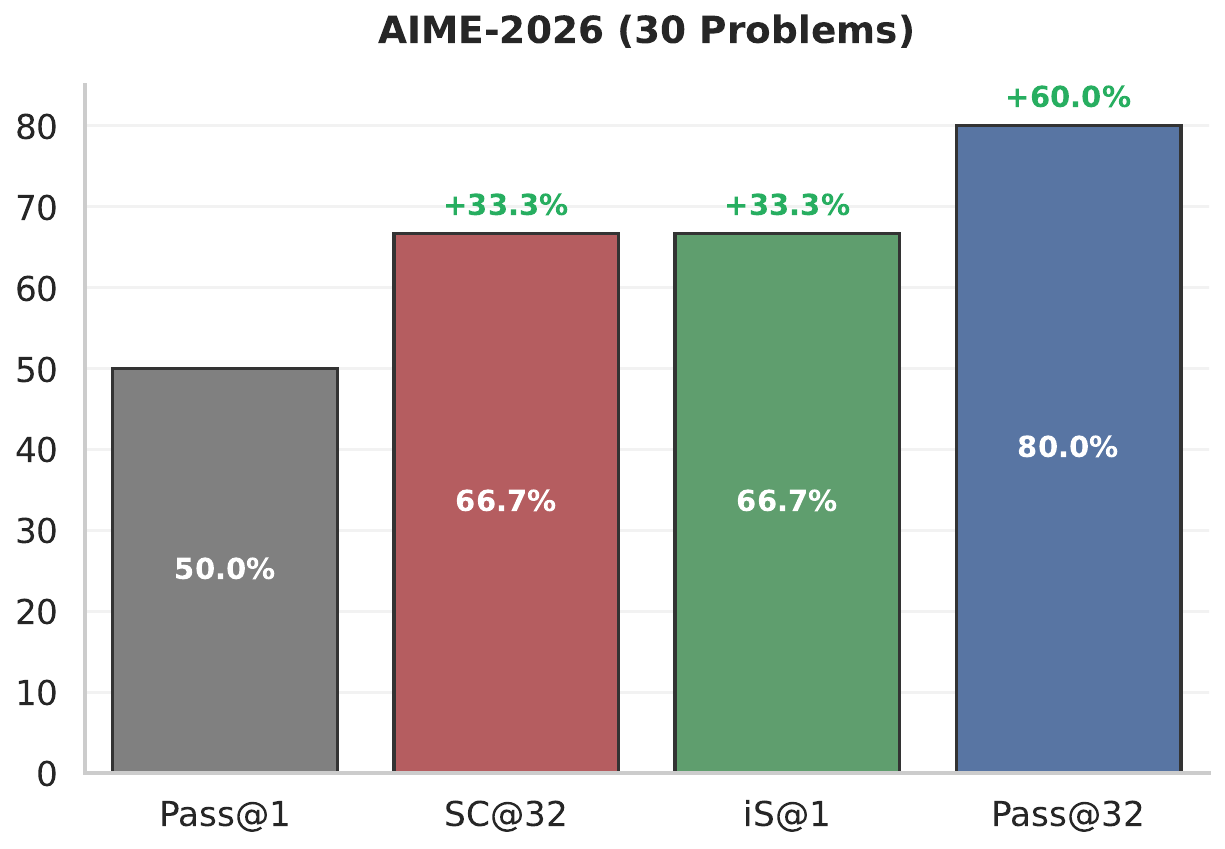}
        \caption{AIME 2026}
        \label{fig:aime-intro}
    \end{subfigure}
    \caption{Accuracy for pass@1, Self-Consistency, \texttt{iS} (ours), and pass@N across recent math and reasoning benchmarks. We evaluate the most recent (2026) math problems to minimize training-data contamination for Qwen3-4B-Instruct-2507 (released in 2025).}
    \label{fig:global}
\end{figure}

\begin{wraptable}{r}{0.4\textwidth}
\centering
\caption{Level of Domain Verifiability.}
\resizebox{\linewidth}{!}{
\begin{tabular}{lll}
\toprule
Domain & Verification & Datasets \\
\midrule
Mathematics & Symbolic & AIME, HMMT \\
Reasoning & Output & GPQA-Diamond \\
Coding & Execution & LiveCodeBench-v6 \\
Healthcare & Semantic & HealthBench-Hard \\
Engineering & Solver & Fusion360 \\
\bottomrule
\end{tabular}
}
\label{tab:verification_pros_cons}
\end{wraptable}
Evaluating across the verifiability spectrum (Appendix~\ref{appx:details} Table~\ref{tab:verification_pros_cons}), we show that: 
\emph{(i)} intrinsic statistics offer a verification-free selection signal rivaling consensus methods (Sec.~\ref{subsec:intrinsic_selection}); 
\emph{(ii)} they reliably gauge problem difficulty for adaptive compute allocation (Sec.~\ref{subsec:difficulty});  and 
\emph{(iii)} verifier-free resampling (\texttt{iPF}) and guided distillation (\texttt{dPF}) enhance inference scaling (Sec.~\ref{subsec:intrinsic_resampling}, \ref{subsec:steering_guidance}).

\paragraph{Models} We evaluate four diverse architectures: {Qwen3-4B-Instruct-2507}~\citep{yang2025qwen3}, {gemma-4-E2B-it}~\citep{team2025gemma}, {medgemma-1.5-4b-it}~\citep{sellergren2026medgemma} (medical), and {Qwen2.5-VL-7B-Instruct}~\citep{bai2025qwen2} (vision-language).

\paragraph{Datasets} We benchmark across five domains using AIME and HMMT~\citep{balunovic2025matharena}, GPQA-Diamond~\citep{rein2023gpqa}, LiveCodeBench-v6~\citep{jain2024livecodebench}, HealthBench-Hard~\citep{arora2025healthbench}, and Fusion360~\citep{willis2021fusion}. These span symbolic, output, execution, semantic, and solver-based verification. Details in Appendix~\ref{appx:details}.

\paragraph{Evaluation} We report pass@1, pass@N, and top@k metrics. We generate $\le 128$ samples for \texttt{iS} selection, and $\le 32$ for \texttt{iPF}/\texttt{dPF} resampling. To optimize quality, we apply logit blending to short sequences and trigger KL-guided resampling for longer outputs (average length $>10 \times 128$-token steps). Step-level methods (\texttt{iPF}, \texttt{dPF}) default to \texttt{iS}-based estimators. Baselines include Self-Certainty~\citep{kang2025scalable}, Self-Consistency~\citep{wang2022self}, and DeepConf~\citep{fu2025deep}.

\subsection{Intrinsic Selection (\texttt{iS}) with Adjusted Tail Entropy}
\label{subsec:intrinsic_selection}

\begin{table}[b]
  \centering
  \caption{Evaluation accuracy (\%) across seven benchmark datasets, reporting mean performance over four independent seeds. Aggregate metrics include the unweighted average (avg), median (med), and dataset-size weighted average (w avg). For \texttt{iS}, we use a single estimator---adjusted tail entropy---across all datasets, without symbolic or output verification. \texttt{iS} remains competitive with a strong Self-Consistency (SC) baseline.}
  \label{tab:main_results}
  \resizebox{\textwidth}{!}{%
  \begin{tabular}{lccccccc | ccc}
    \toprule
    Method & A2024 & A2025 & A2026 & H2025 & H2026 & GPQA & LCB & avg & med & wt avg \\
    \midrule
    pass@1 & $56.5_{\pm 0.4}$ & $45.3_{\pm 0.5}$ & $50.0_{\pm 0.1}$ & $30.2_{\pm 0.3}$ & $28.5_{\pm 0.4}$ & $42.3_{\pm 0.1}$ & $19.0_{\pm 0.9}$ & 38.8 & 42.3 & 37.0 \\
    SC & $72.5_{\pm 1.6}$ & $60.8_{\pm 2.8}$ & $65.8_{\pm 1.6}$ & $30.8_{\pm 0.8}$ & $29.5_{\pm 0.9}$ & $44.1_{\pm 0.6}$ & $18.2_{\pm 1.2}$ & \textbf{46.0} & 44.1 & 40.9 \\
    \texttt{iS}@1 (ours) & $67.5_{\pm 2.1}$ & $49.9_{\pm 2.4}$ & $54.2_{\pm 1.6}$ & $35.0_{\pm 0.9}$ & $34.8_{\pm 0.7}$ & $47.7_{\pm 0.5}$ & $21.5_{\pm 1.2}$ & 44.1 & \textbf{47.7} & \textbf{41.9} \\
    pass@32 & $87.5_{\pm 0.8}$ & $75.8_{\pm 2.1}$ & $83.3_{\pm 1.4}$ & $56.7_{\pm 3.3}$ & $52.3_{\pm 1.5}$ & $79.3_{\pm 0.7}$ & $35.0_{\pm 0.4}$ & 66.8 & 75.8 & 66.4 \\
    \bottomrule
  \end{tabular}
  }
\end{table}

Table~\ref{tab:main_results} reports accuracy on seven benchmark datasets. We summarize performance using three aggregate metrics: the unweighted mean (avg), the median (med), which is less sensitive to outliers, and the dataset-size weighted average (wt avg). For \texttt{iS}, we apply a single estimator across all datasets, without symbolic verification or output matching. Bold values indicate the best practical estimator, excluding the pass@32 upper bound. \texttt{iS} performs robustly across benchmarks, achieving the best median (47.7\%) and weighted average (41.9\%) among practical estimators, while Self-Consistency (SC) obtains the best unweighted average (46.0\%). Both methods substantially outperform the pass@1 baseline. As shown in Figure~\ref{fig:global}, \texttt{iS} consistently narrows the gap between pass@1 and the pass@32 upper bound across all seven benchmarks. 

These results establish intrinsic selection via tail entropy as a strong verification-free alternative to SC. Unlike SC, which relies on output parsing and canonical-form matching, \texttt{iS} operates directly on token-level statistics. This makes it applicable wherever sampling is possible, including open-ended settings in which SC is difficult to apply, providing evidence that the tail-entropy signal reflects a fundamental property rather than an architecture-specific artifact.

\paragraph{Engineering Design}
\begin{figure}[t]
  \centering
  \begin{minipage}[c]{0.58\textwidth}
    \centering
    \captionof{table}{Performance comparison across in-distribution (DeepCAD, finetuning) and out-of-distribution (Fusion360) datasets. We report the mean, median, and Valid Sample Rate (VSR) for pass@1, \texttt{iS}@1, and pass@N.}
    \label{tab:cad-performance}
    \vspace{2mm} %
    \resizebox{\linewidth}{!}{%
    \begin{tabular}{lcccccc}
      \toprule
      & \multicolumn{3}{c}{DeepCAD (in-distro)} & \multicolumn{3}{c}{Fusion360 (out-distro)} \\
      \cmidrule(lr){2-4} \cmidrule(lr){5-7}
      Method & Mean & Median & VSR & Mean & Median & VSR \\
      \midrule
      pass@1 & 0.62 & 0.75 & 97.5\% & 0.41 & 0.38 & 94.4\% \\
      \texttt{iS}@1 (ours) & \textbf{0.66} & \textbf{0.76} & \textbf{100\%} & \textbf{0.45} & \textbf{0.48} & \textbf{97.0\%} \\
      pass@128 & 0.89 & 0.98 & 100\% & 0.83 & 0.88 & 100\% \\
      \bottomrule
    \end{tabular}%
    }
  \end{minipage}
  \hfill %
  \begin{minipage}[c]{0.38\textwidth} 
    \centering
    \includegraphics[width=\linewidth]{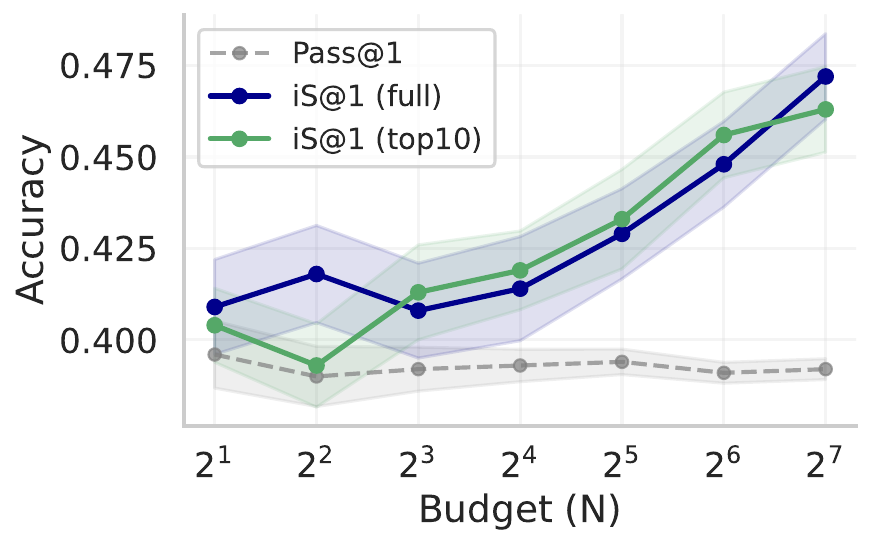}
    \caption{Inference Scaling behavior for image-to-CAD generation using \texttt{iS}@1 on Fusion360 test set.}
    \label{fig:cad-ent1-scaling}
  \end{minipage}
\end{figure}
While generation scaling (pass@128) improves image-to-CAD performance (Table~\ref{tab:cad-performance}), output verification becomes a major computational bottleneck at scale. Executing geometric kernels and computing Intersection over Union (IoU) can account for up to 40\% of the generation loop. In addition, the absence of lightweight reward models in engineering makes standard inference-time scaling methods, such as Best-of-$N$ or step-level reward guidance, especially challenging to apply. 

To address these limitations, \texttt{iS} uses set-level length distributions and adjusted tail entropy to identify high-quality designs. This yields a reliable scaling mechanism without the substantial overhead of external geometric solvers (Table~\ref{tab:cad-performance}). Unlike our other evaluations, the CAD experiments use a custom fine-tuned Qwen2.5-VL model~\citep{doris2025cad}. The strong performance of \texttt{iS} on this specialized multimodal architecture suggests that the selection mechanism is largely agnostic to model family and fine-tuning procedure. As shown in Figure~\ref{fig:cad-ent1-scaling}, \texttt{iS} also scales predictably on Fusion360 test set. As the sample budget $N$ increases, \texttt{iS} steadily narrows the gap to the pass@$N$ oracle, indicating that the intrinsic signal becomes more informative with additional samples, even without a domain-specific reward model. Appendix~\ref{appx:image-to-code} for more details.

\paragraph{Problem Difficulty Estimation}
\label{subsec:difficulty}
We evaluate whether intrinsic statistics can reliably estimate problem difficulty to dynamically route compute (Appendix~\ref{appx:difficulty}). As shown in Table~\ref{tab:tail_entropy_difficulty_results}, length-trimmed tail entropy is a highly reliable difficulty estimator for structured reasoning. It achieves strong ranking quality (AUROC $> 0.79$) and $>90\%$ precision for adaptive compute allocation on AIME-2026 and HMMT-2026.

\paragraph{Ablations} The appendix provides extensive ablations of \texttt{iS}, including comparisons with extrinsic reward models and hybrid selection methods (Appx~\ref{appx:reward_models}), larger sampling budgets (Appx~\ref{appx:budget}), advanced confidence- and consensus-based baselines (Appx~\ref{appx:consensus_confidence}), selection with and without the final answer (Appx~\ref{appx:answer}), robustness, trajectory scoring, and tail selection (Appx~\ref{appx:tail}), and transfer to Gemma-family models (Appx~\ref{appx:gemma}). Additional ablations are provided in Appendix~\ref{appx:additional_experiments}.

\subsection{Intrinsic Particle Filtering (\texttt{iPF}) on Hard Problems}
\label{subsec:intrinsic_resampling}
We evaluate whether intrinsic, entropy-guided step-level resampling (\texttt{iPF}) improves generation quality on problems where the base model rarely succeeds. We compare \texttt{iPF} against a standard i.i.d.\ baseline on the hardest 25\% of the AIME (2024--2026) datasets---where \texttt{iS} underperforms SC (Sec~\ref{subsec:intrinsic_selection} Table~\ref{tab:main_results})---as well as on clinical reasoning tasks from HealthBench-Hard. For \texttt{iPF}, we deploy 16 particles for up to 120 steps (128 tokens per step) and apply systematic resampling with z-score normalization, yielding an approximate step resampling rate of 20\%. 

As detailed in Table~\ref{tab:hard_math_main}, \texttt{iPF} significantly outperforms the 16-sample baseline on structured math tasks, achieving an average +6.1 percentage point gain in pass@1 across the AIME datasets, highlighted by a notable +9.2 point increase on AIME-2026. 

\begin{wraptable}{r}{0.5\textwidth}
  \caption{Accuracy on Hard 25\% Subset (Budget \(N=16\)) for AIME datasets using Intrinsic Particle Filtering (\texttt{iPF}) as resampling mechanism. We run 10 seeds and report Min, Avg, Max performance.
  }
  \label{tab:hard_math_main}
  \centering
  \setlength{\tabcolsep}{3pt}
  \resizebox{\linewidth}{!}{
  \begin{tabular}{llccccccccc}
    \toprule
    &
    & \multicolumn{3}{c}{A24}
    & \multicolumn{3}{c}{A25}
    & \multicolumn{3}{c}{A26} \\
    \cmidrule(lr){3-5} \cmidrule(lr){6-8} \cmidrule(lr){9-11}
    Metric & Sampling & Min & Avg & Max & Min & Avg & Max & Min & Avg & Max \\
    \midrule
    \multicolumn{10}{l}{\textit{Gen Quality}} \\
    pass@1
    & parallel
      & 6.7  & 17.0 & 29.2
      & 1.6  & 4.9  & 9.4
      & 2.0  & 5.6  & 10.5 \\
    pass@1
    &  \texttt{iPF}
      & 10.7 & 22.8 & 36.6
      & 2.0  & 8.3  & 16.1
      & 4.9  & 14.7 & 26.3 \\
    {$\Delta$ pass@1}
    &
      & \multicolumn{3}{c}{\textbf{+5.8}}
      & \multicolumn{3}{c}{\textbf{+3.3}}
      & \multicolumn{3}{c}{\textbf{+9.2}} \\
    \midrule
    \multicolumn{10}{l}{\textit{Self-Consistency}} \\
    SC
    & parallel
      & 7.1  & 21.4 & 35.7
      & 0.0  & 10.7 & 25.0
      & 0.0  & 3.6  & 10.7 \\
    SC 
    & \texttt{iPF}
      & 14.3 & 28.6 & 46.4
      & 0.0  & 10.7 & 25.0
      & 3.6  & 17.9 & 32.1 \\
    {$\Delta$ SC}
    &
      & \multicolumn{3}{c}{\textbf{+7.1}}
      & \multicolumn{3}{c}{0.0}
      & \multicolumn{3}{c}{\textbf{+14.3}} \\
    \midrule
    \multicolumn{10}{l}{\textit{Entropy Selection}} \\
    \texttt{iS}
    & parallel
      & 10.7 & 25.0 & 42.9
      & 0.0  & 7.1  & 17.9
      & 0.0  & 10.7 & 21.4 \\
    \texttt{iS} 
    & \texttt{iPF}
      & 10.7 & 25.0 & 42.9
      & 3.6  & 17.9 & 32.1
      & 3.6  & 14.3 & 28.6 \\
    {$\Delta$ \texttt{iS}}
    &
      & \multicolumn{3}{c}{0.0}
      & \multicolumn{3}{c}{\textbf{+10.7}}
      & \multicolumn{3}{c}{\textbf{+3.6}} \\
    \bottomrule
  \end{tabular}
  }
\end{wraptable}
By iteratively refining trajectories, \texttt{iPF} successfully navigates problems where the standard sampling policy occasionally identifies, but fails to execute, correct approaches. Furthermore, because this resampling mechanism concentrates particles around promising reasoning paths, self-consistency (SC@16) effectively amplifies the correct signal once it is found, driving SC accuracy from 3.6\% to 17.9\% on AIME-2026. 

Crucially, this capability extends to open-ended, hard-to-verify domains. As shown in Figure~\ref{fig:healthbench_bar}, \texttt{iPF} consistently improves problem-weighted rubric scores over the i.i.d.\ baseline on HealthBench-Hard, excelling specifically on reasoning-heavy clinical themes. In contrast, Particle Distillation (\texttt{dPF}, discussed next) provides complementary steering on this same dataset for tasks that demand strict adherence to complex, rubric-specific criteria.

\subsection{Steering Resampling with Particle Distillation (\texttt{dPF})}
\label{subsec:steering_guidance}
\begin{wraptable}{r}{0.5\textwidth}
\centering
\caption{Math particle distillation results comparing \texttt{dPF} (\(N=8\), guided resampling) against \(N=32\) i.i.d.\ samples. Guidance hints are generated via {Qwen3.5-27B}. Crucially, the hint generator, guide \(q\), and sampling model \(p\) access only \(\vh\), without ground-truth answers or extrinsic verification. See Appx~\ref{appx:additional_experiments} Table~\ref{tab:phase2-distill-appx} for ablations.}
\label{tab:phase2-distill}
\resizebox{\linewidth}{!}{%
\begin{tabular}{ll ccccc}
\toprule
\multirow{2}{*}{Metric} & \multirow{2}{*}{Sampling} &  \multicolumn{5}{c}{Dataset} \\
\cmidrule(lr){3-7}
 & & A24 & A25 & A26 & H25 & H26 \\
\midrule
pass@1 & parallel          & 56.7 & 44.3 & 50.1 & 30.0 & 28.2 \\
pass@1 & \texttt{dPF}       & \textbf{64.6} & \textbf{47.9} & \textbf{62.5} & \textbf{32.5} & \textbf{37.1}\\
\midrule
\texttt{iS}@1 & parallel    & 66.7 & 53.3 & 53.3 & \textbf{33.3} & 36.4\\
\texttt{iS}@1 & \texttt{dPF}      & \textbf{76.7} & \textbf{63.3} & \textbf{63.3} & \textbf{33.3} & \textbf{45.5} \\
\midrule
pass@32 & parallel & 86.7 & 70.0 & 80.0 & 56.7 & 51.5\\
\bottomrule
\end{tabular}%
}
\end{wraptable}
We evaluate \texttt{dPF} in two complementary settings prone to \emph{systematic failure}: competition mathematics, where the model systematically errs due to incorrect assumptions or problem complexity, and clinical healthcare, where quality is multidimensional and requires strict adherence to expert rubrics and safety constraints rather than binary verification. In both cases, \texttt{dPF} injects privileged information to steer sampling past these structural bottlenecks.

\paragraph{Mathematical Reasoning (Teacher Guidance)}
For mathematics, we generate hints (e.g., procedures or specific sub-problem strategies) from a strong teacher model using trajectories with the final answers removed. Crucially, neither the hint generator, the guide distribution $q(\vz \mid \vc, \vh)$, nor the sampling model $p_{\theta}(\vz \mid \vc)$ accesses the ground truth answer. 

As detailed in Tables~\ref{tab:phase2-distill} and~\ref{tab:phase2-distill-appx}, distilling this privileged guidance allows the model to correct faulty procedural assumptions early in the trajectory. With just 8 particles, \texttt{dPF} consistently matches or exceeds the 32-sample i.i.d.\ baseline across all metrics, achieving a 4--5 point improvement in weighted average pass@1. We find that logit blending is particularly effective on the hardest datasets, driving the largest performance gains (e.g., +8.9 points on HMMT-2026). 

Furthermore, critique-based hints and teacher demonstrations offer complementary benefits depending on the domain's specific failure mode: critiques excel on AIME, where identifying faulty assumptions is critical, while teacher demonstrations perform better on HMMT, where the solution paths themselves are highly non-obvious. 

\begin{wrapfigure}{r}{0.4\textwidth}
    \centering
    \includegraphics[width=.9\linewidth]{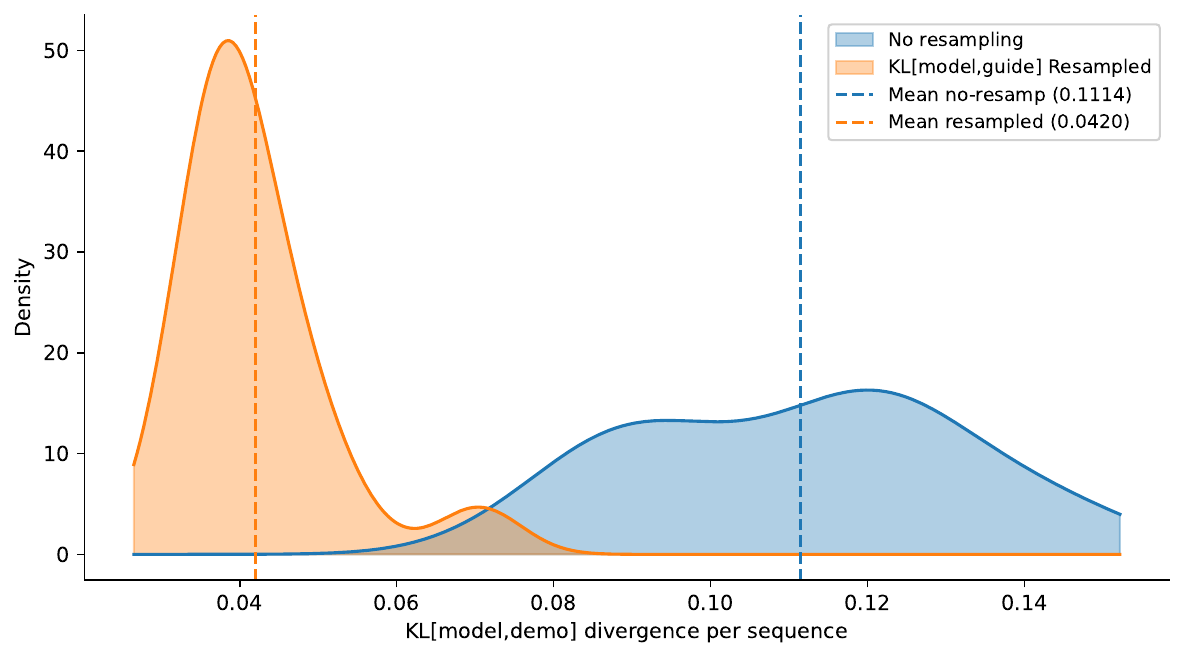}
  \caption{KL divergence between the base and hint-guided models on the unsolved AIME problem 2024-II-15. Guided resampling (orange) shifts particles toward low-divergence regions (mean KL: 0.042 vs.\ 0.111), indicating that \texttt{dPF} steers generation toward trajectories consistent with the privileged hint.}
    \label{fig:kl-resampling-kde}
\end{wrapfigure}

Finally, Figure~\ref{fig:kl-resampling-kde} illustrates this steering mechanism on an unsolved AIME problem (pass@$N = 0$). By applying guided resampling, \texttt{dPF} shifts the KL distribution from a diffuse regime (mean 0.111) to a concentrated, low-divergence one (mean 0.042), effectively forcing particles toward trajectories where the base model and the guide agree.

\begin{figure}[t]
    \centering
    \includegraphics[width=\linewidth]{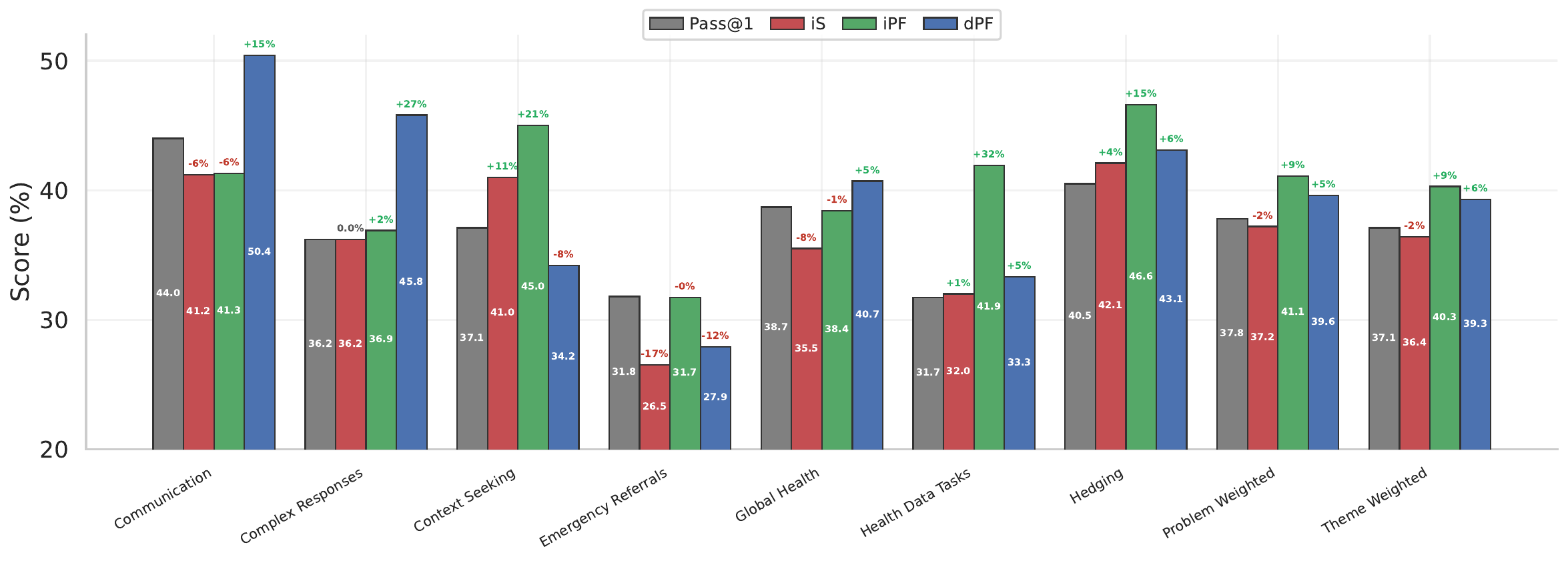}
    \caption{HealthBench Results using \texttt{iPF} and \texttt{dPF}.}
    \label{fig:healthbench_bar}
\end{figure}

\paragraph{Clinical Healthcare (Rubric Guidance)}
To test transferability, we evaluate HealthBench-Hard - comprising 100 problems across seven clinical themes - using MedGemma-4B-IT for generation and MedGemma-27B as the judge. Despite the shift in domain and architecture, both \texttt{iPF} and \texttt{dPF} yield substantial gains without requiring any retuning of the particle filtering hyperparameters. 
We compare a 32-sample i.i.d.\ baseline against \texttt{iPF} ($N=16$) and \texttt{dPF} ($N=8$). In this setting, \texttt{dPF} employs a rubric-conditioned KL divergence strategy: rather than using teacher hints, a guide model conditioned solely on the problem's rubric criteria directs resampling and logit blending over the first 25\% of steps.

As shown in Figure~\ref{fig:healthbench_bar} and Appendix Table~\ref{tab:healthbench_mean-appx}, \texttt{iPF} emerges as the strongest overall method, achieving a 0.411 mean score (+8.7\% over pass@1). It specifically excels on themes requiring deep reasoning - such as health data tasks (+32.2\%) and context seeking (+21.3\%) - where sustained low entropy strongly correlates with successfully completing a logical chain. Meanwhile, \texttt{dPF} serves as the second-best overall strategy (+4.8\% over pass@1), but dominates on themes with concrete, checkable criteria. For instance, the rubric-conditioned guide provides a highly targeted signal for communication (+14.5\%) and complex responses (+26.5\%), successfully steering generation to include mandatory pharmacological or guideline-based content.
Conversely, standard \texttt{iS} degrades performance by -1.5\% compared to pass@1.

In HealthBench, single-forward-pass entropy tends to select the most ``confident'' response, but confidence alone does not guarantee multidimensional rubric coverage. This contrast perfectly highlights the complementary nature of our framework: while \texttt{iS} serves as a highly effective selection and routing mechanism for scalar quality metrics, \texttt{iPF} and \texttt{dPF} are essential for driving generation quality in complex, open-ended spaces.

\section{Conclusion and Limitations}
\label{sec:conclusion}

We present a framework for inference-time scaling that uses intrinsic model statistics - in particular, length distributions and tail entropy - to estimate problem difficulty and rank candidates without process-level rewards. For more challenging tasks, Intrinsic Particle Filtering (\texttt{iPF}) performs step-level resampling using internal signals, while Particle Distillation (\texttt{dPF}) steers generation with privileged information. Together, these components recast inference-time compute as a practical and robust mechanism driven by intrinsic signals.

\paragraph{Limitations} 
Our resampling weights are heuristic and do not define a proper logit distribution, which may introduce instability. Difficulty estimation also incurs an upfront cost of $N$ parallel samples. In addition, \texttt{dPF} depends on privileged information, such as teacher hints, that may be unavailable in open-ended settings. Future work includes using intrinsic statistics as training-time signals to complement reinforcement learning, enabling adaptive compute routing at deployment, and replacing external teacher hints with self-generated critiques.

\clearpage
\small
\bibliographystyle{abbrvnat}
\bibliography{biblio}

\clearpage
\appendix
\tableofcontents

\clearpage
\section{Extended Related Work}
\label{appx:related-work}

\paragraph{Domain Verifiability}
Appendix Table~\ref{tab:verifiability-appx} contrasts verifiable and hard-to-verify domains. Verifiable fields, such as mathematics and coding, offer canonical outputs that provide cheap, automated reward signals (\(V(x,y) \in \{0,1\}\)) for reinforcement learning and inference-time filtering. In contrast, extending inference-time scaling to hard-to-verify or open-ended domains - such as healthcare and engineering design - presents major challenges~\citep{jaech2024openai}. These settings lack cheap verification, forcing reliance on expensive subjective rubrics, human intervention, LLM-as-a-judge pipelines~\citep{wang2025mcts}, or computationally heavy geometric simulators. High verification costs bottleneck scalability, and in these environments, external reward models are prone to over-optimization~\citep{gao2023scaling,el2024goodhart} and generalize poorly~\citep{park2025know}. Furthermore, practical verification remains fragile even in structured fields: heuristic parsers break easily, and solver-based validation can take hours per sample, effectively rendering them hard-to-verify at scale~(Appendix~\ref{appx:verifiability}).

\paragraph{Inference-Time Scaling}
Recent advancements have demonstrated that allocating additional compute during generation, often termed test-time or inference-time scaling, significantly improves LLM reasoning performance~\citep{snell2024scaling, brown2024large, wu2025inference}. Scaling strategies typically fall into two categories: parallel sampling, where multiple independent trajectories are generated to form a consensus, and sequential step-level searches, such as Tree of Thoughts~\citep{yao2023tree} or Monte Carlo Tree Search (MCTS~\citep{misaki2025wider}), which systematically explore reasoning trees. While parallel scaling is straightforward and highly effective for standard problems, step-level methods navigate high-density regions of the solution space to tackle harder tasks where correct trajectories lie in low-probability regions.

\paragraph{Consensus-based Selection}
Self-Consistency (SC)~\citep{wang2022self} is the most successful verification-free approach in modern parallel inference, marginalizing over diverse reasoning trajectories via majority voting to achieve strong results whenever answers admit a canonical form. It marginalizes latent trajectories \(\{\vz\}^K_{k=1} \sim p(\vz \mid \vc)\) to approximate the conditional answer distribution:
\[
p(\vx \mid \vc) = \sum_{\vz} p(\vx \mid \vz, \vc) p(\vz \mid \vc)
\]
In practice, systems extract consensus via majority voting over a discrete set of \(M\) choices \(\{a_m\}^{M}_{m=1}\):
\[
x_a^* = \arg\max_{a} \sum^M_{m=1} \mathbb{I}(a_m = a)
\]
While inexpensive and robust to noise, SC is heavily confined to domains with cheap syntactic verification, which can be non-trivial even in mathematics. Universal Self-Consistency~\citep{chen2023universal} extends this to open-ended tasks by using an LLM to judge agreement among candidates. Other works group answers by meaning to compute semantic entropy~\citep{kuhn2023semantic}, which helps measure uncertainty but requires expensive clustering. Fundamentally, these approaches depend on output-level comparison or LLM-based parsing, limiting their applicability when outputs are long-form, multi-part, or lack a single correct answer. Our intrinsic selection sidesteps this by scoring at the token-probability level, requiring no output parsing.

\paragraph{Confidence-based Selection}
Building on early findings that base LLMs are often well-calibrated~\citep{kadavath2022language}, confidence-based methods leverage the model's internal signals to rank solutions, bypassing external verifiers. Reasoning models trained via reinforcement learning~\citep{guo2025deepseek,jaech2024openai} intrinsically verify their reasoning during generation. Orthogonally, approaches like Self-Certainty~\citep{kang2025scalable} measure the divergence from a uniform distribution to the model's predictive distribution to gauge intrinsic confidence. For a probabilistic model defined over a discrete vocabulary \(\mathcal{V}\) and a token sequence of length \(T\), denoted as \(p(\vz) = p(\vz_T) \prod^{T-1}_{t=1} p(\vz_t \mid \vz_{<t})\), we define the entropy at token \(t\) as~\citep{shannon1948mathematical}:
\[
    \mathbb{H}[\vz_t \mid \vz_{<t}] = - \sum_{v \in \mathcal{V}} p(\vz_t = v \mid \vz_{<t}) \log p(\vz_t = v \mid \vz_{<t})
\]
This can also be expressed as a negated KL divergence relative to a uniform distribution \(U\):
\[
\mathbb{KL}[p, U]_t = \sum_{v \in \mathcal{V}} p(\vz^v_t) \log \dfrac{p(\vz^v_t)}{1/|\mathcal{V}|} = - \mathbb{H}\left[p\right]_t + \log(|\mathcal{V}|)
\]
A sequence demonstrates relative confidence when its entropy is lower, or its KL divergence from uniform is larger, compared to competing sequences. More complex pipelines, such as DeepConf~\citep{fu2025deep} and TokUR~\citep{zhang2025tokur}, utilize trace-level scores from token probabilities and entropy-aware generation to identify promising candidates without extrinsic rewards. Our approach complements RL-trained models; while they internalize verification, our work advances this shift by introducing a purely intrinsic, length-adjusted tail entropy metric that robustly isolates reasoning quality, generalizing these ideas to step-level resampling and extracting discriminative signals from the collective statistics of an \(N\)-sample set.

\paragraph{Particle-based Inference and Guidance}
Guided decoding seeks to iteratively steer generation using dynamic feedback. Drawing from Bayesian inference, Sequential Monte Carlo (SMC)~\citep{doucet2001introduction,zhao2024probabilistic,lew2023sequential} methods leverage external rewards to define tilting distributions and step-level resampling. Particle Filtering (PF), a subset of SMC, aims to approximate the posterior \(p(\vz \mid \vc, \vo) \propto p(\vo \mid \vz, \vc) p(\vz \mid \vc)\) via sequential importance resampling~\citep{liu2001theoretical}, introducing a proposal \(q(\vz \mid \vc)\):
\[
w_{t} \propto w_{t-1}~\dfrac{p(\vz_{t} \mid \vz_{t-1}, \vc, \vo_{t})}{q(\vz_{t} \mid \vz_{t-1}, \vc)}
\]
Choosing the model itself as the proposal, the weight update reduces to \(w_t \propto w_{t-1}\,p(\vo_t \mid \vz_t, \vc)\), where the observation model can be parameterized by a Process Reward Model (PRM), i.e., \(r_{\phi}(\vz_t, \vc) = p(\vo_t = 1 \mid \vz_t, \vc)\)~\citep{lightman2023let,puri2025probabilistic}. However, training PRMs is expensive, domain-specific, and requires calibration, as they often degrade and become overconfident on long sequences outside their training distribution~\citep{zhang2025lessons,park2025know}. To overcome this, recent approaches steer generation using intrinsic metrics~\citep{wang2025think,sanyal2025mixing} or Contrastive Decoding~\citep{li2022contrastive, o2023contrastive}. Our logit blending approach draws a direct parallel to classifier-free guidance for language models~\citep{sanchez2023stay, ho2022classifier}, interpolating an unconditional base policy \(p_\theta\) and a hint-guided policy \(q(\mathbf{z} \mid \mathbf{c}, \mathbf{h})\) to steer trajectories away from incorrect early assumptions without relying on external PRMs.

\clearpage
\section{Verifiability}
\label{appx:verifiability}
The distinction between verifiable and non-verifiable tasks fundamentally dictates a model's capacity to scale during post-training. In verifiable domains, the availability of ``cheap approximate verification''---such as executing unit tests or running mathematical solvers---provides a low-cost, high-speed mechanism for evaluating outputs. Because grading each sample costs mere fractions of a cent and milliseconds of compute, researchers can feasibly execute millions of rollouts for Reinforcement Learning (RL) or build vast search trees at inference time. Conversely, non-verifiable tasks rely on expensive and slow evaluation methods, such as human annotators or large LLM-as-a-Judge systems. This high cost acts as a strict bottleneck, making it computationally and financially prohibitive to scale post-training optimization to the same massive extent.
\begin{table}[ht!]
\centering
\caption{Conceptual Map: Domain Verifiability.}
\label{tab:verifiability-appx}
\resizebox{\textwidth}{!}{%
\begin{tabular}{@{}p{3.2cm} p{5.8cm} p{5.8cm}@{}}
\toprule
{Feature} & {Verifiable Domains} & {Hard-to-Verify Domains} \\
\midrule
\textbf{Domain Examples} & Engineering, Math, Reasoning, Code & Healthcare, Enterprise \\
\addlinespace
\textbf{Verification Methods} & Solvers, Symbolic Output, Test Execution. & LLM-as-Verifier, Human Evaluation. Model-based \\
\addlinespace
\textbf{Core Characteristics} & Output Canonicalization, Reference Solutions & No Canonical Form (Open-ended), No References (relies on rubrics) \\
\addlinespace
\textbf{Cost \& Scalability} & Cheap Approx Verification $\pi(x)$. Scaling computationally viable. Reward Models.  & Expensive verification creates a strict bottleneck, severely limiting scalability. \\
\addlinespace
\textbf{Ranking \& Selection} & Allows for absolute ranking and selection based on exact correctness. & No absolute ranking or selection without verification; strictly relative preferences. \\
\addlinespace
\textbf{Training Dynamics} & 
\(f(x) \leftrightarrow y\). Exact verifier \(V(x,y) \in \{0,1\}\) provides the high-throughput reward signal needed for Reinforcement Learning (RL). & 
Optimization relies on proxy rewards. Highly challenging to scale post-training without cheap verification. \\
\addlinespace
\textbf{Inference Dynamics} & 
\(\pi(x) \rightarrow \hat{y}\). Allows for inference-time scaling and filtering using \(V(x,\hat{y})\). & 
\(\pi(x) \rightarrow ?\) Exact correctness is undefined or subjective at test time. \\
\bottomrule
\end{tabular}
}
\end{table}

Training, Validating and Evaluating foundation models fundamentally diverges based on the intrinsic verifiability of the target task. In domains such as mathematics, software engineering, and formal reasoning, tasks are highly verifiable because they possess canonical forms and definitive reference solutions. This structure allows for inexpensive, scalable verification methods like symbolic solvers or direct test execution, where a model trained on a deterministic mapping $f(x) \leftrightarrow y$ produces a readily testable inference output $\pi(x) \rightarrow \hat{y}$. Conversely, complex applications in healthcare and enterprise contexts push evaluation beyond traditional verifiability. These hard-to-verify tasks typically lack exact reference solutions or canonical forms, relying instead on open-ended generation or relative preferences where the definitive correctness of an inference output remains ambiguous ($\pi(x) \rightarrow ?$). Because these subjective domains preclude cheap, programmatic approximate verification, evaluating model performance necessitates highly expensive, rubric-driven mechanisms, primarily relying on expert human evaluators or LLM-as-a-Judge frameworks.

\begin{figure}[ht!]
    \centering
    \includegraphics[width=\linewidth]{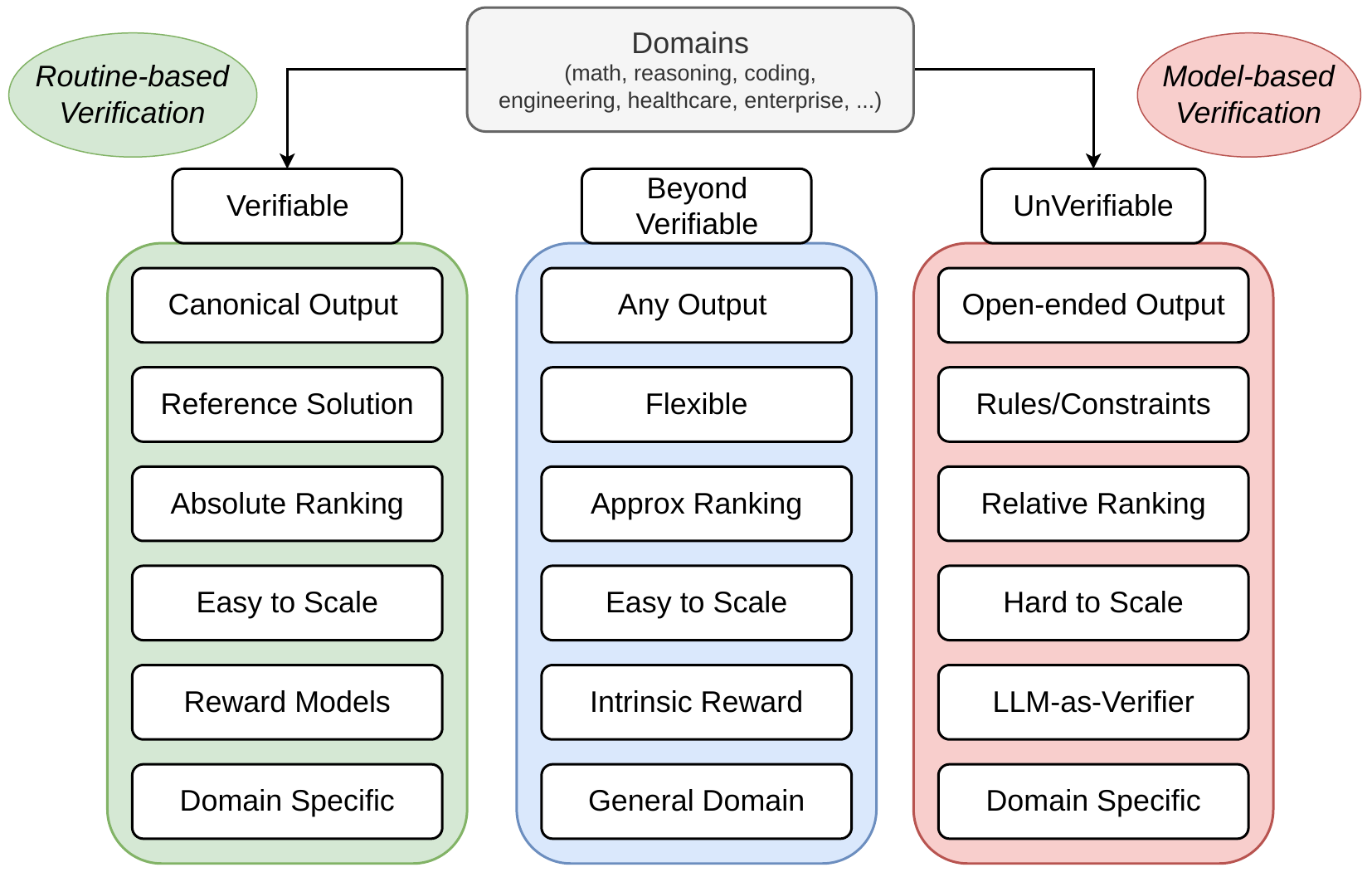}
    \caption{Beyond Domain Verifiability. We would like to build ITS methods that can approximate verifiability for non-verifiable domains, without requiring human intervention or expensive extrinsic reward.}
    \label{fig:domain_verifiability-appx}
\end{figure}

\subsection{The Bottleneck of Extrinsic Verification}
Relying on extrinsic verification presents significant limitations for scaling. Training PRMs and ORMs is expensive~\citep{zhang2025lessons}, require calibration~\citep{park2025know}, and become increasingly difficult to build as the underlying models and reasoning budgets grow stronger. Furthermore, rubric-driven reward models generalize poorly and fail entirely on the hardest problems where the base success rate is zero. Most importantly, extrinsic verification is largely unavailable for open-ended or non-verifiable domains.

To understand this bottleneck, we categorize solution domains by their verification complexity (Table \ref{tab:verification_pros_cons}). Traditionally, domains are considered verifiable if a post-processing routine can extract a canonical representation for the generated output and if a reference solution for the given problem exists, or non-verifiable/unverifiable if no such routine exists. However, verifiability is ultimately a function of available resources and compute~\footnote{Notice that in practice almost nothing is strictly unverifiable if enough time, human expertise, and computing power are available. The true constraint for scaling is the cost, automation level, and availability of verification at training and validation time. We will see examples of such problem in engineering.}. 

A more practical distinction is between easy-to-verify domains at scale, where checking a solution is much cheaper than generating it, and hard-to-verify domains at scale, where the verification process itself is complex, computationally expensive, or requires expert judgment. Doing so, we can think of domain verification as a spectrum. For example, evaluating the aerodynamics of an engineering design is technically verifiable (and deterministic) through a fluid dynamics simulation, making it easy-to-verify at the prototype level. However, running a high-fidelity simulation takes hours and massive compute, making it act as a hard-to-verify domain at scale, when there is need to evaluate thousand parallel designs s of generated by a custom foundation model. Such compute requirements make large-scale RL training or model validation hard to scale, making extrinsic verification a bottleneck.

\subsection{Challenges in Verifiable Domains}
Even among domains traditionally considered verifiable, practical implementation can be challenging. Consensus algorithms~\citep{wang2022self,fu2025deep} rely heavily on symbolic and output verification, meaning solutions must be parsed and matched at the string level. In mathematics, this requires post-processing tools like \texttt{math\_verify}\footnote{\texttt{https://github.com/huggingface/Math-Verify}} to normalize solutions into canonical forms. These heuristic-based parsers are often brittle, and differing implementations, model behaviors or edge cases can drastically alter results, forcing practitioners to rely on custom solutions.

Other verifiable domains present their own distinct hurdles. Coding requires access to executable test environments, while engineering requires expensive solvers or slow simulators that limit large-scale verification. Among hard-to-verify domains like healthcare and enterprise tool calling, progress has been even slower. Verification in these areas requires checking for correctness at the trajectory level, rubric-informed scoring, safety requirements, and open-ended nuances, which demands expensive human-in-the-loop systems or complex LLM-judge pipelines. Finally, even when output verification is cheap and easy, verifying intermediate steps (reasoning trajectories, mathematical proofs) can still be challenging and require large amount of compute and specialized post-processing pipelines.

For these reasons, a general method that can approximate verification at inference time would be of great utility, irrespective of the level of verifiability of the underlying domain.

\clearpage
\section{Perspective on Intrinsic Scoring and Resampling}
\label{app:derivation}

This section provides a compact interpretation of the intrinsic scoring and resampling rules used by iS, iPF, and dPF. The goal is not to derive a formal guarantee of correctness, but to clarify the statistical intuition behind the surrogate quantities used in the main method.

Let $\vc$ denote a problem instance and let $\vz = (\vz_1,\dots,\vz_T)$ be a sampled trajectory from the base model $p(\vz \mid \vc)$. Recall that the token-level entropy is:
\begin{equation}
\mathbb{H}_t(\vz) = \mathbb{H}(\vz_t \mid \vz_{<t}, \vc).
\end{equation}

For \texttt{iS}, the method scores each sampled trajectory using an aggregated entropy over an adaptively chosen suffix region. Writing $A(z) \subseteq \{1,\dots,T\}$ for the tail index set selected by the procedure in Section~\ref{sec:method}, the resulting score is
\begin{equation}
S(\vz) = \mathrm{agg}_{t \in A(\vz)} \mathbb{H}_t(\vz),
\end{equation}
where $\mathrm{agg}$ is the same robust aggregation operator used in the manuscript.

The underlying intuition is that, for many reasoning tasks, the final answer region is more informative than the earlier exploratory portion of the trajectory. Under this view, lower suffix entropy serves as a proxy for local confidence on the part of the sequence most relevant to the final output. This is precisely the quantity used by \texttt{iS} to rank candidates.

The adaptive tail construction can be interpreted as follows: if the tail window is too long, the score is contaminated by earlier reasoning tokens that are weakly related to the final answer. If the window is too short, the resulting estimate of suffix confidence becomes noisy. The heuristic cutoff used in the paper therefore aims to select a region that is long enough to stabilize the score while remaining focused on the answer-bearing suffix.

This same perspective also helps explain the benefit of response-length filtering and robust aggregation. In practice, sampled trajectories may include pathological short or long traces, for example due to truncation, formatting failures, or runaway reasoning. Trimming such outliers before estimating set-level statistics reduces distortion in the tail estimate, while median-style aggregation reduces sensitivity to localized entropy spikes.

For \texttt{iPF}, the method replaces an external reward model with entropy-based pseudo-observations during resampling. A useful way to interpret this update is as sampling from a confidence-tilted trajectory distribution of the form
\begin{equation}
\tilde p(\vz \mid \vc) \propto p(\vz \mid \vc)\exp\left(-\lambda \sum_{t=1}^{T} \mathbb{H}_{t}(\vz)\right),
\end{equation}
where $\mathbb{H}_{t}(\vz)$ denotes the standardized last step entropy statistic used in the resampling rule, and $\lambda > 0$ is an implicit temperature induced by normalization.

This interpretation should be read as heuristic rather than exact. The paper does not claim that iPF recovers a true posterior over correct trajectories, and the resampling weights are not calibrated likelihoods. Instead, iPF biases computation toward continuations that remain locally stable under the base model, which is consistent with its empirical role on problems where standard sampling occasionally finds the correct direction but fails to sustain it.

For dPF, the same logic applies with privileged guidance. After the initial logit-blending phase, the method performs resampling using KL divergence between the base policy and a guide conditioned on hints or rubric information. This can be interpreted as favoring trajectories that stay close to a guide-aware policy during the most consequential part of generation:
\begin{equation}
\tilde p_h(\vz \mid \vc, \vh) \propto p(\vz \mid \vc)\exp\left(
-\lambda \sum_{t=1}^{T}
D_{\mathrm{KL}}\!\left(
p_t(\cdot \mid \vz_{<t}, \vc)\,\|\,q_t(\cdot \mid \vz_{<t}, \vc, \vh)
\right)
\right),
\end{equation}
where $q_t$ denotes the guide policy conditioned on privileged information $\vh$. Notice that the guide $q_t$ is the policy model $p_{\theta}$ conditioned on in-context information.

Again, this expression is best understood as an interpretation of the steering effect rather than a formal target distribution. In particular, dPF does not attempt exact imitation of the guide at every step. Rather, it preferentially retains trajectories whose local decisions remain compatible with useful guidance early enough to avoid systematic failure.

Taken together, these interpretations clarify the intended operating regime of the framework. Intrinsic selection is most natural when suffix-level uncertainty is informative about answer quality, while entropy-guided or KL-guided resampling is most useful when additional inference-time compute can concentrate mass on promising reasoning paths. Conversely, when confidence is poorly aligned with quality or when trajectories have little variation in their answer-bearing region, the intrinsic signal may be weaker, which is consistent with the empirical limitations discussed in the manuscript.

\paragraph{Answer Region Scaling}
The adaptive tail cutoff, $t_c$, is designed to isolate the concluding answer region without incorporating noise from the preceding exploratory reasoning trace. For a generated trajectory of total length $L$, two scaling properties emerge: the total Shannon entropy grows linearly, $\mathcal{O}(L)$, reflecting the cumulative information volume, whereas the geometric dispersion in the latent space scales sublinearly, $\|\mathbf{z}_{\text{answer}}\|_2 \approx \sigma\sqrt{L}$, following standard concentration bounds for independent increments~\citep{vershynin2018high}. Because our intrinsic scoring mechanism ranks samples based on their relative separability within the entropy landscape - a geometric property rather than a strictly volumetric one - we adopt the sublinear scaling factor, establishing $t_c \propto \sqrt{L}$.

\paragraph{Preventing Premature Collapse in \texttt{iPF}}
Standard sequential particle filtering is notoriously susceptible to sample impoverishment, wherein resampling dynamics greedily trap the generation process in sub-optimal local states~\citep{doucet2001introduction, doucet2001sequential}. This risk is particularly acute in our verifier-free setting, as intrinsic entropy signals are inherently noisier than explicit, calibrated reward models. A single transient drop in entropy could theoretically force the particle set to collapse entirely around an incomplete or flawed reasoning path. To maintain broad exploration over long generation horizons, we modify the resampling distribution to dynamically normalize weights and explicitly detect absorbing configurations. By intervening before particle diversity catastrophically collapses, our intrinsic particle filtering (\texttt{iPF}) preserves alternative reasoning pathways without requiring computationally expensive Markov Chain Monte Carlo (MCMC) rejuvenation steps~\citep{hastings1970monte} or rigid thresholding algorithms~\citep{gilks1995markov}, thus fully retaining the verifier-free benefits of intrinsic scoring~\citep{lightman2023let, wang2024math}.

\clearpage
\section{Algorithms}
\label{appx:algo}

\subsection{Intrinsic Selection}

\begin{algorithm}[ht!]
\caption{Intrinsic Selection (\texttt{iS})}
\label{alg:is}
\begin{algorithmic}[1]
\Require Problem $\mathbf{c}$, model $p_\theta$, sample budget $N$
\Ensure Selected solution $\mathbf{z}^*$

\Statex \textit{// Parallel generation}
\For{$n = 1, \ldots, N$}
    \State Generate $\mathbf{z}_n \sim p_\theta(\cdot \mid \mathbf{c})$
    \State Record $l_n \gets l(\mathbf{z}_n)$ and
           $\{\mathbb{H}[\mathbf{z}_{n,t} \mid \mathbf{z}_{n,<t}]\}_{t=1}^{l_n}$
\EndFor

\Statex \textit{// Adaptive tail window}
\State $\bar{l} \gets N^{-1} \sum_{n} l_n$
    \Comment{mean length}
\State $\bar{h} \gets N^{-1} \sum_{n}
    \bigl( l_n^{-1} \sum_{t} \mathbb{H}[\mathbf{z}_{n,t}] \bigr)$
    \Comment{mean entropy}
\State $t_c \gets \bigl\lfloor \sqrt{\bar{l}} \,/\, \bar{h} \bigr\rfloor$
    \Comment{tail cutoff (Eq.~\ref{eq:tail})}

\Statex \textit{// Score and select}
\For{$n = 1, \ldots, N$}
    \State $\mathcal{A}_n \gets
        \{t \in \mathbb{Z} \mid t_c \le t \le l_n\}$
        \Comment{tail index (Eq.~\ref{eq:tail-entropy})}
    \State $s_n \gets \operatorname*{agg}_{t \in \mathcal{A}_n}\;
        \mathbb{H}[\mathbf{z}_{n,t} \mid \mathbf{z}_{n,<t}]$
\EndFor
\State \Return $\mathbf{z}^* \gets
    \mathbf{z}_{n^*}$ where $n^* = \operatorname*{arg\,min}_{n}\; s_n$
    \Comment{lowest tail entropy} %
\end{algorithmic}
\end{algorithm}

\clearpage
\subsection{Intrinsic Particle Filtering}

\begin{algorithm}[ht!]
\caption{Intrinsic Particle Filtering (\texttt{iPF})}
\label{alg:ipf}
\begin{algorithmic}[1]
\Require Problem $\mathbf{c}$, model $p_\theta$, particles $N$,
         max steps $S$, tokens per step $B$
\Ensure Refined particle set $\{\mathbf{z}_n\}_{n=1}^N$

\State $w_n \gets 1/N$ for all $n$  %
\For{$t = 1, \ldots, S$}
    \For{$n = 1, \ldots, N$}
        \State Generate $B$ tokens:
            $\mathbf{z}_n^{(t)} \sim
            p_\theta(\cdot \mid \mathbf{z}_n^{(<t)}, \mathbf{c})$
        \State $h_n \gets \operatorname{agg}_{b \in [B]}\;
            \mathbb{H}[\mathbf{z}_{n,b}^{(t)} \mid
            \mathbf{z}_{n,<b}^{(t)}]$
            \Comment{step entropy}
    \EndFor
    \State $\tilde{h}_n \gets
        (h_n - \bar{h}) / (\mathrm{std}(h) + \epsilon)$
        for all $n$
        \Comment{z-score}
    \State $w_n \gets \sigma(-\tilde{h}_n)$
        \Comment{weight update (Eq.~\ref{eq:ipf})} %
    \State Resample indices $\{n'\}$ from
        $\mathrm{Cat}(w_1, \ldots, w_N)$
    \State $\mathbf{z}_n^{(\le t)} \gets
        \mathbf{z}_{n'}^{(\le t)}$ for all $n$
\EndFor
\State \Return $\{\mathbf{z}_n\}_{n=1}^N$;
\end{algorithmic}
\end{algorithm}

\clearpage
\subsection{Particle Distillation}

\begin{algorithm}[ht!]
\caption{Particle Distillation (\texttt{dPF})}
\label{alg:dpf}
\begin{algorithmic}[1]
\Require Problem $\mathbf{c}$, hint $\mathbf{h}$, model $p_\theta$,
         guide $q(\cdot \mid \mathbf{c}, \mathbf{h})$,
         particles $N$, steps $S$, blend steps $S_b$, tokens per step $B$
\Ensure Steered particle set $\{\mathbf{z}_n\}_{n=1}^N$
\State Initialize $N$ particles; $w_n \gets 1/N$ for all $n$
\For{$t = 1, \ldots, S$}
    \If{$t \le S_b$}
        \Comment{Phase 1: logit blending}
        \State $\alpha_t \gets 1 - t \,/\, S_b$
            \Comment{anneal from guide toward base (Eq.~\ref{eq:dpf-logits})} %
        \For{$n = 1, \ldots, N$}
            \State $\ell_n \gets \alpha_t \cdot
                \mathrm{logits}_{p_\theta}(\mathbf{z}_n, \mathbf{c})
                + (1 - \alpha_t) \cdot
                \mathrm{logits}_q(\mathbf{z}_n, \mathbf{c}, \mathbf{h})$
            \State Sample $B$ tokens from
                $\mathrm{softmax}(\ell_n)$
        \EndFor
    \Else
        \Comment{Phase 2: KL-guided resampling}
        \For{$n = 1, \ldots, N$}
            \State Generate $B$ tokens from
                $p_\theta(\cdot \mid \mathbf{z}_n^{(<t)}, \mathbf{c})$
            \State $d_n \gets \mathbb{KL}\bigl[
                p_\theta(\mathbf{z}_n^{(t)} \mid \mathbf{z}_n^{(<t)}, \mathbf{c})
                \,\|\,
                q(\mathbf{z}_n^{(t)} \mid \mathbf{z}_n^{(<t)}, \mathbf{c}, \mathbf{h})
                \bigr]$ %
        \EndFor
        \State $\tilde{d}_n \gets
            (d_n - \bar{d}) / (\mathrm{std}(d) + \epsilon)$
            for all $n$
        \State $w_n \gets \sigma(\exp(-\tilde{d}_n))$
            \Comment{cumulative weight update (Eq.~\ref{eq:dpf})} %
        \State Resample from
            $\mathrm{Cat}(w_1, \ldots, w_N)$
    \EndIf
\EndFor
\State \Return $\{\mathbf{z}_n\}_{n=1}^N$
\end{algorithmic}
\end{algorithm}

\clearpage
\section{Code for Intrinsic Selection}

\begin{lstlisting}[style=pythonstyle, caption={iS@1 selection. Given $N$ candidate responses with per-token entropy, select the one with lowest aggregated entropy in an adaptive tail window.}, label={lst:tail-entropy}]
import numpy as np

def adaptive_tail_window(
    entropies: list[list[float]], included: list[int],
    tail_min: int = 64, tail_max: int = 2048,
) -> int:
    lengths: list[int] = [len(entropies[i]) for i in included]
    mean_len: float = float(np.mean(lengths))
    mean_ent: float = float(np.mean([
        np.mean(entropies[i])
        for i in included if entropies[i]]))
    mean_ent = max(mean_ent, 1e-6)
    return int(np.clip(
        np.sqrt(mean_len) / mean_ent, tail_min, tail_max))

def trim_length_outliers(
    lengths: list[int], trim_pct: float = 0.05,
) -> list[int]:
    n: int = len(lengths)
    if n < 16:
        return list(range(n))
    lo = float(np.percentile(lengths, trim_pct * 100))
    hi = float(np.percentile(lengths, (1-trim_pct) * 100))
    inc = [i for i in range(n) if lo <= lengths[i] <= hi]
    return inc if len(inc) >= 2 else list(range(n))

def tail_scores(
    entropies: list[list[float]], included: list[int],
    tail: int, agg: str = "median",
) -> list[float]:
    inc: set[int] = set(included)
    fn = np.median if agg == "median" else np.mean
    scores: list[float] = []
    for i in range(len(entropies)):
        if i not in inc or not entropies[i]:
            scores.append(float("inf"))
        else:
            scores.append(float(fn(entropies[i][-tail:])))
    return scores

def select_by_tail_entropy(
    entropies: list[list[float]],
    tail_min: int = 64, tail_max: int = 2048,
    agg: str = "median",
) -> tuple[int, list[float]]:
    lengths: list[int] = [len(e) for e in entropies]
    included = trim_length_outliers(lengths)
    tail = adaptive_tail_window(
        entropies, included, tail_min, tail_max)
    scores = tail_scores(entropies, included, tail, agg)
    return int(np.argmin(scores)), scores
\end{lstlisting}

\clearpage
\section{Intrinsic Selection and Resampling: Workflow Analysis}

\subsection{Graphical Model}

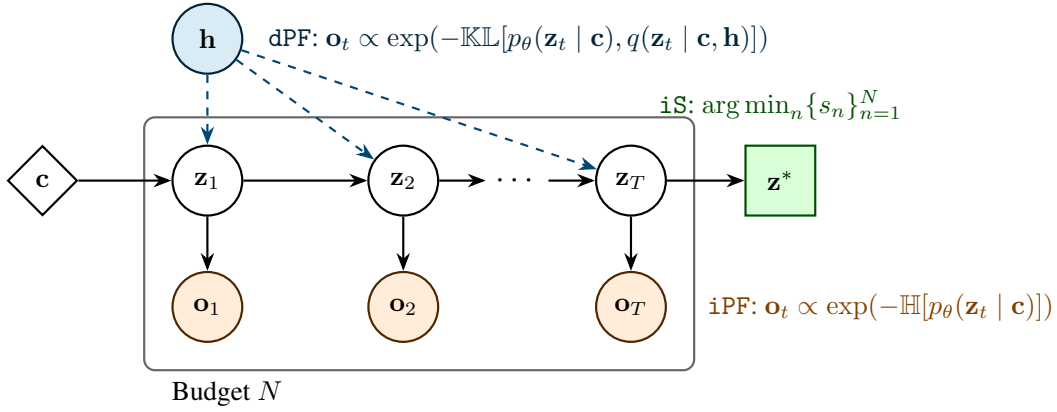
\begin{figure}[ht!]
\centering
\resizebox{\linewidth}{!}{%
\begin{tikzpicture}[
    >=Stealth,
    node distance=1.4cm and 1.6cm,
    latent/.style={
        circle, draw, thick, minimum size=0.9cm
    },
    cond/.style={
        diamond, draw, thick, minimum size=0.9cm
    },
    observed/.style={
        circle, draw, thick, minimum size=0.9cm,
        fill=black!12,
    },
    deterministic/.style={
        rectangle, draw, thick, minimum size=0.9cm,
    },
    plate/.style={
        rectangle, draw, rounded corners=4pt,
        inner sep=10pt, thick, draw=black!60
    },
    arrow/.style={->, thick, >=Stealth},
    intrinsic/.style={->, thick, >=Stealth},
    guide/.style={->, thick, >=Stealth, draw=blue!60!black, dashed},
    platelabel/.style={text=black!30!black},
]

\node[cond] (c) {$\mathbf{c}$};

\node[latent, right=1.2cm of c] (z1) {$\mathbf{z}_1$};
\node[latent, right=of z1] (z2) {$\mathbf{z}_2$};
\node[right=0.6cm of z2, font=\large] (dots) {$\cdots$};
\node[latent, right=0.6cm of dots] (zT) {$\mathbf{z}_T$};

\draw[arrow] (c) -- (z1);
\draw[arrow] (z1) -- (z2);
\draw[arrow] (z2) -- (dots);
\draw[arrow] (dots) -- (zT);

\node[observed, below=0.7cm of z1, draw=orange!30!black,fill=orange!15]
    (o1) {$\mathbf{o}_1$};
\node[observed, below=0.7cm of z2, draw=orange!30!black,fill=orange!15]
    (o2) {$\mathbf{o}_2$};
\node[observed, below=0.7cm of zT, draw=orange!30!black, fill=orange!15]
    (oT) {$\mathbf{o}_T$};

\draw[intrinsic] (z1) -- (o1);
\draw[intrinsic] (z2) -- (o2);
\draw[intrinsic] (zT) -- (oT);

\node[observed, above=0.9cm of z1, fill=blue!15, draw=blue!30!black]
    (h) {$\mathbf{h}$};

\draw[guide] (h) -- (z1);
\draw[guide] (h) to[bend left=0] (z2);
\draw[guide] (h) to[bend left=0] (zT);

\node[deterministic, right=1.0cm of zT, fill=green!15, draw=green!30!black]
    (sel) {$\mathbf{z}^*$};

\draw[arrow] (zT) -- (sel);

\begin{scope}[on background layer]
    \node[plate, fit=(z1)(z2)(zT)(o1)(o2)(oT)(dots),
          label={[platelabel]below left: Budget $N$}]{};
\end{scope}

\node[above=0.2cm of sel, text=green!30!black, align=left]
    {\texttt{iS}: $\argmin_n \{s_n\}^N_{n=1}$};

\node[right=0.4cm of oT, text=orange!50!black, align=center]
    {\texttt{iPF}: $\mathbf{o}_t \propto \exp(-\mathbb{H}[p_{\theta}(\mathbf{z}_t \mid \mathbf{c} )])$};

\node[right=0.2cm of h, text=blue!30!black, align=center]
    {\texttt{dPF}: $\mathbf{o}_t \propto \exp(-\mathbb{KL}[p_{\theta}(\mathbf{z}_t \mid \mathbf{c}), q(\mathbf{z}_t \mid \mathbf{c}, \mathbf{h})])$};

\end{tikzpicture}%
}
\caption{
Probabilistic graphical model for our intrinsic selection and resampling methods.
Step trajectories $\mathbf{z}_1, \ldots, \mathbf{z}_T$ are generated 
conditioned on the prompt $\mathbf{c}$. At each step $t$, intrinsic pseudo-observations 
$\mathbf{o}_t$ (gray nodes) are attached to the latents, representing step-wise intrinsic 
signals used by \textcolor{orange!30!black}{\textbf{iPF}}. The hint node $\mathbf{h}$ (blue) 
injects privileged information that guides early latent variables through dashed arrows, 
corresponding to \textcolor{blue!30!black}{\textbf{dPF}}. The selection node $\mathbf{z}^*$ 
(green) denotes the chosen trajectory at the end of the chain, as in 
\textcolor{green!30!black}{\textbf{iS}}.
}
\label{fig:pgm}
\end{figure}

\subsection{Workflow}

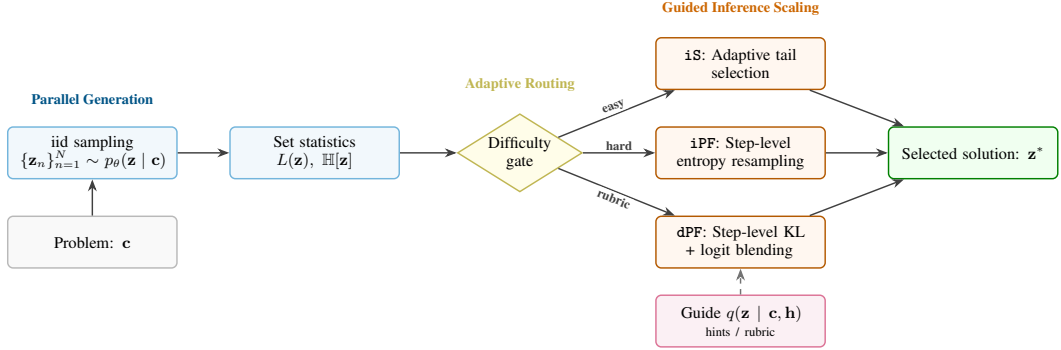
\begin{figure}[ht!]
\centering
\resizebox{\linewidth}{!}{%
\begin{tikzpicture}[
    >=Stealth,
    node distance=7mm and 10mm,
    every node/.style={font=\small},
    box/.style={
        rectangle,
        rounded corners=3pt,
        draw,
        thick,
        minimum height=10mm,
        text width=3.0cm,
        align=center,
        inner xsep=4pt,
        inner ysep=3pt
    },
    phase/.style={
        box,
        fill=blue!5,
        draw=blue!55
    },
    context/.style={
        box,
        fill=gray!5,
        draw=gray!55
    },
    method/.style={
        box,
        fill=orange!6,
        draw=orange!70!black
    },
    output/.style={
        box,
        fill=green!6,
        draw=green!50!black
    },
    hintbox/.style={
        box,
        fill=purple!6,
        draw=purple!55
    },
    decision/.style={
        diamond,
        draw=yellow!70!black,
        fill=yellow!10,
        thick,
        aspect=1.6,
        align=center,
        inner sep=1pt,
        minimum width=18mm,
        minimum height=13mm
    },
    arrow/.style={->, thick, draw=black!75},
    dasharrow/.style={->, thick, dashed, draw=black!55},
    lab/.style={
        font=\scriptsize\bfseries,
        text=black!75,
        fill=white,
        inner sep=1pt
    }
]

\node[context] (input) {Problem: $\mathbf{c}$};

\node[phase, above=of input] (sample) {iid sampling $ \{\vz_n\}^N_{n=1}\sim p_\theta(\vz \mid \vc)$};

\node[phase, right=of sample] (stats) {Set statistics\\$L(\mathbf{z}),\ \mathbb{H}[\mathbf{z}]$};

\node[decision, right=11mm of stats] (diff) {Difficulty\\gate};

\node[method, right=14mm of diff] (ipf) {\textbf{\texttt{iPF}}: Step-level\\entropy resampling};

\node[method, above=7mm of ipf] (is) {\textbf{\texttt{iS}}: Adaptive tail\\selection};

\node[method, below=7mm of ipf] (dpf) {\textbf{\texttt{dPF}}: Step-level KL\\+ logit blending};

\node[hintbox, below=5mm of dpf] (hint) {Guide $q(\mathbf{z}\mid\mathbf{c},\mathbf{h})$\\{\scriptsize hints / rubric}};

\node[output, right=12mm of ipf] (out) {Selected solution: $\mathbf{z}^*$};

\draw[arrow] (input) -- (sample);
\draw[arrow] (sample) -- (stats);
\draw[arrow] (stats) -- (diff);

\draw[arrow] (diff) -- node[lab, above, sloped] {easy} (is);
\draw[arrow] (diff) -- node[lab, above] {hard} (ipf);
\draw[arrow] (diff) -- node[lab, below, sloped] {rubric} (dpf);

\draw[arrow] (is) -- (out);
\draw[arrow] (ipf) -- (out);
\draw[arrow] (dpf) -- (out);

\draw[dasharrow] (hint) -- (dpf);

\node[above=3mm of sample, font=\footnotesize\bfseries, text=blue!70!black]
    {Parallel Generation};

\node[above=3mm of diff, font=\footnotesize\bfseries, text=yellow!70!black]
    {Adaptive Routing};

\node[above=3mm of is, font=\footnotesize\bfseries, text=orange!80!black]
    {Guided Inference Scaling};

\end{tikzpicture}
}
\caption{
    \textbf{Adaptive Intrinsic Inference Scaling Pipeline.}
    Given a problem $\mathbf{c}$, set-level statistics from an initial parallel pass provide a difficulty gate that dynamically routes compute.
    Easy problems are resolved instantly via Intrinsic Selection (\texttt{iS}).
    Hard problems trigger step-level Intrinsic Particle Filtering (\texttt{iPF}) driven by entropy.
    Tasks requiring specific adherence to multidimensional criteria utilize Particle Distillation (\texttt{dPF}) guided by privileged hints or rubrics.
    All paths yield a refined solution $\mathbf{z}^*$ without external reward models.
}
\label{fig:pipeline}
\end{figure}

\clearpage
\subsection{Parallel Sampling}
\begin{table}[ht!]
    \centering
    \caption{AIME 2024 Statistics (Hard problems: 4/30) sorted by number of correct solutions. Notice how methods based on Self-Consistency are not robust to syntactic output diversity, and cases where $pass@N = 1$ but $pass@1 \rightarrow n/N$ with $n<<N$ and $n \in [1, N/2]$. 
    In these cases, most selection methods break and $top@k \rightarrow 0$ for $k=\{1,2,3\}$. We define as hard problems where $pass@1 \rightarrow n/N$, with $n=\{0,1,2\}$. Notice how the hardness of a problem (under the model distribution) is strongly correlated with $pass@1$ (extrinsic metric) and inversely correlated with adjusted tail entropy (intrinsic metric). When verification is not available or hard, we can use the tail entropy as a proxy for hardness, without the need or external verifiers or ground truth. 
    }
\begin{tabular}{lccccc}
\toprule
{uid} & {n\_correct} & {n\_samples} & {pass@1} & {pass@N} & {is\_hard} \\
\midrule
2024-I-12 & 0 & 32 & 0.000 & 0 & True \\
2024-I-11 & 0 & 32 & 0.000 & 0 & True \\
2024-II-15 & 0 & 32 & 0.000 & 0 & True \\
2024-II-9 & 1 & 32 & 0.031 & 1 & True \\
\midrule
2024-II-10 & 3 & 32 & 0.094 & 1 & False \\
2024-II-8 & 3 & 32 & 0.094 & 1 & False \\
2024-I-8 & 3 & 32 & 0.094 & 1 & False \\
2024-I-9 & 5 & 32 & 0.156 & 1 & False \\
2024-I-10 & 10 & 32 & 0.312 & 1 & False \\
2024-I-13 & 10 & 32 & 0.312 & 1 & False \\
2024-II-11 & 11 & 32 & 0.344 & 1 & False \\
2024-I-14 & 13 & 32 & 0.406 & 1 & False \\
2024-II-12 & 14 & 32 & 0.438 & 1 & False \\
2024-II-7 & 15 & 32 & 0.469 & 1 & False \\
\midrule
2024-II-14 & 18 & 32 & 0.562 & 1 & False \\
2024-II-3 & 22 & 32 & 0.688 & 1 & False \\
2024-II-1 & 25 & 32 & 0.781 & 1 & False \\ 
2024-II-13 & 26 & 32 & 0.812 & 1 & False \\
2024-II-2 & 26 & 32 & 0.812 & 1 & False \\
2024-II-5 & 28 & 32 & 0.875 & 1 & False \\
2024-I-5 & 28 & 32 & 0.875 & 1 & False \\
2024-I-7 & 28 & 32 & 0.875 & 1 & False \\
2024-II-6 & 29 & 32 & 0.906 & 1 & False \\
2024-I-1 & 30 & 32 & 0.938 & 1 & False \\ 
2024-I-6 & 31 & 32 & 0.969 & 1 & False \\
2024-I-3 & 31 & 32 & 0.969 & 1 & False \\
2024-I-4 & 31 & 32 & 0.969 & 1 & False \\
\midrule
2024-I-2 & 32 & 32 & 1.000 & 1 & False \\
2024-I-15 & 32 & 32 & 1.000 & 1 & False \\
2024-II-4 & 32 & 32 & 1.000 & 1 & False \\
\bottomrule
\end{tabular}
\end{table}

\clearpage
\subsection{Intrinsic Selection}
\begin{table}[ht!]
    \centering
    \caption{Detailed Problem Results. 
    We rank each problem by adjusted mean tail entropy. We can see that there is strong correlation between low pass@1 (pass\_rate), hardness of the problem, and increase in mean tail entropy. 
    So we can use such quantity (for example selecting the worst 10/20\% of the samples) as proxy for problem complexity, select the hard problems and use more compute for inference-time scaling (larger N, step-level steering, or using demonstrations). 
    }
    \resizebox{\textwidth}{!}{%
    \begin{tabular}{llccccccccccccc}
        \toprule
        & & & & \multicolumn{3}{c}{{Entropy}} & \multicolumn{3}{c}{{Certainty}} & \multicolumn{3}{c}{{Length}} & \\
        \cmidrule(lr){5-7} \cmidrule(lr){8-10} \cmidrule(lr){11-13}
        {uid} & {is\_hard} & {n\_correct} & {pass\_rate} & {Mean} & {Std} & {Sum Mean} & {Mean} & {Std} & {Sum Mean} & {Mean} & {Std} & {CV} & {n\_samples} \\
        \midrule
        2024-II-9  & True  & 1  & 0.03 & 0.36 & 0.10 & 506.08 & 28.27 & 2.14 & 39345.19 & 24162.72 & 5319.21 & 0.22 & 32 \\
        2024-II-15 & True  & 0  & 0.00 & 0.35 & 0.06 & 486.94 & 28.51 & 1.86 & 39690.78 & 23192.78 & 3515.64 & 0.15 & 32 \\
        2024-I-12  & True  & 0  & 0.00 & 0.32 & 0.08 & 451.34 & 28.94 & 1.49 & 40288.83 & 19258.34 & 1208.11 & 0.06 & 32 \\
        2024-I-11  & True  & 0  & 0.00 & 0.29 & 0.10 & 406.70 & 28.10 & 3.07 & 39119.66 & 25822.25 & 3908.03 & 0.15 & 32 \\
        \midrule
        2024-I-8   & False & 3  & 0.09 & 0.27 & 0.07 & 379.02 & 28.12 & 2.44 & 39149.96 & 26822.72 & 2417.06 & 0.09 & 32 \\
        2024-I-6   & False & 31 & 0.97 & 0.27 & 0.04 & 355.44 & 34.18 & 1.47 & 45486.80 & 4343.81  & 649.49  & 0.15 & 32 \\
        2024-II-1  & False & 25 & 0.78 & 0.26 & 0.04 & 316.24 & 33.26 & 1.43 & 40237.73 & 8421.97  & 4786.78 & 0.57 & 32 \\
        2024-I-15  & False & 32 & 1.00 & 0.26 & 0.05 & 359.33 & 29.52 & 2.42 & 41094.30 & 18204.16 & 1822.06 & 0.10 & 32 \\
        2024-II-8  & False & 3  & 0.09 & 0.26 & 0.07 & 357.10 & 28.24 & 4.04 & 39316.17 & 26011.81 & 3266.31 & 0.13 & 32 \\
        2024-II-2  & False & 26 & 0.81 & 0.26 & 0.05 & 356.77 & 31.01 & 2.78 & 43168.00 & 16381.31 & 5945.18 & 0.36 & 32 \\
        2024-II-5  & False & 28 & 0.88 & 0.25 & 0.08 & 354.26 & 33.52 & 2.54 & 46657.12 & 13180.72 & 4949.95 & 0.38 & 32 \\
        2024-I-5   & False & 28 & 0.88 & 0.24 & 0.10 & 339.43 & 34.07 & 3.38 & 47428.58 & 12911.13 & 3516.81 & 0.27 & 32 \\
        2024-I-9   & False & 5  & 0.16 & 0.24 & 0.08 & 177.87 & 33.98 & 2.69 & 25441.03 & 3821.31  & 4808.08 & 1.26 & 32 \\
        2024-II-11 & False & 11 & 0.34 & 0.23 & 0.05 & 326.24 & 32.72 & 2.68 & 45550.70 & 17319.03 & 2986.65 & 0.17 & 32 \\
        2024-II-14 & False & 18 & 0.56 & 0.22 & 0.05 & 306.68 & 30.29 & 1.42 & 42165.72 & 19931.22 & 1763.68 & 0.09 & 32 \\
        2024-I-4   & False & 31 & 0.97 & 0.21 & 0.03 & 256.61 & 35.09 & 0.79 & 41620.87 & 3603.72  & 611.90  & 0.17 & 32 \\
        2024-II-7  & False & 15 & 0.47 & 0.20 & 0.05 & 275.74 & 29.95 & 3.02 & 41689.50 & 19006.72 & 3434.10 & 0.18 & 32 \\
        2024-II-3  & False & 22 & 0.69 & 0.19 & 0.07 & 266.54 & 34.73 & 2.86 & 48350.23 & 11397.84 & 6344.92 & 0.56 & 32 \\
        2024-II-10 & False & 3  & 0.09 & 0.19 & 0.07 & 266.02 & 31.44 & 3.31 & 43762.68 & 17632.31 & 3025.67 & 0.17 & 32 \\
        2024-I-13  & False & 10 & 0.31 & 0.18 & 0.05 & 253.45 & 35.12 & 1.59 & 48670.15 & 11595.38 & 4451.38 & 0.38 & 32 \\
        2024-I-3   & False & 31 & 0.97 & 0.17 & 0.03 & 239.67 & 34.97 & 0.70 & 48684.48 & 12489.47 & 2042.10 & 0.16 & 32 \\
        2024-I-10  & False & 10 & 0.31 & 0.17 & 0.07 & 234.81 & 33.48 & 4.42 & 46601.31 & 20580.13 & 2871.57 & 0.14 & 32 \\
        2024-II-12 & False & 14 & 0.44 & 0.15 & 0.07 & 206.22 & 33.24 & 4.95 & 46274.16 & 18409.88 & 5885.25 & 0.32 & 32 \\
        2024-II-6  & False & 29 & 0.91 & 0.14 & 0.03 & 195.71 & 39.17 & 1.06 & 54521.82 & 5705.25  & 1973.39 & 0.35 & 32 \\
        2024-I-2   & False & 32 & 1.00 & 0.13 & 0.05 & 172.35 & 38.52 & 2.38 & 46360.82 & 5611.31  & 3042.39 & 0.54 & 32 \\
        2024-II-13 & False & 26 & 0.81 & 0.11 & 0.03 & 152.72 & 37.99 & 2.34 & 52887.98 & 9952.16  & 3460.27 & 0.35 & 32 \\
        2024-I-14  & False & 13 & 0.41 & 0.11 & 0.06 & 149.41 & 38.58 & 3.29 & 53706.37 & 11407.22 & 4367.01 & 0.38 & 32 \\
        2024-I-1   & False & 30 & 0.94 & 0.09 & 0.02 & 119.32 & 40.58 & 0.73 & 54157.97 & 3566.91  & 989.76  & 0.28 & 32 \\
        2024-I-7   & False & 28 & 0.88 & 0.09 & 0.02 & 118.09 & 40.32 & 0.82 & 53401.82 & 3110.28  & 565.81  & 0.18 & 32 \\
        2024-II-4  & False & 32 & 1.00 & 0.06 & 0.02 & 78.75  & 42.47 & 0.81 & 58968.78 & 3775.69  & 559.02  & 0.15 & 32 \\
        \bottomrule
    \end{tabular}%
    }
\end{table}

\clearpage
\subsection{Intrinsic Selection for Hard Problems}
\begin{table}[ht!]
\centering
\setlength{\tabcolsep}{3pt}
\caption{Entropy statistics and correct answer tracking across particles for problem \texttt{2024-II-9}. 
For this problem, extrinsic and intrinsic selection methods (self-consistency, self-certainty, entropy, output reward models) fail. This problem is hard to solve using inference-time scaling alone.}
\resizebox{\textwidth}{!}{
\begin{tabular}{cccccccc}
\toprule
& & & & \multicolumn{4}{c}{Entropy} \\
\cmidrule(lr){5-8}
{particle} & {correct} & {answer} & {n\_tokens} & {Mean} & {Tail Mean} & {Tail Median} & {Tail p90} \\
\midrule
0  & False & 1022 & 8032 & 0.3721 & 0.2417 & 0.0015 & 0.9007 \\
1  & False & 2 & 8039 & 0.4095 & 0.2727 & 0.0092 & 0.9706 \\
2  & False & 252 & 3667 & 0.3079 & 0.2819 & 0.0002 & 1.0730 \\
3  & False & 32282 & 9359 & 0.3766 & 0.2870 & 0.0050 & 1.0191 \\
4  & False & 900 & 8850 & 0.4270 & 0.3048 & 0.0090 & 1.0742 \\
5  & False & 1022 & 6710 & 0.4122 & 0.3098 & 0.0042 & 1.1130 \\
6  & False & 34 & 7300 & 0.4253 & 0.3152 & 0.0121 & 1.0785 \\
7  & False & 900 & 6151 & 0.4498 & 0.3191 & 0.0066 & 1.0749 \\
8  & False & 570 & 7051 & 0.4462 & 0.3211 & 0.0039 & 1.1135 \\
9  & False & 1022 & 9017 & 0.4052 & 0.3249 & 0.0597 & 1.0323 \\
10 & False & 512 & 10050 & 0.4236 & 0.3262 & 0.0423 & 1.0472 \\
\midrule
11 & True  & \color{dartmouthgreen}{902} & 7397 & 0.4526 & 0.3388 & 0.0217 & 1.1343 \\ \midrule
12 & False & 1142 & 8570 & 0.4218 & 0.3445 & 0.0149 & 1.1611 \\
13 & False & 900 & 6942 & 0.4267 & 0.3465 & 0.0294 & 1.1598 \\
14 & False & 1022 & 8525 & 0.4308 & 0.3594 & 0.0754 & 1.1023 \\
15 & False & 122 & 7711 & 0.4096 & 0.3639 & 0.0841 & 1.1132 \\
16 & False & 482 & 5072 & 0.4187 & 0.3850 & 0.0637 & 1.2463 \\
17 & False & 2 & 7035 & 0.5032 & 0.3983 & 0.1442 & 1.1851 \\
18 & False & 126 & 6832 & 0.4476 & 0.4052 & 0.0963 & 1.2312 \\
19 & False & 2 & 8636 & 0.3920 & 0.4109 & 0.1091 & 1.2209 \\
20 & False & 2 & 6848 & 0.4688 & 0.4329 & 0.1629 & 1.2490 \\
21 & False & 2 & 5104 & 0.4701 & 0.4463 & 0.1606 & 1.2983 \\
22 & False & 1922 & 5162 & 0.4756 & 0.4521 & 0.1641 & 1.3111 \\
23 & False & 1022 & 5315 & 0.4892 & 0.4535 & 0.1660 & 1.3235 \\
24 & False & 1024 & 7999 & 0.4447 & 0.4540 & 0.2047 & 1.2958 \\
25 & False & 2 & 5939 & 0.4547 & 0.4559 & 0.1638 & 1.3046 \\
26 & False & 2 & 5525 & 0.5062 & 0.4654 & 0.2123 & 1.3222 \\
27 & False & 1024 & 6274 & 0.4634 & 0.4743 & 0.2272 & 1.3574 \\
28 & False & 2 & 6083 & 0.4666 & 0.4900 & 0.2468 & 1.3810 \\
29 & False & 2 & 5819 & 0.5136 & 0.4939 & 0.2955 & 1.3238 \\
30 & False & 2 & 4633 & 0.4789 & 0.4945 & 0.2297 & 1.4065 \\
31 & False & 2 & 4186 & 0.5228 & 0.5481 & 0.3036 & 1.5093 \\
\bottomrule
\end{tabular}
}
\label{tab:entropy_data}
\end{table}

\clearpage
\subsection{Hint Generation}

The guide $q(\vz | \vc, \vh)$ sees the problem $\vc$ as conditioning (exactly as the model we are sampling)
\begin{mdframed}[
  linecolor=gray,
  linewidth=2pt,
  roundcorner=10pt,
  backgroundcolor=lightgray!15,
  frametitle={Problem AIME~\texttt{2024-II-9}},
  frametitlerule=true,
  frametitlebackgroundcolor=gray!20
]
There is a collection of $25$ indistinguishable white chips and $25$ indistinguishable black chips. Find the number of ways to place some of these chips in the $25$ unit cells of a $5 \times 5$ grid such that: each cell contains at most one chip all chips in the same row and all chips in the same column have the same colour any additional chip placed on the grid would violate one or more of the previous two conditions.
\end{mdframed}

and privileged information $\vh$ concatenated, in the form of an hint or procedure for critical issues in the problem:

\begin{mdframed}[
  linecolor=gray,
  linewidth=2pt,
  roundcorner=10pt,
  backgroundcolor=lightgray!15,
  frametitle={Hint for Problem AIME~\texttt{2024-II-9}},
  frametitlerule=true,
  frametitlebackgroundcolor=gray!20
]
<hint>
The key insight is that maximality forces every row and every column to be non-empty, meaning each is assigned exactly one color. A cell gets a chip if and only if its row and column share the same color. So the problem reduces to choosing subsets of rows and columns to be 'white' vs 'black', with edge cases when all rows are the same color. 

Each row must be uniformly white or black (not empty, by maximality), and similarly for columns. Consider how choosing which rows are white and which columns are white determines the entire chip placement.
</hint>
\end{mdframed}

\clearpage
\subsection{Particle Distillation for Hard Problems}

\begin{table}[ht!]
\centering
\caption{Entropy statistics and correct answer tracking for particle distillation. We use logits blending for the first 25\% of the sampling steps with $\alpha=0.7$ and linear annealing schedule: $\alpha~\texttt{logits}(\vz, \vc, \vh) + (1-\alpha)~\texttt{logits}(\vz, \vc)$.
}
\resizebox{\textwidth}{!}{
\begin{tabular}{cccccccc}
\toprule
& & & & \multicolumn{4}{c}{{Entropy}} \\
\cmidrule(lr){5-8}
{particle} & {correct} & {answer} & {n\_tokens} & {Mean} & {Tail Mean} & {Tail Median} & {Tail p90} \\
\midrule
0 & True & \color{dartmouthgreen}{902}  & 8189 & 0.3866 & 0.2576 & 0.0024 & 0.9026 \\ %
\midrule
1 & True & 902 & 8189 & 0.3866 & 0.2576 & 0.0024 & 0.9026 \\
2 & False & 1022 & 9710 & 0.4293 & 0.2896 & 0.0260 & 0.9853 \\
3 & False & 1022 & 4899 & 0.4882 & 0.4379 & 0.1297 & 1.2943 \\
4 & False & 1022 & 4899 & 0.4882 & 0.4379 & 0.1297 & 1.2943 \\
5 & False & 1022 & 4899 & 0.4882 & 0.4379 & 0.1297 & 1.2943 \\
6 & False & 1022 & 4899 & 0.4882 & 0.4379 & 0.1297 & 1.2943 \\
7 & False & 1022 & 4899 & 0.4882 & 0.4379 & 0.1297 & 1.2943 \\
\bottomrule
\end{tabular}
}
\label{tab:entropy_data_parts}
\end{table}

\clearpage
\section{Problem Complexity}
\label{appx:complexity}

To accurately model how inference-time scaling (ITS) behaves across diverse tasks, we must understand the relationship between a model's inherent capacity and its specific failure modes. We categorize problem complexity into three distinct scenarios to better allocate computational resources and determine the most effective scaling strategy. Assuming \(p_{\theta}\) represents the language model sampler, \(N\) is the maximum parallel sampling budget, and \(n\) is the number of correct solutions generated within that budget, we classify tasks as follows:

\begin{itemize}
    \item \textbf{Level I: Solved-Easy} (\(n/N \approx k\), pass@\(N\)=1). In this regime, the correct solution resides in high-density regions of the model's probability space. Finding the answer requires minimal exploration because the model's default policy naturally aligns with the correct reasoning path. Consensus-based algorithms, such as self-consistency or majority voting, perform optimally here. Because the success rate \(k\) represents a sizable portion of the compute budget, standard parallel sampling is highly efficient without requiring intermediate interventions.

    \item \textbf{Level II: Solved-Hard} (\(n/N \ll k\), pass@\(N\)=1). Correct trajectories exist for these tasks, but they are sparse and buried in low-density regions of the probability space. While a solution is theoretically reachable, relying on blind parallel scaling is computationally inefficient and akin to finding a needle in a haystack. These cases necessitate step-level guidance, tree-search algorithms, and intermediate resampling (such as \texttt{iPF}). By evaluating partial trajectories, these methods dynamically concentrate the compute budget on high-confidence reasoning paths rather than wasting resources on dead ends.

    \item \textbf{Level III: Systematic Failure} (\(n/N \approx 0\), pass@\(N \approx 0\)). Under this condition, the model completely fails to generate correct solutions due to \textbf{faulty initial assumptions} or an inability to satisfy \textbf{multidimensional constraints}. The model's policy becomes trapped in logically flawed or hallucinated reasoning regions, rendering both parallel scaling and internal verification entirely ineffective. Overcoming systematic failure requires privileged steering via \texttt{dPF} or external verifiers to actively realign the generation process. Domain-specific rubrics, expert hints, or environmental feedback are necessary to break the model out of its incorrect prior distributions.
\end{itemize}

\clearpage
\subsection{Compute and Wall-Clock Overhead}
\label{appx:compute_overhead}

A major advantage of using intrinsic model statistics over external reward models or geometric solvers is drastically reduced verification latency. We quantify the wall-clock overhead of our intrinsic methods against a standard unguided parallel generation baseline (\texttt{pass@$N$}). Both methods rely solely on the base model's top-10 log-probabilities natively provided by vLLM. Neither requires an external reward model, verifier, or additional GPU.

As shown in Table~\ref{tab:wall_clock}, Intrinsic Selection (\texttt{iS}) adds nearly zero overhead ($< 1\%$). It computes adaptive tail entropy in microseconds using highly optimized, CPU-only vectorized operations on vLLM's existing token log-probabilities. Intrinsic Particle Filtering (\texttt{iPF}) introduces an approximately $2\times$ overhead (105\%). Crucially, this stems from the sequential nature of step-level resampling rather than an increased token budget. Both intrinsic approaches remain substantially faster than execution-based verification, which can consume up to 40\% of pipeline time in domains like engineering design.

\paragraph{Token budget equivalence} \texttt{iPF@$N$}, \texttt{dPF@$N$}, and \texttt{pass@$N$} share the same total token budget. For $N=16$ particles averaging 64 rounds of 128 tokens, the effective budget is roughly 100K tokens (accounting for early termination at an average of $\bar{n}_\text{active}=12.2$ out of 16). This matches the 96K tokens used by \texttt{pass@16}. Thus, the overhead for step-level intrinsic methods is purely architectural.

\paragraph{iPF $\approx$ pass@$2N$ in wall-clock} Because \texttt{iPF@16} requires roughly the same wall-clock time as \texttt{pass@32}, a fair practical evaluation compares \texttt{iPF@16} against \texttt{iS@32} to determine if particle filtering offers benefits beyond simply doubling the sample budget.

\begin{table}[ht!]
  \centering
  \caption{Estimated wall-clock latency per problem on math benchmarks (avg.\ response length 6000 tokens). Model: Qwen3-4B on a single H100. Timings reflect the full generation and selection loop. Token budgets for iPF@16, dPF@16, and pass@16 are matched (100K tokens). iPF overhead arises from sequential step generation, while dPF incurs additional KL scoring overhead.}
  \label{tab:wall_clock}
  \resizebox{\textwidth}{!}{%
  \begin{tabular}{l c c c l}
    \toprule
    {Method} & {Budget ($N$)} & {Avg.\ Wall-Clock} & {Overhead} & {Primary Cost} \\
    \midrule
    pass@128 (Baseline) & 128 & 268\,s & --- & Parallel token generation \\
    \texttt{iS}@128 & 128 & 268\,s & $< 1\%$ & Vectorized entropy (CPU) \\
    \midrule
    pass@32 (Baseline) & 32 & 67\,s & --- & Parallel token generation \\
    \texttt{iS}@32 & 32 & 67\,s & $< 1\%$ & Vectorized entropy (CPU) \\
    \midrule
    pass@16 (Baseline) & 16 & 33\,s & --- & Parallel token generation \\
    \texttt{iS}@16 & 16 & 33\,s & $< 1\%$ & Vectorized entropy (CPU) \\
    \texttt{iPF}@16 & 16 & 68\,s & 105\% & Sequential step generation \\
    \texttt{dPF}@16 & 16 & 85\,s & 155\% & Sequential step generation + KL scoring \\
    \bottomrule
  \end{tabular}
  }
\end{table}

\paragraph{Overhead decomposition for iPF and dPF} The 105\% overhead of \texttt{iPF@16} compared to \texttt{pass@16} decomposes into three sources, estimated from 1,890 generation rounds. The primary bottleneck is short-batch inefficiency (60\%), as vLLM optimizes for long-sequence parallel decoding, meaning 128-token batches underutilize prefill and attention kernels. Step synchronization accounts for another 30\%, since each of the 64 steps requires an API roundtrip and a resampling barrier before proceeding. Finally, KV cache management contributes 10\% due to the memory copies required for particle duplication during the 22\% of steps that trigger resampling (ESS/$N < 0.5$). 

Assuming \texttt{dPF@16} is executed entirely within vLLM and utilizes only KL-guided resampling (omitting the early logit blending phase), it inherits all of \texttt{iPF}'s sequential overheads. However, \texttt{dPF} incurs an additional 50\% relative latency penalty on top of the baseline because the guide model must perform a forward scoring pass on the generated step tokens at each resampling barrier to compute the KL divergence ($D_\text{KL}(p \parallel q)$).

\clearpage

\section{Extrinsic Reward Ablation}
\label{appx:reward_models}

In this section, we evaluate various selection strategies for choosing the final response from $N=32$ sampled solutions per problem. All evaluations are performed using \texttt{Qwen3-4B-Instruct}, with accuracy (in \%) averaged over five independent seeds ($\pm$ standard error of the mean). 

\paragraph{Baselines}
\emph{Random} selects a solution uniformly at random. \emph{Oracle} (best-of-$N$) selects a correct solution if at least one exists in the sampled set, providing an upper bound on selection performance.

\subsection{Model-Free Selection}
Model-free approaches (Table~\ref{tab:selection-free}) do not rely on an external reward model:
\begin{itemize}
    \item \emph{Entropy@1}: Selects the response with the lowest adaptive tail entropy, denoted as $H_{\text{tail}}$. The adaptive tail window is scaled by $\sqrt{\bar{L}}/\bar{H}$, incorporating percentile-based sequence length outlier trimming and median aggregation.
    \item \emph{Self-Consistency (SC)}: Applies majority voting over raw string generations without task-specific normalization.
    \item \emph{Entropy-Weighted SC (EW-SC)}: Extends SC by weighting each vote by $\exp(-H_{\text{tail}})$.
    \item \emph{Entropy-Filtered SC (E$k{\to}$SC)}: Restricts the majority vote to the top-$k$ candidates with the lowest tail entropy, explicitly filtering out highly uncertain responses prior to aggregation.
\end{itemize}

\subsection{Reward Model and Hybrid Selection}
These strategies (Table~\ref{tab:selection-orm}) leverage an external model, \texttt{Qwen2.5-Math-PRM-7B}, to score each candidate. Scores are derived from the final 2048 tokens via single-step aggregation to estimate the probability of correctness:
\begin{itemize}
    \item \emph{ORM@1}: Greedily selects the candidate with the highest ORM score.
    \item \emph{ORM-SC}: Uses the predicted ORM probabilities as weights for SC voting.
    \item \emph{Rank Fusion}: Computes the sum of the ranks derived independently from entropy and ORM scoring.
    \item \emph{Product}: Selects the response maximizing the joint heuristic $P(\text{correct}) \times \exp(-H_{\text{tail}})$.
    \item \emph{E$k{\to}$O} / \emph{O$k{\to}$E}: Filters the candidate pool to the top-$k$ responses based on lowest entropy (or highest ORM score), then selects the argmax ORM score (or argmin entropy) from the filtered subset.
    \item \emph{E$k{\to}$OSC}: First filters to the top-$k$ responses by entropy, followed by ORM-weighted majority voting on the filtered subset.
\end{itemize}

\begin{table}[ht!]
\centering
\caption{Selection accuracy (\%) ($N=32$, Qwen3-4B-Instruct). \textbf{Bold} indicates the best performance among selection methods. E$k{\to}$SC denotes entropy top-$k$ filtering followed by SC. E$k{\to}$OSC denotes entropy top-$k$ filtering followed by ORM-weighted SC. Product refers to $P(\text{correct}) \times \exp(-H_{\text{tail}})$.}
\label{tab:selection}
\footnotesize
\setlength{\tabcolsep}{2.8pt}
\resizebox{\textwidth}{!}{%
\begin{tabular}{l cc ccccc ccccc}
\toprule
 & \multicolumn{2}{c}{\textit{Bounds}} & \multicolumn{5}{c}{\textit{No external model}} & \multicolumn{5}{c}{\textit{With Qwen2.5-Math-PRM-7B}} \\
\cmidrule(lr){2-3} \cmidrule(lr){4-8} \cmidrule(lr){9-13}
 & Rand. & Oracle & Ent@1 & SC & EW-SC & E8$\to$SC & E16$\to$SC & ORM@1 & ORM-SC & E8$\to$OSC & E16$\to$OSC & Product \\
\midrule
  AIME 24 & $56.5_{\pm 0.4}$ & $87.5_{\pm 0.8}$ & $66.7_{\pm 4.9}$ & $72.5_{\pm 1.6}$ & $73.3_{\pm 0.0}$ & $75.8_{\pm 0.8}$ & $75.8_{\pm 0.8}$ & $75.0_{\pm 1.7}$ & $78.3_{\pm 2.2}$ & $79.2_{\pm 1.6}$ & $\textbf{80.0}_{\pm 1.4}$ & $75.8_{\pm 2.5}$ \\
  AIME 25 & $45.3_{\pm 0.5}$ & $75.8_{\pm 2.1}$ & $52.5_{\pm 2.1}$ & $60.8_{\pm 2.8}$ & $59.2_{\pm 2.5}$ & $54.2_{\pm 3.2}$ & $60.0_{\pm 2.4}$ & $51.7_{\pm 3.2}$ & $60.8_{\pm 2.5}$ & $55.8_{\pm 1.6}$ & $\textbf{61.7}_{\pm 2.2}$ & $55.0_{\pm 1.7}$ \\
  AIME 26 & $50.0_{\pm 0.1}$ & $83.3_{\pm 1.4}$ & $52.5_{\pm 0.8}$ & $65.8_{\pm 1.6}$ & $67.5_{\pm 2.1}$ & $60.8_{\pm 3.2}$ & $61.7_{\pm 2.2}$ & $59.2_{\pm 2.1}$ & $\textbf{68.3}_{\pm 1.0}$ & $67.5_{\pm 1.6}$ & $\textbf{68.3}_{\pm 1.0}$ & $61.7_{\pm 2.2}$ \\
  HMMT 25 & $30.3_{\pm 0.3}$ & $56.7_{\pm 3.3}$ & $34.2_{\pm 2.8}$ & $30.8_{\pm 0.8}$ & $30.8_{\pm 0.8}$ & $36.7_{\pm 1.4}$ & $35.0_{\pm 1.7}$ & $36.7_{\pm 1.4}$ & $36.7_{\pm 1.9}$ & $\textbf{39.2}_{\pm 1.6}$ & $\textbf{39.2}_{\pm 0.8}$ & $32.5_{\pm 1.6}$ \\
  HMMT 26 & $28.5_{\pm 0.4}$ & $52.3_{\pm 1.5}$ & $34.1_{\pm 0.8}$ & $28.8_{\pm 0.9}$ & $30.3_{\pm 2.1}$ & $32.6_{\pm 1.5}$ & $30.3_{\pm 1.2}$ & $34.1_{\pm 0.8}$ & $30.3_{\pm 1.2}$ & $33.3_{\pm 1.2}$ & $32.6_{\pm 0.8}$ & $\textbf{35.6}_{\pm 0.8}$ \\
\midrule
  LCB-100 & $18.9_{\pm 0.1}$ & $35.0_{\pm 0.7}$ & $\textbf{22.0}_{\pm 1.2}$ & $19.0_{\pm 1.2}$ & $19.2_{\pm 1.0}$ & $19.5_{\pm 1.7}$ & $18.2_{\pm 1.1}$ & $14.8_{\pm 0.9}$ & $18.5_{\pm 1.6}$ & $20.2_{\pm 1.0}$ & $20.0_{\pm 1.9}$ & $16.2_{\pm 1.3}$ \\
\midrule
  GPQA & $42.3_{\pm 0.1}$ & $79.3_{\pm 0.7}$ & $46.3_{\pm 0.4}$ & $43.4_{\pm 0.6}$ & $44.2_{\pm 0.9}$ & $46.7_{\pm 0.6}$ & $44.7_{\pm 0.6}$ & $47.2_{\pm 1.0}$ & $\textbf{47.9}_{\pm 1.2}$ & $46.8_{\pm 1.0}$ & $46.8_{\pm 0.7}$ & $47.7_{\pm 0.5}$ \\
\midrule
  \textit{Math} & $42.1_{\pm 0.2}$ & $71.1_{\pm 0.7}$ & $48.0_{\pm 1.5}$ & $51.8_{\pm 0.4}$ & $52.2_{\pm 0.4}$ & $52.0_{\pm 1.0}$ & $52.6_{\pm 0.6}$ & $51.3_{\pm 0.9}$ & $54.9_{\pm 0.7}$ & $55.0_{\pm 1.0}$ & $\textbf{56.3}_{\pm 0.4}$ & $52.1_{\pm 0.3}$ \\
  \textit{Code} & $18.9_{\pm 0.1}$ & $35.0_{\pm 0.7}$ & $\textbf{22.0}_{\pm 1.2}$ & $19.0_{\pm 1.2}$ & $19.2_{\pm 1.0}$ & $19.5_{\pm 1.7}$ & $18.2_{\pm 1.1}$ & $14.8_{\pm 0.9}$ & $18.5_{\pm 1.6}$ & $20.2_{\pm 1.0}$ & $20.0_{\pm 1.9}$ & $16.2_{\pm 1.3}$ \\
  \textit{Reason.} & $42.3_{\pm 0.1}$ & $79.3_{\pm 0.7}$ & $46.3_{\pm 0.4}$ & $43.4_{\pm 0.6}$ & $44.2_{\pm 0.9}$ & $46.7_{\pm 0.6}$ & $44.7_{\pm 0.6}$ & $47.2_{\pm 1.0}$ & $\textbf{47.9}_{\pm 1.2}$ & $46.8_{\pm 1.0}$ & $46.8_{\pm 0.7}$ & $47.7_{\pm 0.5}$ \\
\bottomrule
\end{tabular}
}
\end{table}

\begin{table}[ht!]
\centering
\caption{Model-free selection methods ($N=32$). E$k{\to}$SC denotes entropy top-$k$ filtering followed by majority voting. \textbf{Bold} indicates the best performance per row (excluding bounds).}
\label{tab:selection-free}
\footnotesize
\setlength{\tabcolsep}{3pt}
\resizebox{\textwidth}{!}{%
\begin{tabular}{l cccccccc}
\toprule
 & \multicolumn{2}{c}{\textit{Bounds}} & \multicolumn{3}{c}{\textit{Base}} & \multicolumn{3}{c}{\textit{Entropy-filtered SC}} \\
\cmidrule(lr){2-3} \cmidrule(lr){4-6} \cmidrule(lr){7-9}
 & Random & Oracle & Ent@1 & SC & EW-SC & E4$\to$SC & E8$\to$SC & E16$\to$SC \\
\midrule
  AIME 24 & $56.5_{\pm 0.4}$ & $87.5_{\pm 0.8}$ & $66.7_{\pm 4.9}$ & $72.5_{\pm 1.6}$ & $73.3_{\pm 0.0}$ & $71.7_{\pm 1.0}$ & $\textbf{75.8}_{\pm 0.8}$ & $\textbf{75.8}_{\pm 0.8}$ \\
  AIME 25 & $45.3_{\pm 0.5}$ & $75.8_{\pm 2.1}$ & $52.5_{\pm 2.1}$ & $\textbf{60.8}_{\pm 2.8}$ & $59.2_{\pm 2.5}$ & $52.5_{\pm 3.9}$ & $54.2_{\pm 3.2}$ & $60.0_{\pm 2.4}$ \\
  AIME 26 & $50.0_{\pm 0.1}$ & $83.3_{\pm 1.4}$ & $52.5_{\pm 0.8}$ & $65.8_{\pm 1.6}$ & $\textbf{67.5}_{\pm 2.1}$ & $56.7_{\pm 2.7}$ & $60.8_{\pm 3.2}$ & $61.7_{\pm 2.2}$ \\
  HMMT 25 & $30.3_{\pm 0.3}$ & $56.7_{\pm 3.3}$ & $34.2_{\pm 2.8}$ & $30.8_{\pm 0.8}$ & $30.8_{\pm 0.8}$ & $\textbf{37.5}_{\pm 0.8}$ & $36.7_{\pm 1.4}$ & $35.0_{\pm 1.7}$ \\
  HMMT 26 & $28.5_{\pm 0.4}$ & $52.3_{\pm 1.5}$ & $\textbf{34.1}_{\pm 0.8}$ & $28.8_{\pm 0.9}$ & $30.3_{\pm 2.1}$ & $30.3_{\pm 1.2}$ & $32.6_{\pm 1.5}$ & $30.3_{\pm 1.2}$ \\
\midrule
  LCB-100 & $18.9_{\pm 0.1}$ & $35.0_{\pm 0.7}$ & $\textbf{22.0}_{\pm 1.2}$ & $19.0_{\pm 1.2}$ & $19.2_{\pm 1.0}$ & $20.0_{\pm 1.7}$ & $19.5_{\pm 1.7}$ & $18.2_{\pm 1.1}$ \\
\midrule
  GPQA & $42.3_{\pm 0.1}$ & $79.3_{\pm 0.7}$ & $46.3_{\pm 0.4}$ & $43.4_{\pm 0.6}$ & $44.2_{\pm 0.9}$ & $46.1_{\pm 1.1}$ & $\textbf{46.7}_{\pm 0.6}$ & $44.7_{\pm 0.6}$ \\
\midrule
  \textit{Math} & $42.1_{\pm 0.2}$ & $71.1_{\pm 0.7}$ & $48.0_{\pm 1.5}$ & $51.8_{\pm 0.4}$ & $52.2_{\pm 0.4}$ & $49.7_{\pm 0.8}$ & $52.0_{\pm 1.0}$ & $\textbf{52.6}_{\pm 0.6}$ \\
  \textit{Code} & $18.9_{\pm 0.1}$ & $35.0_{\pm 0.7}$ & $\textbf{22.0}_{\pm 1.2}$ & $19.0_{\pm 1.2}$ & $19.2_{\pm 1.0}$ & $20.0_{\pm 1.7}$ & $19.5_{\pm 1.7}$ & $18.2_{\pm 1.1}$ \\
  \textit{Reason.} & $42.3_{\pm 0.1}$ & $79.3_{\pm 0.7}$ & $46.3_{\pm 0.4}$ & $43.4_{\pm 0.6}$ & $44.2_{\pm 0.9}$ & $46.1_{\pm 1.1}$ & $\textbf{46.7}_{\pm 0.6}$ & $44.7_{\pm 0.6}$ \\
\bottomrule
\end{tabular}
}
\end{table}

\begin{table}[ht!]
\centering
\caption{ORM-based selection methods ($N=32$, \texttt{Qwen2.5-Math-PRM-7B}). \textbf{Bold} indicates the best performance per row.}
\label{tab:selection-orm}
\footnotesize
\setlength{\tabcolsep}{2.5pt}
\resizebox{\textwidth}{!}{%
\begin{tabular}{l ccccccccccccc}
\toprule
 & \multicolumn{2}{c}{\textit{Base}} & \multicolumn{2}{c}{\textit{Simple comb.}} & \multicolumn{3}{c}{\textit{E$k{\to}$ORM}} & \multicolumn{3}{c}{\textit{O$k{\to}$Ent}} & \multicolumn{3}{c}{\textit{E$k{\to}$ORM-SC}} \\
\cmidrule(lr){2-3} \cmidrule(lr){4-5} \cmidrule(lr){6-8} \cmidrule(lr){9-11} \cmidrule(lr){12-14}
 & ORM@1 & ORM-SC & Rank fus. & Product & E4$\to$O & E8$\to$O & E16$\to$O & O4$\to$E & O8$\to$E & O16$\to$E & E4$\to$OSC & E8$\to$OSC & E16$\to$OSC \\
\midrule
  AIME 24 & $75.0_{\pm 1.7}$ & $78.3_{\pm 2.2}$ & $75.0_{\pm 1.7}$ & $75.8_{\pm 2.5}$ & $70.8_{\pm 1.6}$ & $74.2_{\pm 2.1}$ & $75.0_{\pm 2.2}$ & $78.3_{\pm 1.0}$ & $74.2_{\pm 1.6}$ & $70.0_{\pm 1.4}$ & $71.7_{\pm 1.7}$ & $79.2_{\pm 1.6}$ & $\textbf{80.0}_{\pm 1.4}$ \\
  AIME 25 & $51.7_{\pm 3.2}$ & $60.8_{\pm 2.5}$ & $54.2_{\pm 0.8}$ & $55.0_{\pm 1.7}$ & $51.7_{\pm 3.2}$ & $50.8_{\pm 1.6}$ & $56.7_{\pm 2.7}$ & $53.3_{\pm 3.0}$ & $52.5_{\pm 1.6}$ & $50.0_{\pm 1.4}$ & $54.2_{\pm 2.1}$ & $55.8_{\pm 1.6}$ & $\textbf{61.7}_{\pm 2.2}$ \\
  AIME 26 & $59.2_{\pm 2.1}$ & $\textbf{68.3}_{\pm 1.0}$ & $60.8_{\pm 3.7}$ & $61.7_{\pm 2.2}$ & $58.3_{\pm 2.2}$ & $59.2_{\pm 0.8}$ & $61.7_{\pm 2.9}$ & $61.7_{\pm 4.0}$ & $64.2_{\pm 3.9}$ & $57.5_{\pm 2.8}$ & $59.2_{\pm 2.1}$ & $67.5_{\pm 1.6}$ & $\textbf{68.3}_{\pm 1.0}$ \\
  HMMT 25 & $36.7_{\pm 1.4}$ & $36.7_{\pm 1.9}$ & $34.2_{\pm 3.4}$ & $32.5_{\pm 1.6}$ & $35.0_{\pm 2.2}$ & $34.2_{\pm 0.8}$ & $35.0_{\pm 1.0}$ & $35.8_{\pm 2.5}$ & $35.0_{\pm 2.9}$ & $32.5_{\pm 2.8}$ & $35.8_{\pm 2.5}$ & $\textbf{39.2}_{\pm 1.6}$ & $\textbf{39.2}_{\pm 0.8}$ \\
  HMMT 26 & $34.1_{\pm 0.8}$ & $30.3_{\pm 1.2}$ & $32.6_{\pm 0.8}$ & $\textbf{35.6}_{\pm 0.8}$ & $32.6_{\pm 1.5}$ & $33.3_{\pm 0.0}$ & $34.1_{\pm 0.8}$ & $31.8_{\pm 0.9}$ & $31.1_{\pm 0.8}$ & $33.3_{\pm 1.2}$ & $31.1_{\pm 1.5}$ & $33.3_{\pm 1.2}$ & $32.6_{\pm 0.8}$ \\
\midrule
  LCB-100 & $14.8_{\pm 0.9}$ & $18.5_{\pm 1.6}$ & $18.8_{\pm 1.3}$ & $16.2_{\pm 1.3}$ & $20.0_{\pm 1.8}$ & $20.0_{\pm 1.2}$ & $19.2_{\pm 1.8}$ & $15.2_{\pm 0.6}$ & $17.8_{\pm 1.3}$ & $20.0_{\pm 1.1}$ & $\textbf{20.5}_{\pm 1.8}$ & $20.2_{\pm 1.0}$ & $20.0_{\pm 1.9}$ \\
\midrule
  GPQA & $47.2_{\pm 1.0}$ & $47.9_{\pm 1.2}$ & $48.4_{\pm 0.9}$ & $47.7_{\pm 0.5}$ & $48.2_{\pm 0.4}$ & $47.5_{\pm 0.7}$ & $47.5_{\pm 0.9}$ & $47.7_{\pm 1.3}$ & $47.0_{\pm 0.8}$ & $\textbf{48.5}_{\pm 0.5}$ & $47.1_{\pm 0.3}$ & $46.8_{\pm 1.0}$ & $46.8_{\pm 0.7}$ \\
\midrule
  \textit{Math} & $51.3_{\pm 0.9}$ & $54.9_{\pm 0.7}$ & $51.3_{\pm 0.7}$ & $52.1_{\pm 0.3}$ & $49.7_{\pm 1.1}$ & $50.3_{\pm 0.6}$ & $52.5_{\pm 0.9}$ & $52.2_{\pm 0.8}$ & $51.4_{\pm 0.5}$ & $48.7_{\pm 0.7}$ & $50.4_{\pm 0.7}$ & $55.0_{\pm 1.0}$ & $\textbf{56.3}_{\pm 0.4}$ \\
  \textit{Code} & $14.8_{\pm 0.9}$ & $18.5_{\pm 1.6}$ & $18.8_{\pm 1.3}$ & $16.2_{\pm 1.3}$ & $20.0_{\pm 1.8}$ & $20.0_{\pm 1.2}$ & $19.2_{\pm 1.8}$ & $15.2_{\pm 0.6}$ & $17.8_{\pm 1.3}$ & $20.0_{\pm 1.1}$ & $\textbf{20.5}_{\pm 1.8}$ & $20.2_{\pm 1.0}$ & $20.0_{\pm 1.9}$ \\
  \textit{Reason.} & $47.2_{\pm 1.0}$ & $47.9_{\pm 1.2}$ & $48.4_{\pm 0.9}$ & $47.7_{\pm 0.5}$ & $48.2_{\pm 0.4}$ & $47.5_{\pm 0.7}$ & $47.5_{\pm 0.9}$ & $47.7_{\pm 1.3}$ & $47.0_{\pm 0.8}$ & $\textbf{48.5}_{\pm 0.5}$ & $47.1_{\pm 0.3}$ & $46.8_{\pm 1.0}$ & $46.8_{\pm 0.7}$ \\
\bottomrule
\end{tabular}
}
\end{table}

The results across the evaluated selection strategies reveal three key takeaways. First, cascaded approaches that integrate both intrinsic signals and external reward models - specifically, entropy-filtered ORM-weighted Self-Consistency ($E_k{\to}$OSC) - consistently achieve the highest accuracy across most benchmarks, demonstrating the complementary strengths of internal model confidence and external verification. Second, in the absence of an external reward model, using intrinsic tail entropy to filter out highly uncertain responses prior to aggregation ($E_k{\to}$SC) generally outperforms raw Self-Consistency (SC), indicating that pruning noisy, low-confidence reasoning traces cleans up the consensus signal. Finally, greedily selecting the highest-scored response from a reward model (ORM@1) is frequently suboptimal; instead, utilizing ORM probabilities as weights for majority voting (ORM-SC) yields noticeably more robust performance across both mathematical and reasoning domains.

\clearpage

\section{Gemma Ablation}
\label{appx:gemma}

In our evaluation of selection strategies, entropy-based methods demonstrate significant performance improvements over standard self-consistency (SC) on complex reasoning tasks (where pass@1 \(<30\%\)) and code generation benchmarks. This advantage arises because SC struggles to find exact structural matches for code or lengthy reasoning chains at higher sampling temperatures. Conversely, SC remains the most robust choice for easier datasets, such as AIME-2024, where models naturally converge on the correct solution. Furthermore, our results indicate that adaptive tail formulas transfer effectively across diverse model architectures. 

Practitioners should leverage entropy-based aggregation for high-complexity or open-ended generation tasks, while defaulting to standard self-consistency for highly constrained, convergent problem-solving.

\begin{table}[ht!]
\centering
\caption{Selection accuracy (\%) for \texttt{Qwen3-4B-Instruct} across mathematics, reasoning, and code generation benchmarks (\(N{=}32\)). The \texttt{iS}@1 metric evaluates both median and mean aggregations. The best-performing selection method per aggregated column is \textbf{bolded} (excluding the pass@32 oracle).}
\label{tab:qwen-results}
\resizebox{\textwidth}{!}{%
\begin{tabular}{l ccccccc ccc}
\toprule
& \multicolumn{7}{c}{Dataset} & \multicolumn{3}{c}{Aggregated} \\
\cmidrule(lr){2-8} \cmidrule(lr){9-11}
{Method} & {A24} & {A25} & {A26} & {H25} & {H26} & {GPQA} & {LCB} & {Mean} & {Median} & {w Mean} \\
\midrule
pass@1             & 56.5 & 45.3 & 50.0 & 30.2 & 28.5 & 42.3 & 19.0 & 38.8 & 42.3 & 37.0 \\
pass@32            & 87.5 & 75.8 & 83.3 & 54.2 & 52.3 & 79.3 & 35.0 & 66.8 & 75.8 & 66.4 \\
\midrule
SC@32              & \textbf{72.5} & \textbf{60.8} & \textbf{65.8} & 30.8 & 28.8 & 44.1 & 18.2 & \textbf{45.9} & 44.1 & 40.8 \\
\texttt{iS}@1 (median)      & 66.7 & 52.5 & 52.5 & \textbf{34.2} & \textbf{34.1} & \textbf{46.3} & \textbf{22.0} & 44.0 & \textbf{46.3} & \textbf{41.4} \\
\texttt{iS}@1 (mean)        & 66.7 & 45.0 & 53.3 & 30.0 & 31.8 & 46.6 & 21.0 & 42.1 & 45.0 & 40.4 \\
\bottomrule
\end{tabular}%
}
\end{table}

\begin{table}[ht!]
\centering
\caption{Selection accuracy (\%) for \texttt{gemma-4-E2B-it} evaluated on identical benchmarks (\(N{=}32\)). Method configurations and tail hyperparameters mirror those detailed in Table~\ref{tab:qwen-results}. The best-performing selection method per aggregated column is \textbf{bolded} (excluding the pass@32 oracle).}
\label{tab:gemma-results}
\resizebox{\textwidth}{!}{%
\begin{tabular}{l ccccccc ccc}
\toprule
& \multicolumn{7}{c}{Dataset} & \multicolumn{3}{c}{Aggregated} \\
\cmidrule(lr){2-8} \cmidrule(lr){9-11}
{Method} & {A24} & {A25} & {A26} & {H25} & {H26} & {GPQA} & {LCB} & {Mean} & {Median} & {w Mean} \\
\midrule
pass@1             & 44.8 & 29.3 & 39.6 & 19.2 & 21.6 & 40.4 & 19.6 & 30.6 & 29.3 & 32.5 \\
pass@32            & 76.7 & 63.3 & 63.3 & 53.3 & 45.5 & 84.3 & 36.0 & 60.3 & 63.3 & 65.3 \\
\midrule
SC@32              & 60.0 & 36.7 & 50.0 & 20.0 & \textbf{33.3} & 45.5 & 20.0 & 37.9 & \textbf{36.7} & 37.9 \\
\texttt{iS}@1 (median)      & 60.0 & 36.7 & \textbf{56.7} & 20.0 & 21.2 & 46.0 & \textbf{25.0} & 37.9 & \textbf{36.7} & 38.8 \\
\texttt{iS}@1 (mean)        & 60.0 & 36.7 & 53.3 & \textbf{23.3} & 24.2 & \textbf{47.5} & 24.0 & \textbf{38.4} & \textbf{36.7} & \textbf{39.4} \\
\bottomrule
\end{tabular}%
}
\end{table}

\clearpage
\section{Additional Ablations}
\label{appx:additional_experiments}

\begin{figure}[htbp]
    \centering
    \begin{subfigure}[b]{0.32\textwidth}
        \centering
        \includegraphics[width=\textwidth]{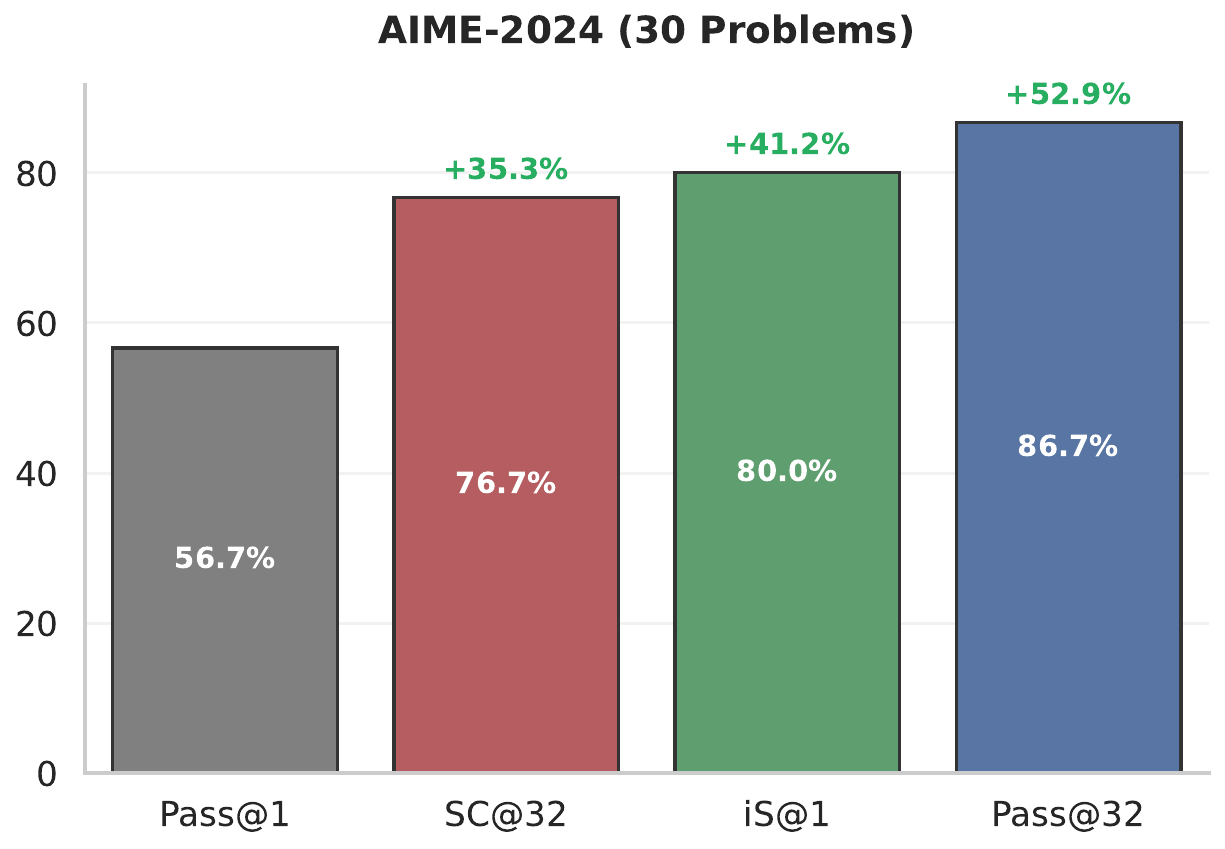}
        \caption{AIME-2024}
        \label{fig:a24-intro-appx}
    \end{subfigure}
    \hfill %
    \begin{subfigure}[b]{0.32\textwidth}
        \centering
        \includegraphics[width=\textwidth]{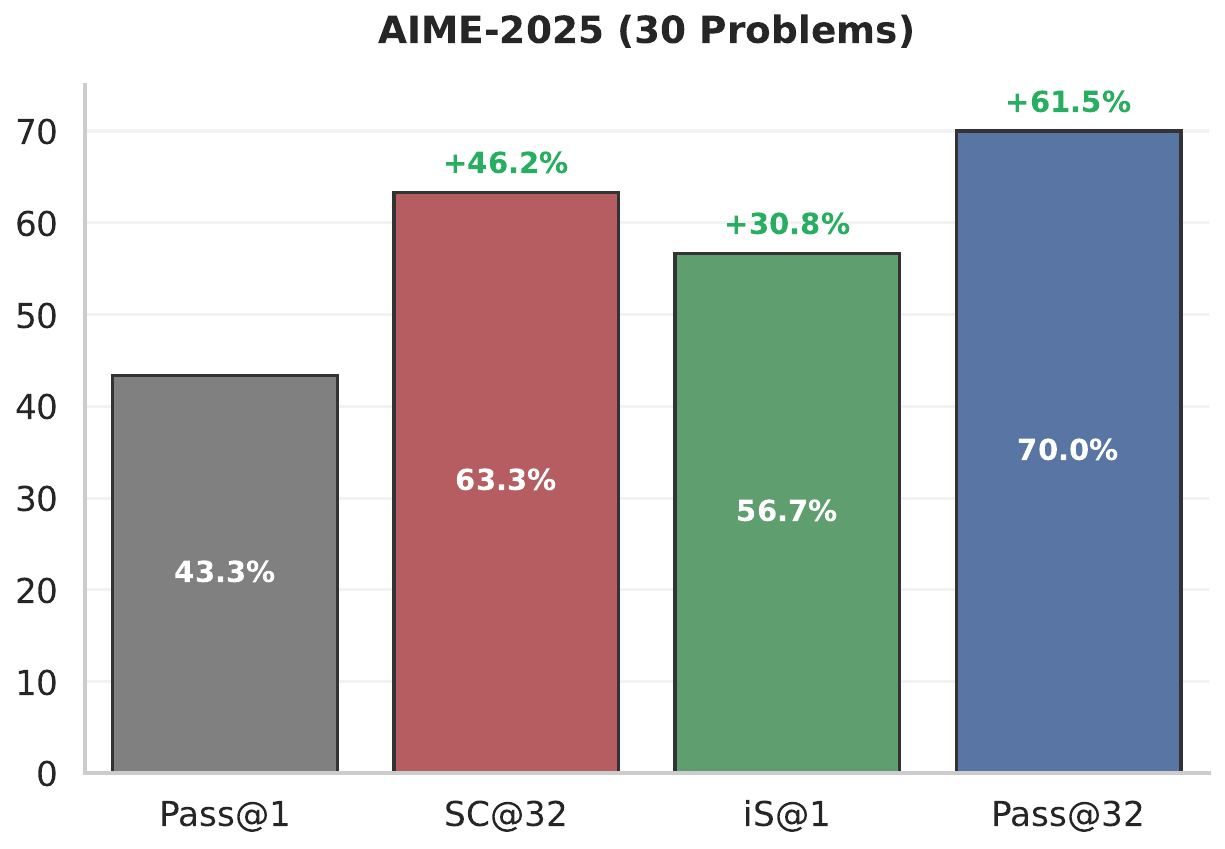}
        \caption{AIME-2025}
        \label{fig:a25-intro-appx}
    \end{subfigure}
    \hfill %
    \begin{subfigure}[b]{0.32\textwidth}
        \centering
        \includegraphics[width=\textwidth]{img/selection/aime-2026_relative_improvement_paper.pdf}
        \caption{AIME-2026}
        \label{fig:a26-intro-appx}
    \end{subfigure}
    
    \vspace{20pt}
    
    \begin{subfigure}[b]{0.24\textwidth}
        \centering
        \includegraphics[width=\textwidth]{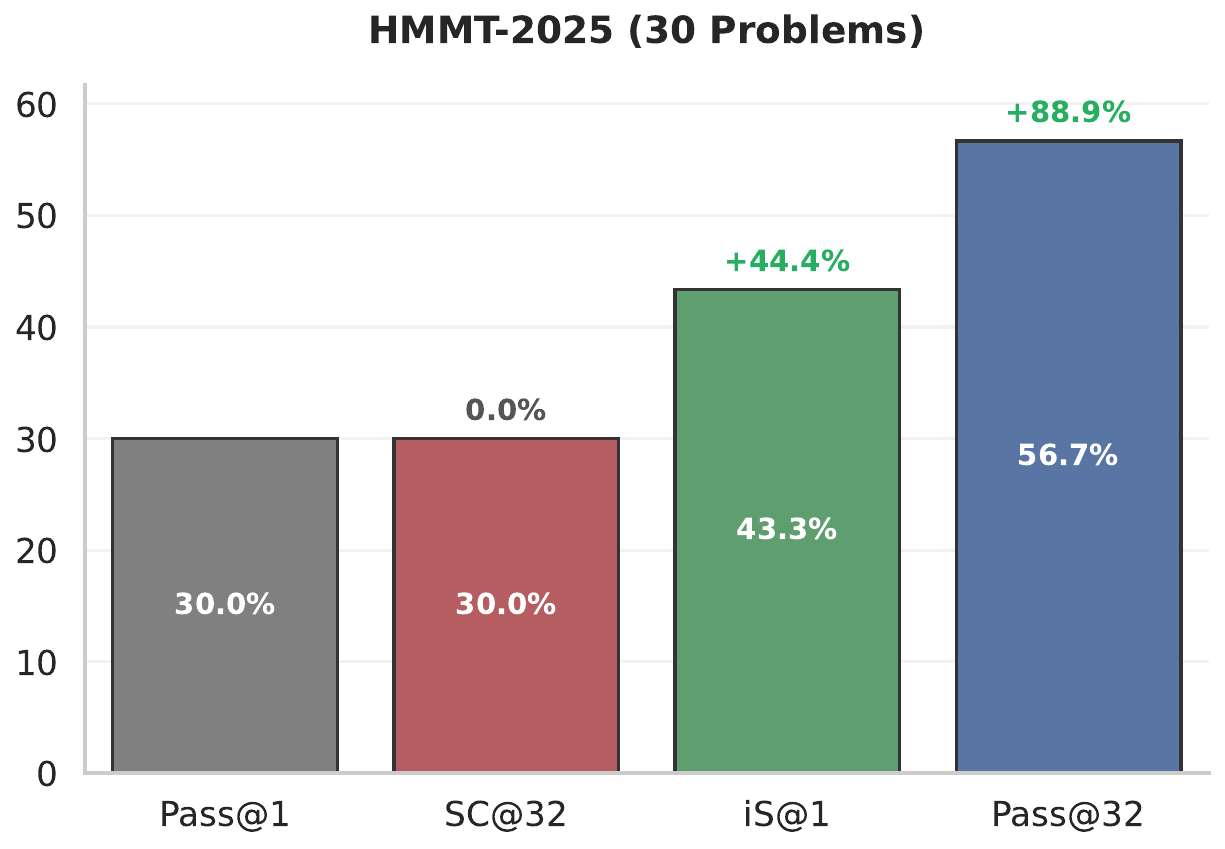}
        \caption{HMMT-2025}
        \label{fig:h25-intro-appx}
    \end{subfigure}
    \hfill %
    \begin{subfigure}[b]{0.24\textwidth}
        \centering
        \includegraphics[width=\textwidth]{img/selection/gpqa_relative_improvement_paper.pdf}
        \caption{GPQA-Diamond}
        \label{fig:gpqa-intro-appx}
    \end{subfigure}
    \hfill %
    \begin{subfigure}[b]{0.24\textwidth}
        \centering
        \includegraphics[width=\textwidth]{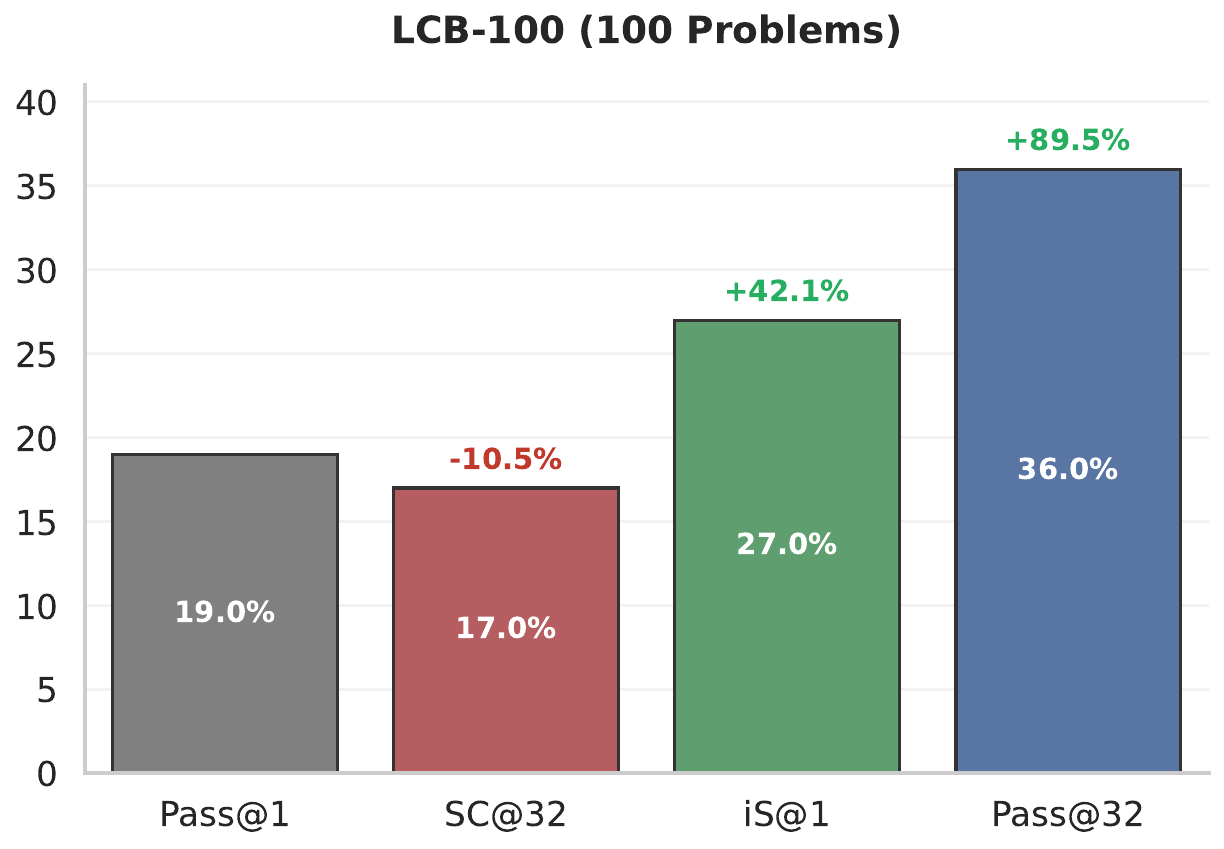}
        \caption{LiveCodeBench-v6}
        \label{fig:lcb-intro-appx}
    \end{subfigure}
    \hfill %
    \begin{subfigure}[b]{0.24\textwidth}
        \centering
        \includegraphics[width=\textwidth]{img/selection/hmmt-2026_relative_improvement_paper.pdf}
        \caption{HMMT-2026}
        \label{fig:h26-intro-appx}
    \end{subfigure}
    
    \caption{Accuracy pass@1, Self-Consistency, \texttt{iS}@1, and pass@N. We provide validation results for our method evaluated over complex math datasets, general reasoning, and coding. The domains require different level or verification.}
    \label{fig:global-appx}
\end{figure}

\clearpage
\subsection{Budget}
\label{appx:budget}

\begin{table}[ht!]
\centering
\caption{Intrinsic selection accuracy (\%) across sample budgets --- \textbf{mean} of 10 bootstrap trials. Pool of 512 samples per problem. Model: Qwen3-4B-Instruct-2507. Best selection method per dataset in \textbf{bold} (excluding pass@$N$).}
\label{tab:phase1-mean-appx}
\resizebox{\textwidth}{!}{%
\begin{tabular}{cl cccccccc c}
\toprule
$N$ & Method & AIME 2024 & AIME 2025 & AIME 2026 & HMMT 2025 & HMMT 2026 & GPQA-D & LCBv6-100 &&  Wt.\ Avg \\
\midrule
\multirow{4}{*}{8}
 & pass@1      & 56.5 & 45.4 & 50.0 & 30.1 & 28.6 & 42.2 & 19.0 && 37.0\\
 & \texttt{iS}@1 (ours) & \textbf{67.6} & 47.5 & 55.7 & \textbf{33.8} & \textbf{31.6} & \textbf{44.9} & \textbf{22.4} && \textbf{40.6} \\
 & SC          & 67.6 & \textbf{56.1} & \textbf{60.1} & 32.1 & 28.9 & 42.6 & 18.9 && 39.4\\
 & pass@$N$    & 81.8 & 66.8 & 75.2 & 46.5 & 42.5 & 69.2 & 29.9 && 58.1 \\
\midrule
\multirow{4}{*}{16}
 & pass@1      & 56.6 & 45.4 & 49.9 & 30.2 & 28.5 & 42.3 & 19.0 && 37.0\\
 & \texttt{iS}@1 (ours) & 68.8 & 47.0 & 55.2 & \textbf{34.5} & \textbf{31.7} & \textbf{45.5} & \textbf{22.4} && \textbf{40.9} \\
 & SC          & \textbf{70.1} & \textbf{58.2} & \textbf{61.9} & 32.0 & 28.1 & 42.5 & 18.3 && 39.5 \\
 & pass@$N$    & 85.3 & 71.3 & 79.3 & 50.9 & 47.9 & 74.7 & 32.3 && 62.6 \\
\midrule
\multirow{4}{*}{32}
 & pass@1      & 56.5 & 45.3 & 50.0 & 30.1 & 28.5 & 42.3 & 18.9 && 36.9 \\
 & \texttt{iS}@1 (ours) & 68.7 & 46.4 & 54.7 & \textbf{35.4} & \textbf{32.0} & \textbf{45.6} & \textbf{21.8} && \textbf{40.8} \\
 & SC          & \textbf{71.1} & \textbf{58.8} & \textbf{63.1} & 31.6 & 28.3 & 42.6 & 17.6 && 39.6 \\
 & pass@$N$    & 87.7 & 75.4 & 82.2 & 55.0 & 53.3 & 79.4 & 34.7 && 66.4 \\
\midrule
\multirow{4}{*}{64}
 & pass@1      & 56.5 & 45.3 & 50.0 & 30.2 & 28.5 & 42.3 & 18.9 && 36.9 \\
 & \texttt{iS}@1 (ours) & 69.5 & 46.0 & 54.4 & \textbf{36.8} & \textbf{31.5} & \textbf{45.5} & \textbf{21.1} && \textbf{40.7} \\
 & SC          & \textbf{71.9} & \textbf{59.9} & \textbf{64.1} & 30.8 & 27.8 & 42.7 & 16.8 && 39.6 \\
 & pass@$N$    & 89.1 & 79.0 & 85.2 & 60.7 & 59.5 & 83.4 & 37.2 && 70.1 \\
\midrule
\multirow{4}{*}{128}
 & pass@1      & 56.5 & 45.3 & 50.0 & 30.2 & 28.5 & 42.3 & 19.0 && 37.0 \\
 & \texttt{iS}@1 (ours) & 70.0 & 46.7 & 56.7 & \textbf{40.0} & \textbf{33.3} & \textbf{44.9} & \textbf{20.0} && \textbf{40.8} \\
 & SC          & \textbf{73.3} & \textbf{60.0} & \textbf{63.3} & 30.0 & 27.3 & 41.9 & 16.0 && 39.0 \\
 & pass@$N$    & 90.0 & 80.0 & 86.7 & 70.0 & 63.6 & 86.9 & 39.0 && 73.2 \\
\bottomrule
\end{tabular}%
}
\end{table}

\begin{table}[ht!]
\centering
\caption{Intrinsic selection accuracy (\%) across sample budgets --- \textbf{median} of 10 bootstrap trials. Pool of 512 samples per problem. Model: Qwen3-4B-Instruct-2507. Best selection method per dataset in \textbf{bold} (excluding pass@$N$).}
\label{tab:phase1-median-appx}
\resizebox{\textwidth}{!}{%
\begin{tabular}{cl cccccccc c}
\toprule
$N$ & Method& AIME 2024 & AIME 2025 & AIME 2026 & HMMT 2025 & HMMT 2026 & GPQA-D & LCBv6-100 && Wt.\ Avg \\
\midrule
\multirow{4}{*}{8}
 & pass@1      & 56.7 & 45.4 & 50.0 & 30.0 & 28.4 & 42.2 & 19.0 && 36.9 \\
 & \texttt{iS}@1 (ours) & 66.7 & 46.7 & \textbf{56.7} & \textbf{33.3} & \textbf{30.3} & \textbf{44.9} & \textbf{22.0} && \textbf{40.4} \\
 & SC          & \textbf{66.7} & \textbf{56.7} & 60.0 & 33.3 & 27.3 & 42.4 & 19.0 && 39.2 \\
 & pass@$N$    & 83.3 & 66.7 & 76.7 & 46.7 & 42.4 & 69.2 & 30.0 && 58.3 \\
\midrule
\multirow{4}{*}{16}
 & pass@1      & 56.5 & 45.4 & 50.0 & 30.2 & 28.4 & 42.3 & 18.9 && 36.9 \\
 & \texttt{iS}@1 (ours) & \textbf{70.0} & 46.7 & 56.7 & \textbf{33.3} & \textbf{30.3} & \textbf{45.5} & \textbf{23.0} && \textbf{41.0} \\
 & SC          & 70.0 & \textbf{56.7} & \textbf{63.3} & 31.7 & 27.3 & 42.4 & 18.0 && 39.4 \\
 & pass@$N$    & 86.7 & 70.0 & 80.0 & 50.0 & 48.5 & 74.7 & 32.0 && 62.5 \\
\midrule
\multirow{4}{*}{32}
 & pass@1      & 56.5 & 45.3 & 49.9 & 30.1 & 28.5 & 42.3 & 18.9 && 36.9 \\
 & \texttt{iS}@1 (ours) & \textbf{70.0} & 46.7 & 53.3 & \textbf{36.7} & \textbf{30.3} & \textbf{45.5} & \textbf{22.0} && \textbf{40.8} \\
 & SC          & 70.0 & \textbf{60.0} & \textbf{63.3} & 30.0 & 27.3 & 42.4 & 18.0 && 39.5 \\
 & pass@$N$    & 86.7 & 76.7 & 83.3 & 53.3 & 54.5 & 79.3 & 35.0 && 66.5 \\
\midrule
\multirow{4}{*}{64}
 & pass@1      & 56.5 & 45.3 & 49.9 & 30.2 & 28.5 & 42.3 & 19.0 && 37.0 \\
 & \texttt{iS}@1 (ours) & 70.0 & 46.7 & 53.3 & \textbf{36.7} & \textbf{30.3} & \textbf{45.5} & \textbf{21.0} && \textbf{40.6} \\
 & SC          & \textbf{73.3} & \textbf{60.0} & \textbf{63.3} & 30.0 & 27.3 & 42.4 & 17.0 && 39.5 \\
 & pass@$N$    & 90.0 & 80.0 & 86.7 & 60.0 & 60.6 & 83.3 & 37.0 && 70.3 \\
\midrule
\multirow{4}{*}{128}
 & pass@1      & 56.5 & 45.3 & 50.0 & 30.2 & 28.5 & 42.3 & 19.0 && 37.0 \\
 & \texttt{iS}@1 (ours) & 70.0 & 46.7 & 56.7 & \textbf{40.0} & \textbf{33.3} & \textbf{44.9} & \textbf{20.0} && \textbf{40.8} \\
 & SC          & \textbf{73.3} & \textbf{60.0} & \textbf{63.3} & 30.0 & 27.3 & 41.9 & 16.0 && 39.0 \\
 & pass@$N$    & 90.0 & 80.0 & 86.7 & 70.0 & 63.6 & 86.9 & 39.0 && 73.2 \\
\bottomrule
\end{tabular}%
}
\end{table}

\clearpage
\subsection{Consensus and Confidence}
\label{appx:consensus_confidence}

To further contextualize our findings, Table~\ref{tab:deepconf} compares validation accuracy across math datasets for Self-Consistency, \texttt{iS}, and state-of-the-art confidence-based selection methods, notably Self-Certainty~\citep{kang2025scalable} and DeepConf~\citep{fu2025deep} ($N = 32$, max length 16,000 tokens). These baselines represent different paradigms for inference-time scaling:
\begin{itemize}
    \item {Self-Consistency:} Relies purely on output consensus by marginalizing reasoning trajectories via majority voting over a canonical answer format.
    \item {Self-Certainty:} Depends entirely on intrinsic confidence, measuring the divergence from a uniform distribution to the model's predictive distribution to bypass external verifiers.
    \item {DeepConf:} Hybridizes the two by weighting majority voting consensus with trace-level confidence scores derived from token log probabilities.
    \item {\texttt{iS} (Ours):} Evaluates solution quality intrinsically by ranking candidates using set-level length distributions and adjusted per-token tail entropy, entirely bypassing external verifiers and consensus matching.
\end{itemize}

\begin{table}[ht!]
  \centering
  \caption{Validation accuracy across math datasets using various baselines involving confidence-based, consensus-based, and refining-based inference scaling. DeepConf merges consensus and confidence metrics for selection. For the metrics involving confidence, we use the selection variant performing the best over each dataset.}
  \resizebox{\textwidth}{!}{%
  \begin{tabular}{lccccc}
  \toprule
  Method & AIME 2024 & AIME 2025 & AIME 2026 & HMMT 2025 & HMMT 2026\\
  \midrule
  Self-Consistency~\citep{wang2022self} & $72.5_{\pm 1.5}$ & $60.8_{\pm 2.8}$ & $65.8_{\pm 1.6}$ & $30.8_{\pm 0.8}$ & $28.5_{\pm 0.4}$ \\
  Self-Certainty~\citep{kang2025scalable} (val)   &  $67.5_{\pm 2.2}$ & $50.0_{\pm 0.4}$ & $51.7_{\pm 0.4}$ & $35.0_{\pm 1.1}$ & $34.8_{\pm 1.9}$ \\
  DeepConf~\citep{fu2025deep} (val) & $72.7_{\pm 1.3}$ & $62.7_{\pm 2.5}$ & $70.7_{\pm 1.3}$ & $30.7_{\pm 2.5}$ & $25.5_{\pm 4.5}$ \\
  \texttt{iS}@1 (our, val) & $74.2_{\pm 1.7}$ & $52.5_{\pm 2.1}$ & $64.2_{\pm 1.5}$ & $38.3_{\pm 0.8}$ & $37.1_{\pm 1.1}$\\
  \bottomrule
  \end{tabular}
  }
  \label{tab:deepconf}
\end{table}

The empirical results highlight the complementary strengths of consensus- and confidence-based approaches, as well as the unique advantages of \texttt{iS}. DeepConf performs exceptionally well on the AIME datasets, where the model successfully produces a sizable proportion of correct solutions and majority voting is already highly effective. By leveraging trace-level confidence to upweight reliable consensus clusters, DeepConf noticeably improves upon the baseline, notably reaching 70.7\% on AIME 2026 compared to the 65.8\% Self-Consistency baseline.

However, this reliance on consensus becomes a bottleneck on harder datasets like HMMT. On HMMT datasets, where the frequency of correct answers within the sample set drops significantly ($n/N \ll 1$), extracting a reliable consensus frequently fails. In these sparse regimes, pure consensus degrades, and hybrid methods like DeepConf struggle to compensate because the underlying answer clusters are too weak.

Conversely, \texttt{iS} achieves the highest overall accuracy on the HMMT datasets (38.3\% and 37.1\%). By selecting the candidate with the lowest adjusted tail entropy, \texttt{iS} efficiently isolates the best solution even when correct trajectories are sparse and lack a strong consensus cluster. Crucially, \texttt{iS} achieves these gains without requiring any output parsing, canonical-form matching, or explicit answer clustering. This lack of dependency on extrinsic output formats means that, unlike Self-Consistency or DeepConf, \texttt{iS} can seamlessly extend to open-ended and hard-to-verify domains.

\clearpage

\subsection{Selection without the Answer}
\label{appx:answer}

A fundamental limitation of consensus-based algorithms, such as Self-Consistency, is their strict reliance on a clearly defined, extractable final answer. These methods must parse a canonical output format (e.g., a \texttt{\textbackslash boxed\{\}} string in mathematics) to group identical solutions and compute majority votes. This dependency effectively precludes their use in open-ended domains - such as clinical drafting, creative writing, or complex engineering design - where solutions are highly variable and no single definitive ``answer'' string exists. 

To demonstrate that Intrinsic Selection (\texttt{iS}) completely bypasses this limitation, we designed an experiment to evaluate the robustness of our entropy-based scoring when the final answer is entirely inaccessible. Specifically, we stripped the \texttt{\textbackslash boxed\{\}} answer region from every generated sequence prior to evaluation, forcing the \texttt{iS} scoring function to operate solely on the intrinsic statistics of the preceding reasoning trace.

\begin{table}[ht!]
  \caption{Performance comparison without and with answer stripping using Intrinsic Selection. Overall, our method is highly robust to answer presence and even sees marginal gains when the final answer is removed.}
  \label{tab:answer-stripping-appx}
  \centering
  \begin{tabular}{lccccccc}
    \toprule
    Method & A24 & A25 & A26 & H25 & H26 & GPQA & wt avg \\
    \midrule
    w/o strip   & 67.5 & 49.2 & 52.5 & 35.0 & 31.1 & 46.5 & 46.96 \\
    w/  strip & 67.5 & 50.0 & 55.0 & 35.8 & 32.6 & 47.6 & 48.08 \\
    \midrule
    $\Delta$ & +0.0 & +0.8 & +2.5 & +0.8 & +1.5 & +1.1 &  +1.06\\
    \bottomrule
  \end{tabular}
\end{table}

\begin{figure}[ht!]
    \centering
    \includegraphics[width=.8\linewidth]{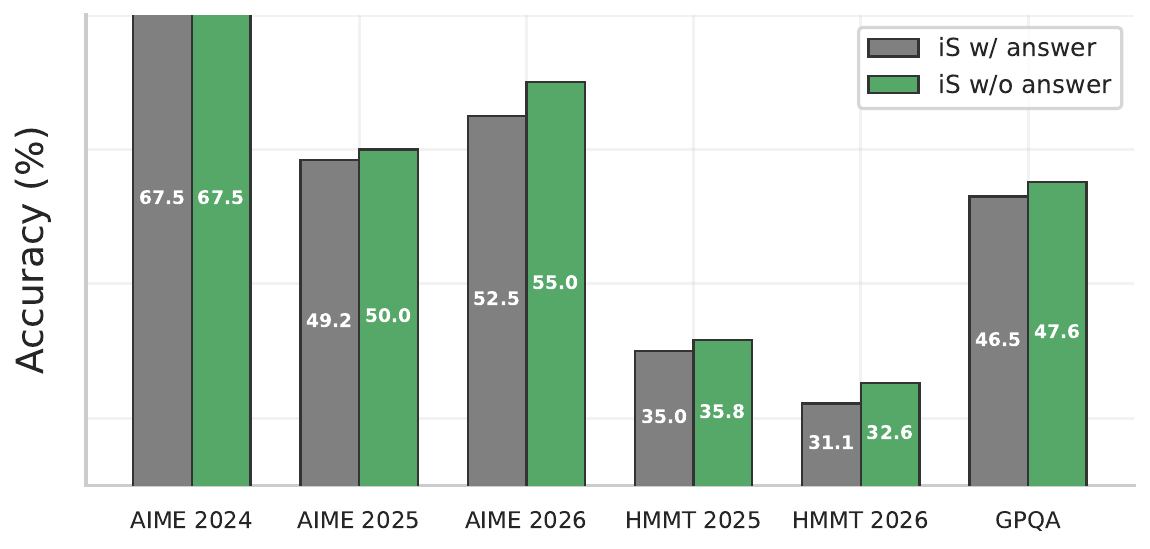}
    \caption{Answer Stripping. \texttt{iS} remains robust without the final answer, demonstrating its effectiveness regardless of domain verifiability.}
    \label{fig:answer-strip}
\end{figure}

As shown in Figure~\ref{fig:answer-strip} and detailed in Table~\ref{tab:answer-stripping-appx}, the performance of \texttt{iS} remains stable or slightly improves across all six math and reasoning datasets when the answer is stripped. We observe consistent positive deltas, ranging from neutral (+0.0 on AIME 2024) to noticeable gains (+2.5 on AIME 2026), yielding an average dataset-size weighted gain of +1.06 percentage points overall. 

This marginal improvement suggests that formatting tokens and standardized answer wrappers may occasionally introduce artificial entropy spikes that add noise to confidence estimations. By evaluating only the reasoning steps, \texttt{iS} captures a purer signal of the model's internal certainty. Most importantly, these results empirically confirm that the discriminative signal captured by adjusted tail entropy resides entirely in the latent structure of the reasoning trajectory rather than the final answer tokens. Consequently, \texttt{iS} functions effectively as a verifier-free inference scaling method applicable even in open-ended domains lacking canonical outputs.

\clearpage
\subsection{Entropy Tail and Aggregation Ablations}
\label{appx:tail}

The core premise of Intrinsic Selection (\texttt{iS}) is that not all tokens in a reasoning trajectory are equally informative for assessing confidence. Early steps in a chain-of-thought sequence often contain exploratory or boilerplate text, while the final tokens encapsulate the model's conclusive stance. However, the length of the relevant "answer region" varies significantly across problems and domains. To validate our design choices for isolating this discriminative signal, we conduct ablation studies on both the tail window definition and the token-level aggregation function.

\begin{table}[ht!]
\centering
\caption{Ablation: tail window and aggregation for iS@1 ($N\!=\!32$, with outlier trimming). \emph{Full}: all tokens. $\sqrt{L}$: tail $= \lfloor\sqrt{\bar{L}}\rfloor$. $\sqrt{L}/\bar{H}$ (ours): tail $= \lfloor\sqrt{\bar{L}}/\bar{H}\rfloor$ where $\bar{H}$ is the mean per-token entropy.}
\label{tab:tail-ablation}
\small
\begin{tabular}{l cc cc cc}
\toprule
 & \multicolumn{2}{c}{$L=2048$} & \multicolumn{2}{c}{$\sqrt{L}$} & \multicolumn{2}{c}{$\sqrt{L}/\bar{H}$ (ours)} \\
\cmidrule(lr){2-3} \cmidrule(lr){4-5} \cmidrule(lr){6-7}
 & Mean & Median & Mean & Median & Mean & Median \\
\midrule
  AIME 24 & $65.8_{\pm 1.6}$ & $66.7_{\pm 4.3}$ & $\textbf{71.7}_{\pm 2.2}$ & $70.8_{\pm 3.7}$ & $65.8_{\pm 0.8}$ & $69.2_{\pm 4.4}$ \\
  AIME 25 & $46.7_{\pm 3.0}$ & $50.0_{\pm 2.4}$ & $43.3_{\pm 4.7}$ & $49.2_{\pm 3.4}$ & $45.0_{\pm 2.2}$ & $\textbf{53.3}_{\pm 3.0}$ \\
  AIME 26 & $53.3_{\pm 3.6}$ & $\textbf{57.5}_{\pm 3.7}$ & $49.2_{\pm 2.8}$ & $52.5_{\pm 2.1}$ & $52.5_{\pm 3.7}$ & $50.8_{\pm 2.1}$ \\
  HMMT 25 & $36.7_{\pm 1.4}$ & $35.0_{\pm 2.2}$ & $35.0_{\pm 1.0}$ & $\textbf{39.2}_{\pm 0.8}$ & $30.0_{\pm 1.4}$ & $34.2_{\pm 2.1}$ \\
  HMMT 26 & $30.3_{\pm 2.1}$ & $29.5_{\pm 1.5}$ & $31.1_{\pm 1.9}$ & $30.3_{\pm 1.2}$ & $31.8_{\pm 0.9}$ & $\textbf{32.6}_{\pm 1.5}$ \\
\midrule
  LCB-100 & $19.8_{\pm 1.1}$ & $21.0_{\pm 1.1}$ & $\textbf{22.2}_{\pm 1.3}$ & $21.5_{\pm 1.0}$ & $21.5_{\pm 1.7}$ & $21.8_{\pm 0.8}$ \\
\midrule
  GPQA & $44.6_{\pm 1.1}$ & $43.9_{\pm 0.6}$ & $44.9_{\pm 1.2}$ & $\textbf{46.0}_{\pm 1.5}$ & $45.5_{\pm 0.7}$ & $\textbf{46.0}_{\pm 0.5}$ \\
\bottomrule
\end{tabular}
\end{table}

Table~\ref{tab:tail-ablation} contrasts different tail formulations and aggregation strategies across mathematics, coding, and reasoning tasks. A fixed full-sequence window ($L=2048$) typically yields sub-optimal performance because it incorporates noise from the early problem-solving phases. By shifting to a sublinear scaling factor, specifically the square root of the mean length ($\sqrt{L}$), the scoring mechanism effectively zeroes in on the concluding steps, leading to measurable improvements across multiple domains.

Our proposed adaptive tail strategy, $\sqrt{L}/\bar{H}$, further scales this window inversely by the mean per-token entropy of the sample set. The underlying intuition is that when the model exhibits high overall uncertainty across trajectories, a narrower tail is necessary to avoid integrating uninformative entropy spikes. The results show that the adaptive tail consistently rivals or outperforms the static $\sqrt{L}$ approach. Furthermore, when comparing aggregation functions, the median is demonstrably superior to the mean. The mean is highly sensitive to isolated entropy spikes caused by punctuation, formatting tokens, or minor syntactic uncertainties, whereas the median provides a statistically robust summary of the trajectory's structural confidence.

\begin{table}[ht!]
\centering
\caption{Effect of length outlier trimming on iS@1 ($\sqrt{L}/\bar{H}$, median). Trimming excludes responses outside the $[5\%, 95\%]$ length percentile before computing the adaptive tail window.}
\label{tab:trim-ablation}
\small
\begin{tabular}{l cc}
\toprule
 & No trim & Trim (ours) \\
\midrule
  AIME 24 & $68.3_{\pm 4.2}$ & $\textbf{69.2}_{\pm 4.4}$ \\
  AIME 25 & $\textbf{53.3}_{\pm 3.0}$ & $\textbf{53.3}_{\pm 3.0}$ \\
  AIME 26 & $\textbf{52.5}_{\pm 2.1}$ & $50.8_{\pm 2.1}$ \\
  HMMT 25 & $\textbf{34.2}_{\pm 2.1}$ & $\textbf{34.2}_{\pm 2.1}$ \\
  HMMT 26 & $\textbf{32.6}_{\pm 1.5}$ & $\textbf{32.6}_{\pm 1.5}$ \\
\midrule
  LCB-100 & $21.5_{\pm 0.6}$ & $\textbf{21.8}_{\pm 0.8}$ \\
\midrule
  GPQA & $45.7_{\pm 0.4}$ & $\textbf{46.0}_{\pm 0.5}$ \\
\bottomrule
\end{tabular}
\end{table}

Additionally, we evaluate the impact of length-based outlier trimming. Since our adaptive tail is a function of the set's mean length ($\bar{L}$) and mean entropy ($\bar{H}$), extreme pathological behaviors - such as infinite generation loops, repeating tokens, or premature stopping - can heavily skew these set-level statistics. Table~\ref{tab:trim-ablation} details the effect of systematically excluding responses outside the $[5\%, 95\%]$ length percentile prior to computing the adaptive window. While the improvements are marginal in perfectly well-behaved generations, trimming acts as a critical stabilizing mechanism. It yields slight but consistent gains in AIME 2024, coding (LCB-100), and general reasoning (GPQA) datasets, effectively ensuring that pathological generations do not compromise the selection of otherwise valid, high-confidence trajectories.

\clearpage
\subsection{Image-to-Code CAD Scaling}
\label{appx:image-to-code}

Extending inference-time scaling to engineering design, specifically image-to-code computer-aided design (CAD) generation, presents a unique systemic challenge. Unlike programmatic domains where unit tests are relatively lightweight, verifying the correctness of generated CAD code requires executing complex geometric kernels to render 3D meshes and subsequently computing volumetric overlap metrics, such as Intersection over Union (IoU). This rendering and evaluation pipeline, executed via engines like CadQuery, introduces a severe computational bottleneck, often dominating the inference loop and rendering traditional Best-of-$N$ scaling with external verifiers practically intractable at scale. Consequently, CAD generation serves as an ideal testbed for evaluating whether Intrinsic Selection (\texttt{iS}) can successfully bypass the need for costly environmental simulators.

\begin{table}[ht!]
\caption{Fusion360 IoU results over 100 test problems. Intrinsic selection methods evaluated across sample budgets $N$. Results show median$\pm$std. Model: Qwen2.5-VL-7B tuned on GenCAD-Code dataset; sampling temperature=1.0. 
We test entropy selection using the full logits and the top10. 
IoU is computed between generated and ground-truth 3D meshes via CadQuery.}
\label{tab:fusion360-median-iou-appx}
\centering
\small
\begin{tabular}{lccccccc}
\toprule
Method & $N{=}2$ & $N{=}4$ & $N{=}8$ & $N{=}16$ & $N{=}32$ & $N{=}64$ & $N{=}128$ \\
\midrule
pass@1 & $.396{\scriptstyle\pm.029}$ & $.390{\scriptstyle\pm.026}$ & $.392{\scriptstyle\pm.019}$ & $.393{\scriptstyle\pm.014}$ & $.394{\scriptstyle\pm.011}$ & $.391{\scriptstyle\pm.009}$ & $.388$ \\
\texttt{iS}@1 (\text{full}) & $.409{\scriptstyle\pm.041}$ & $.418{\scriptstyle\pm.042}$ & $.408{\scriptstyle\pm.041}$ & $.414{\scriptstyle\pm.045}$ & $.429{\scriptstyle\pm.039}$ & $.448{\scriptstyle\pm.037}$ & $.472$ \\
\texttt{iS}@1 (\text{top10}) & $.404{\scriptstyle\pm.032}$ & $.393{\scriptstyle\pm.036}$ & $.413{\scriptstyle\pm.041}$ & $.419{\scriptstyle\pm.034}$ & $.433{\scriptstyle\pm.043}$ & $\mathbf{.456}{\scriptstyle\pm.037}$ & $.463$ \\
\midrule
pass@$N$ (oracle) & $.526{\scriptstyle\pm.029}$ & $.602{\scriptstyle\pm.022}$ & $.677{\scriptstyle\pm.020}$ & $.745{\scriptstyle\pm.020}$ & $.804{\scriptstyle\pm.018}$ & $.842{\scriptstyle\pm.007}$ & $.881$ \\
\bottomrule
\end{tabular}
\end{table}

Table~\ref{tab:fusion360-median-iou-appx} details the scaling behavior of \texttt{iS} on the out-of-distribution Fusion360 dataset using a fine-tuned multimodal architecture (Qwen2.5-VL-7B). As expected, standard parallel sampling (\texttt{pass@1}) remains stagnant at an average IoU of approximately $0.39$ regardless of the sample budget $N$, as it merely selects a random trajectory without any quality-based discrimination. 

In stark contrast, \texttt{iS} demonstrates a clear ability to identify higher-quality geometries without ever invoking the external CadQuery solver. We compare two specific variants of our selection mechanism: computing tail entropy over the full vocabulary distribution versus a truncated top-10 logit distribution. Both variants successfully scale with increased compute budget. At $N=128$, the full-logit \texttt{iS} approach elevates the mean IoU to $0.472$, demonstrating a consistent and predictable scaling trajectory. While the top-10 variant peaks slightly lower at $N=128$, it achieves the highest intermediate performance at $N=64$ ($0.456$), indicating that the primary discriminative signal is heavily concentrated in the most likely token predictions.

Finally, while the oracle upper bound (\texttt{pass@$N$}) reaches an impressive IoU of $0.881$ at $N=128$ - confirming that the model is highly capable of generating accurate meshes given enough attempts - \texttt{iS} effectively bridges a substantial portion of the gap between the zero-shot baseline and the oracle. These results confirm that token-level entropy inherently captures latent signals regarding the structural and geometric validity of generated programmatic code, proving \texttt{iS} to be a highly effective scaling mechanism for complex, hard-to-verify engineering tasks.

\clearpage

\subsection{Difficulty Estimation}
\label{appx:difficulty}

\begin{figure}[b]
    \centering
    \includegraphics[width=\linewidth]{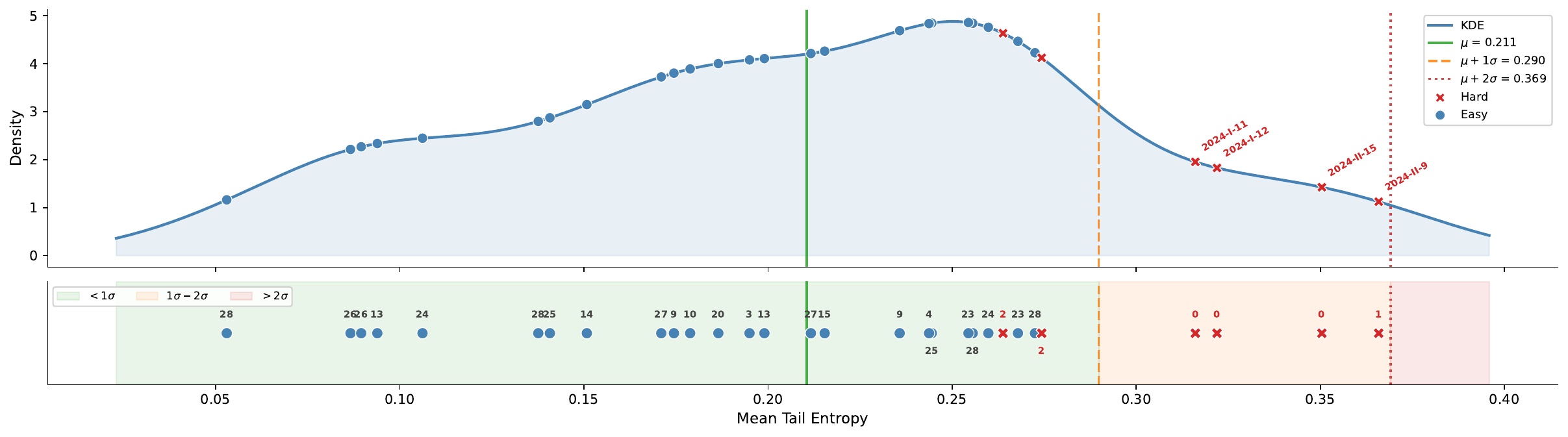}
    \caption{Problem-level entropy and estimated difficulty on math datasets. Higher adjusted mean per-token tail entropy correlates strongly with harder problems. Instances more than $1\sigma$ above the mean are tightly associated with hardness, defined empirically as yielding at most two correct solutions within a 32-sample budget ($n/N$ with $n \in \{0,1,2\}$).}
    \label{fig:entropy-difficulty-appx}
\end{figure}

To evaluate our verifier-free approach for estimating problem difficulty, we analyzed generation statistics from \texttt{Qwen3-4B-Instruct-2507}. For each problem across the AIME and HMMT benchmarks, we generated $N=32$ parallel samples over four random seeds. A problem was empirically classified as ``hard'' if it yielded two or fewer correct solutions per seed ($n \le 2$). 

To compute the intrinsic difficulty score, we first filtered out pathological generations by trimming the top 5\% of length outliers. We then calculated the average mean entropy over the final 2048 tokens (or the full sequence length, if shorter) across the remaining sample set. Based on this problem-level score, we dynamically established an adaptive selection budget ($k$) defined by a threshold of one standard deviation above the mean entropy across the evaluation set. Problems exceeding this threshold trigger the high-entropy gate, indicating that they require advanced inference scaling, such as intrinsic particle filtering, rather than simple parallel selection.

\begin{table}[ht!]
  \centering
  \caption{Performance of the tail entropy difficulty estimation method across mathematical reasoning datasets. Results report the mean $\pm$ standard deviation across 4 random splits of $N=32$ budget per problem. The ``Hard'' subset is defined as problems where the number of correct solutions is very sparse ($n \in \{0,1,2\}$) given the available budget. $S$ denotes the total number of problems in the dataset.}
  \label{tab:tail_entropy_difficulty_results}
  \resizebox{\textwidth}{!}{%
  \begin{tabular}{lccccccc}
    \toprule
    Dataset & $S$ & Hard Problems & Adaptive Budget ($k$) & AUROC & AP & R@k & P@k \\
    \midrule
    AIME 2024     & 30  & $5.2 \pm 0.4$  & $6.2 \pm 0.4$  & $0.951 \pm 0.026$ & $0.848 \pm 0.044$ & $0.808 \pm 0.014$ & $0.679 \pm 0.021$ \\
    AIME 2025     & 30  & $10.0 \pm 0.0$ & $6.5 \pm 1.1$  & $0.850 \pm 0.018$ & $0.800 \pm 0.009$ & $0.500 \pm 0.071$ & $0.779 \pm 0.091$ \\
    AIME 2026     & 30  & $7.2 \pm 0.4$  & $4.2 \pm 0.4$  & $0.952 \pm 0.025$ & $0.895 \pm 0.041$ & $0.589 \pm 0.078$ & $1.000 \pm 0.000$ \\
    HMMT 2025 & 30  & $16.0 \pm 0.7$ & $7.0 \pm 1.2$  & $0.792 \pm 0.018$ & $0.839 \pm 0.009$ & $0.390 \pm 0.047$ & $0.902 \pm 0.057$ \\
    HMMT 2026 & 33  & $19.5 \pm 0.9$ & $5.2 \pm 1.1$  & $0.807 \pm 0.022$ & $0.865 \pm 0.014$ & $0.257 \pm 0.039$ & $0.964 \pm 0.062$ \\
    \bottomrule
  \end{tabular}%
  }
\end{table}

As demonstrated in Table~\ref{tab:tail_entropy_difficulty_results}, set-level tail entropy serves as a highly reliable difficulty estimator for structured mathematical reasoning. The method demonstrates strong ranking quality, achieving an AUROC greater than 0.79 across all five math datasets. Notably, on AIME 2024 and AIME 2026, the AUROC exceeds 0.95, indicating an excellent ability to separate hard problems from easy ones based purely on the model's internal statistics. Furthermore, the precision at the adaptive budget (P@k) is remarkably high, reaching 100\% on AIME 2026 and exceeding 90\% on both HMMT datasets. This confirms that when the intrinsic signal triggers the high-entropy difficulty gate, the problem is almost certainly challenging and necessitates the allocation of sequential compute resources.

\begin{table}[ht!]
    \centering
    \caption{AIME 2024 Summary Metrics (Hard vs. Easy). From a distributional perspective, entropy, certainty, and sequence length are strongly correlated (or inversely correlated) with problem difficulty. Hard problems are defined as those where $pass@N \rightarrow n/N$ with $n \in \{0,1,2\}$. The arrows indicate the direction associated with an increased likelihood of a correct response (i.e., an ``easy'' problem). All metrics are computed per token and averaged over the respective sets.}
    \label{tab:difficulty-appx}
    \begin{tabular}{llccc}
        \toprule
        {Metric} & {Aggregation} & {Hard Set} & {Easy Set} & {Ratio (Hard/Easy)} \\
        \midrule
        Entropy $\downarrow$  & Mean      & 0.3324     & 0.1906     & 1.74 \\
        Entropy $\downarrow$   & Median    & 0.3295     & 0.1878     & 1.75 \\
        Entropy $\downarrow$   & Std       & 0.0851     & 0.0512     & 1.66 \\
        Entropy $\downarrow$   & Sum Mean  & 462.77     & 254.38     & 1.82 \\
        \midrule
        Certainty $\uparrow$  & Mean      & 28.46    & 34.41    & 0.83 \\
        Certainty $\uparrow$ & Median    & 28.05    & 34.32    & 0.82 \\
        Certainty $\uparrow$ & Std       & 2.14     & 2.32     & 0.92 \\
        Certainty $\uparrow$ & Sum Mean  & 39611.16 & 45975.19 & 0.86 \\
        \midrule
        Length $\downarrow$    & Mean      & 23109.02 & 12507.44 & 1.85 \\
        Length $\downarrow$    & Median    & 23291.00 & 12492.61 & 1.86 \\
        Length $\downarrow$    & Std       & 3487.74  & 3097.56  & 1.13 \\
        Length $\uparrow$   & CV        & 0.15     & 0.30     & 0.48 \\
        \bottomrule
    \end{tabular}
\end{table}

To further illustrate the underlying mechanics of this discriminative signal, Table~\ref{tab:difficulty-appx} provides a detailed distributional breakdown of the AIME 2024 dataset, contrasting the ``Hard'' and ``Easy'' subsets. The data clearly shows that token entropy and sequence length are substantially elevated for harder problems. Specifically, the mean per-token entropy for the hard set is 1.74 times higher than that of the easy set, and the mean sequence length is 1.85 times longer. This aligns with the intuition that uncertainty correlates directly with the model's capacity to solve the problem; when faced with a difficult task, the model explores longer, more convoluted reasoning paths with less confidence at each generative step.

\begin{figure}[ht!]
    \centering
    \includegraphics[width=\linewidth]{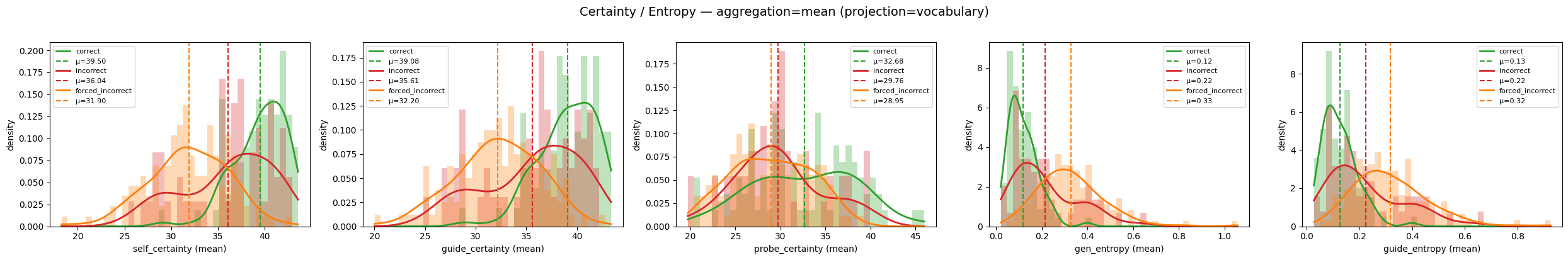}
    \caption{Distributional visualization of token-level entropy and certainty metrics comparing hard and easy problem sets on AIME 2024. The distinct separation between the distributions underscores the reliability of intrinsic statistics as a verifier-free gate for dynamic compute allocation.}
    \label{fig:placeholder}
\end{figure}

\clearpage
\subsection{Intrinsic Particle Filtering}
\label{appx:ipf}
\begin{figure}[ht!]
    \centering
    \begin{subfigure}[b]{0.45\textwidth}
        \centering
        \includegraphics[width=\textwidth]{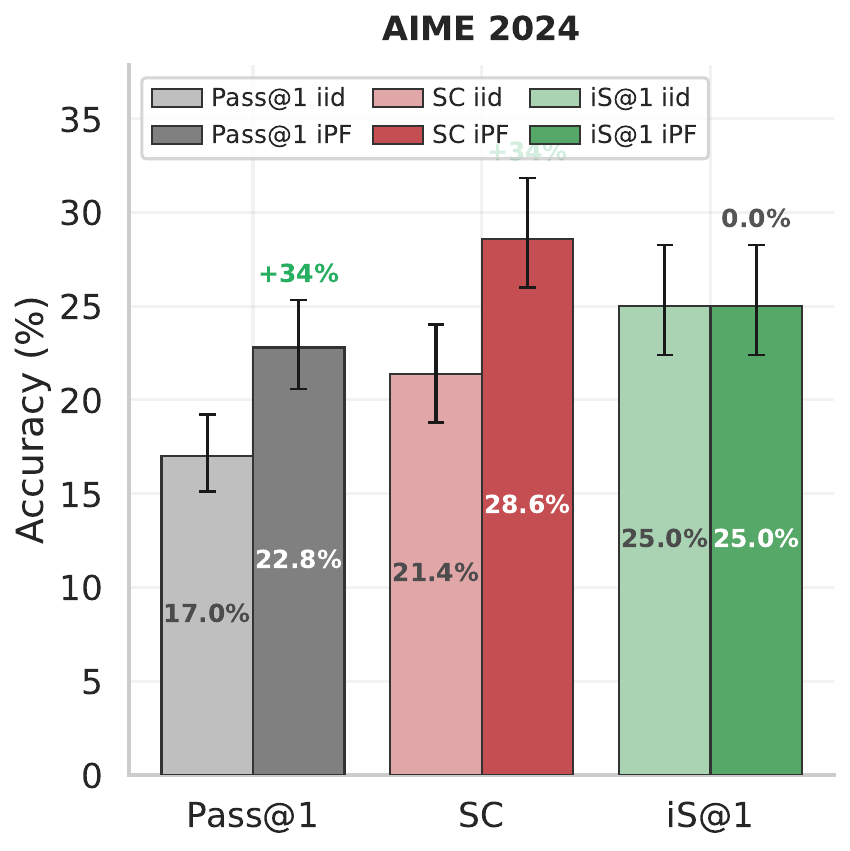}
        \caption{AIME 2024}
        \label{fig:ipf-a24-intro-appx}
    \end{subfigure}
    \begin{subfigure}[b]{0.45\textwidth}
        \centering
        \includegraphics[width=\textwidth]{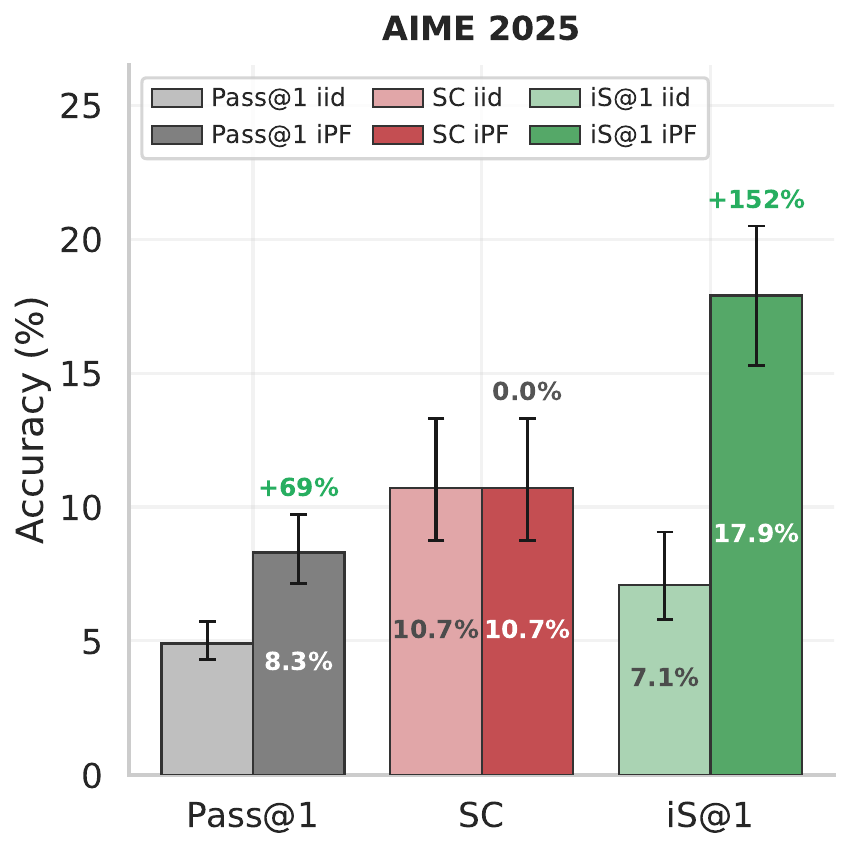}
        \caption{AIME 2025}
        \label{fig:ipf-a25-intro-appx}
    \end{subfigure}

    \par\vspace{1em} %
    
    \begin{subfigure}[b]{0.45\textwidth}
        \centering
        \includegraphics[width=\textwidth]{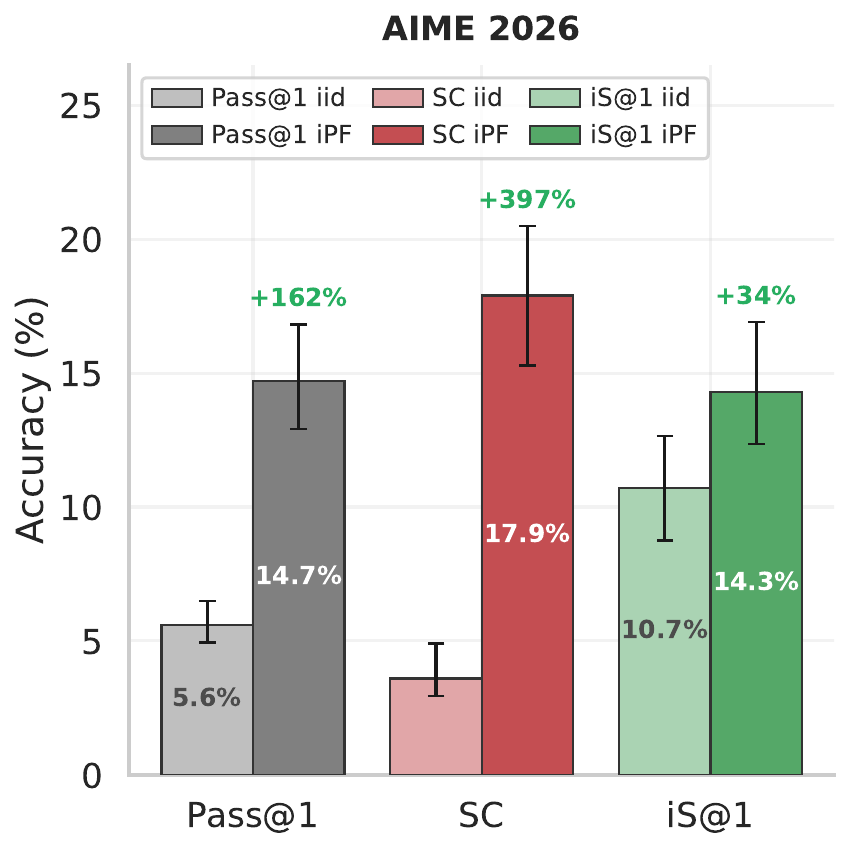}
        \caption{AIME 2026}
        \label{fig:ipf-a26-intro-appx}
    \end{subfigure}
    \begin{subfigure}[b]{0.45\textwidth}
        \centering
        \includegraphics[width=\textwidth]{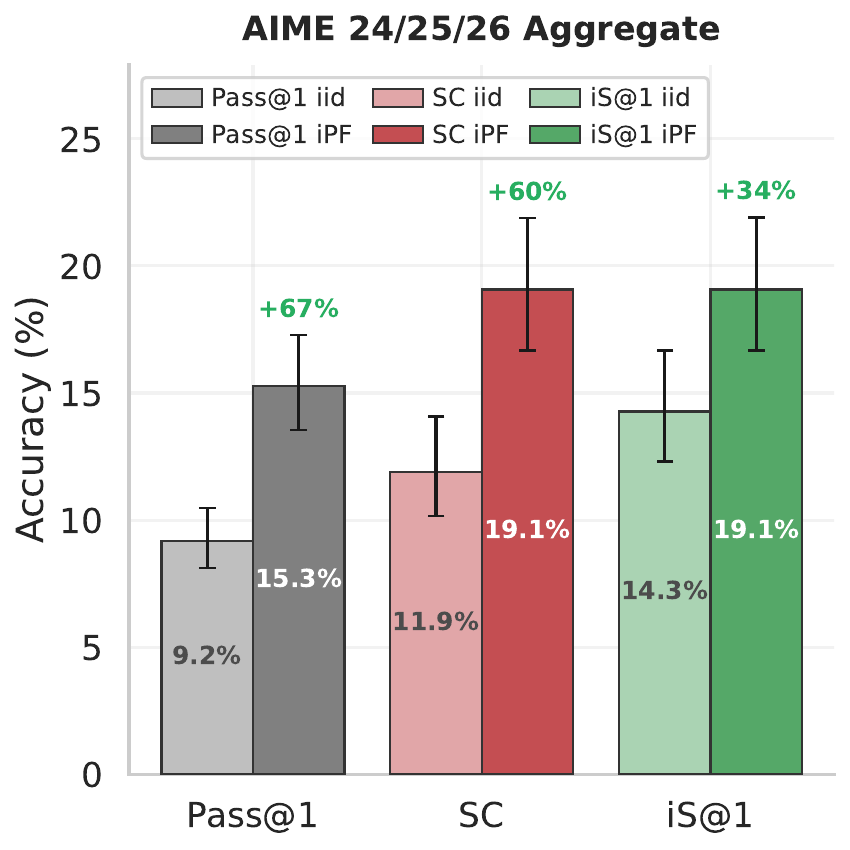}
        \caption{AIME Aggregate (2024--2026)}
        \label{fig:ipf-a242526-intro-appx}
    \end{subfigure}
    \caption{Performance of Intrinsic Particle Filtering (\texttt{iPF}) across the AIME benchmark datasets. The subfigures illustrate accuracy improvements over standard parallel sampling for individual years (a--c) and the overall aggregate performance (d). By leveraging internal token-level entropy to guide step-level resampling, \texttt{iPF} actively concentrates computational resources on high-confidence reasoning trajectories, consistently outperforming the independent sampling baseline on the hardest mathematical problems.}
    \label{fig:ipf-aime-appx}
\end{figure}

\clearpage

\subsection{Particle Distillation}
\label{appx:particle-distillation}

When models encounter problems with systematic failure modes, where early faulty assumptions lead to consistently incorrect reasoning trajectories, simply scaling independent parallel samples (e.g., to $N=32$) yields heavily diminishing returns. To overcome these structural bottlenecks, Particle Distillation (\texttt{dPF}) injects privileged information, such as teacher demonstrations or critiques, via early logit blending and KL-guided resampling. We evaluate this approach on the AIME and HMMT mathematical reasoning datasets to determine if guided resampling can steer the model toward correct solutions more efficiently than unguided scaling.

\begin{table}[ht!]
\centering
\caption{Particle distillation results for mathematics. \texttt{dPF} with
$N{=}8$ particles and guided resampling vs.\ $N{=}32$ i.i.d.\
samples. Guidance: hints from Claude (teacher demonstration or
critique of the best sample selected via \texttt{iS}).}
\label{tab:phase2-distill-appx}
\resizebox{\textwidth}{!}{%
\begin{tabular}{lll ccccc c}
\toprule
\multirow{2}{*}{Method} & \multirow{2}{*}{Hint ($\vh$)} & \multirow{2}{*}{Blend ($\alpha$)} &  \multicolumn{5}{c}{Dataset} & \multirow{2}{*}{wt avg} \\
\cmidrule(lr){4-8}
 & & & AIME 2024 & AIME 2025 & AIME 2026 & HMMT 2025 & HMMT 2026 & \\
\midrule
\multicolumn{7}{l}{\textit{pass@1}} \\
\midrule
Sampling $(N=32)$ & - & -         & 56.7 & 44.3 & 50.1 & 30.0 & 28.2 & 41.6 \\
\texttt{dPF} $(N=8)$ & Teacher & w/o       & 59.2 & 42.5 & \textbf{62.5} & {32.1} & {35.2} & \textbf{46.1} \\
\texttt{dPF} $(N=8)$ & Teacher & w/        & {62.9} & 44.6 & 55.4 & 31.7 & 33.0 & 45.3 \\
\texttt{dPF} $(N=8)$ & Critique & w/o      & \textbf{64.6} & \textbf{47.9} & 51.7 & 29.6 & 32.6 & 45.0 \\
\texttt{dPF} $(N=8)$ & Critique & w/       & 62.5 & 43.3 & 57.5 & \textbf{32.5} & \textbf{37.1} & \textbf{46.4} \\
\midrule
\multicolumn{7}{l}{\textit{pass@8}} \\
\midrule
Sampling $(N=32)$ & - & -    & 81.6 & 66.0 & 74.7 & 46.9 & 41.8 & 61.8 \\
\texttt{dPF} $(N=8)$ & Teacher & w/o        & {83.3} & 66.7 & \textbf{80.0} & 43.3 & \textbf{57.6} & \textbf{66.0} \\
\texttt{dPF} $(N=8)$ & Teacher & w/          & \textbf{86.7} & 63.3 & 73.3 & 43.3 & 51.5 & 63.4 \\
\texttt{dPF} $(N=8)$ & Critique & w/o       & 83.3 & \textbf{73.3} & 70.0 & {46.7} & 42.4 & 62.7 \\
\texttt{dPF} $(N=8)$ & Critique & w/         & \textbf{86.7} & 63.3 & 73.3 & \textbf{50.0} & 45.5 & 63.4 \\
\midrule
\multicolumn{7}{l}{\textit{Intrinsic Selection}} \\
\midrule
Sampling $(N=32)$ & - & -   & 66.7 & 53.3 & 53.3 & 33.3 & 36.4 & 48.4 \\
\texttt{dPF} $(N=8)$ & Teacher & w/o       & 60.0 & 46.7 & \textbf{63.3} & 30.0 & \textbf{45.5} & 49.0 \\
\texttt{dPF} $(N=8)$ & Teacher & w/           & \textbf{76.7} & 46.7 & 53.3 & \textbf{33.3} & 42.4 & \textbf{50.3} \\
\texttt{dPF} $(N=8)$ & Critique & w/o      & 70.0 & \textbf{63.3} & 53.3 & 30.0 & 36.4 & \textbf{50.3} \\
\texttt{dPF} $(N=8)$ & Critique & w/         & 66.7 & 33.3 & {56.7} & \textbf{33.3} & 39.4 & 45.8 \\
\midrule
\multicolumn{7}{l}{\textit{pass@32}} \\
\midrule
Sampling $(N=32)$ & - & - & 86.7 & 70.0 & 80.0 & 56.7 & 51.5 & 68.4 \\
\bottomrule
\end{tabular}%
}
\end{table}

The empirical results, detailed in Table~\ref{tab:phase2-distill-appx}, demonstrate the significant efficiency and steering capabilities of \texttt{dPF}. Using merely 8 particles guided by privileged hints, \texttt{dPF} consistently matches or outperforms the standard 32-sample independent and identically distributed (i.i.d.) parallel generation baseline across multiple evaluation metrics. For instance, under pass@1, \texttt{dPF} utilizing critique-based hints combined with early logit blending (w/ $\alpha$) achieves a weighted average accuracy of 46.4\%, marking a substantial improvement over the 41.6\% unguided baseline. Similar efficiency gains are reflected in the pass@8 and Intrinsic Selection metrics, where 8 guided particles reliably surpass the performance of 32 unguided samples.

A crucial observation from the ablation data is the domain-specific effectiveness of the guidance type. Teacher demonstrations, which provide concrete procedural steps, yield the strongest performance on the highly non-obvious reasoning paths of the HMMT datasets (reaching 57.6\% pass@8 on HMMT 2026 without blending). Conversely, critique-based hints, which actively identify and correct early logical errors in the model's generated candidates, prove more advantageous on the AIME datasets. Furthermore, integrating early logit blending generally enhances performance on the most challenging problems by forcefully steering the model past systematic early assumptions before it commits to a flawed trajectory. Overall, these findings confirm that \texttt{dPF} efficiently distills privileged information to navigate complex problem spaces where standard parallel scaling fails.

\clearpage
\subsection{Healthcare}
\label{appx:healthcare}

To evaluate the transferability of our inference scaling framework beyond structured domains like mathematics and coding, we tested our methods on HealthBench-Hard, an open-ended clinical reasoning benchmark. Unlike exact-match environments, evaluating clinical responses requires multi-dimensional subjective grading against expert rubrics. Table~\ref{tab:healthbench_mean-appx} details the performance of standard parallel sampling (\texttt{pass@1}), Intrinsic Selection (\texttt{iS}), Intrinsic Particle Filtering (\texttt{iPF}), and Particle Distillation (\texttt{dPF}) across seven distinct clinical themes.

\begin{table}[ht!]
    \centering
    \caption{HealthBench-Hard Results. Evaluation utilizes MedGemma-4B-IT for generation and MedGemma-27B as the judge to compute rubric-based scores. Bold values indicate the highest score per theme.}
    \label{tab:healthbench_mean-appx}
    \begin{tabular}{lcccc}
        \toprule
        Theme & pass@1 & iS & iPF & dPF \\
        \midrule
        Communication & 0.440 & 0.412 & 0.413 & \textbf{0.504} \\
        Complex Responses & 0.362 & 0.362 & 0.369 & \textbf{0.458} \\
        Context Seeking & 0.371 & 0.410 & \textbf{0.450} & 0.342 \\
        Emergency Referrals & \textbf{0.318} & 0.265 & 0.317 & 0.279 \\
        Global Health & 0.387 & 0.355 & 0.384 & \textbf{0.407} \\
        Health Data Tasks & 0.317 & 0.320 & \textbf{0.419} & 0.333 \\
        Hedging & 0.405 & 0.421 & \textbf{0.466} & 0.431 \\
        \midrule
        Problem Weighted & 0.378 & 0.372 & \textbf{0.411} & 0.396 \\
        Theme Weighted & 0.371 & 0.364 & \textbf{0.403} & 0.393 \\
        \bottomrule
    \end{tabular}
\end{table}

The empirical results reveal that step-level interventions (\texttt{iPF} and \texttt{dPF}) are essential for driving generation quality in complex, open-ended clinical spaces. \texttt{iPF} emerges as the strongest overall method, achieving the highest problem-weighted (0.411) and theme-weighted (0.403) scores. It specifically excels on themes that require deep, sustained reasoning, such as Health Data Tasks (0.419) and Context Seeking (0.450). In these categories, the ability of \texttt{iPF} to concentrate compute on trajectories with sustained low entropy strongly correlates with successfully completing logical diagnostic chains. 

Conversely, \texttt{dPF} acts as the optimal strategy for themes constrained by strict, highly specific criteria. By utilizing a guide model conditioned on the problem's rubric to steer early logit blending and resampling, \texttt{dPF} provides a highly targeted signal. This results in dominant performance in Communication (0.504) and Complex Responses (0.458), successfully forcing the model to include mandatory pharmacological guidelines or bedside manner protocols that standard sampling often ignores.

\begin{figure}[ht!]
    \centering
    \includegraphics[width=.6\linewidth]{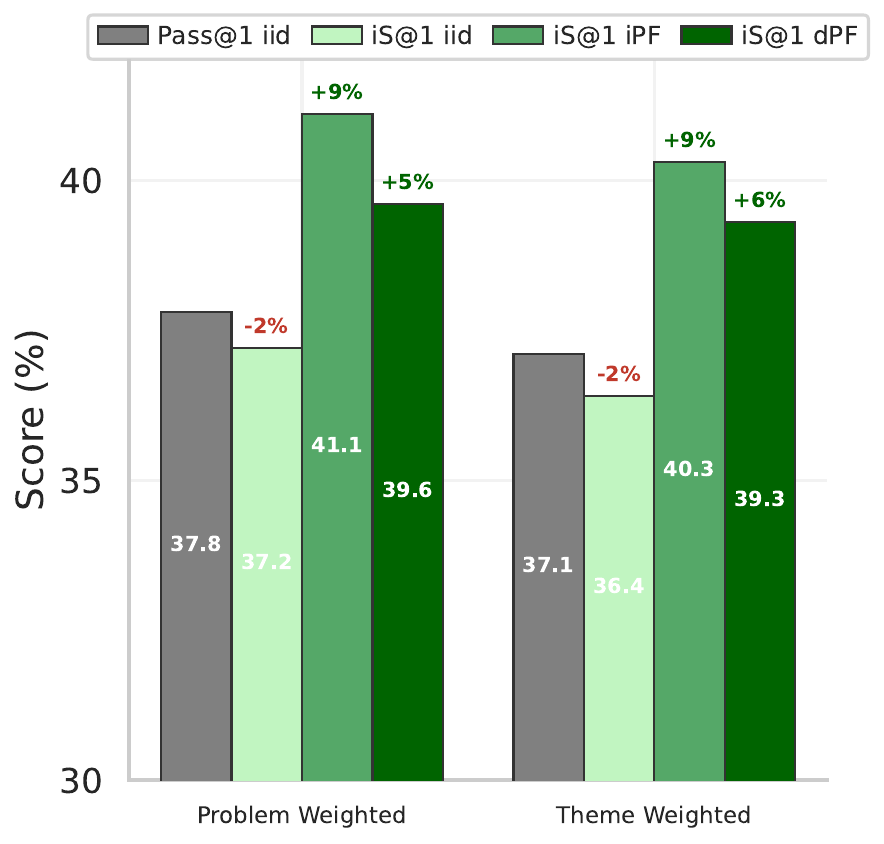}
    \caption{Average rubric scores across HealthBench-Hard clinical themes. While \texttt{iPF} yields the highest overall problem-weighted improvements by enhancing logical consistency, \texttt{dPF} delivers the largest targeted gains on highly constrained themes (e.g., Complex Responses) by distilling rubric criteria directly into the generative trajectory. Intrinsic selection (\texttt{iS}) slightly underperforms the baseline, highlighting that confidence alone does not guarantee adherence to multidimensional clinical rubrics.}
    \label{fig:healthbench_averages_paper}
\end{figure}

Interestingly, standard Intrinsic Selection (\texttt{iS}) generally underperforms the \texttt{pass@1} baseline across most themes, degrading the problem-weighted average slightly from 0.378 to 0.372. While \texttt{iS} reliably selects the most confident response generated by the model, internal confidence does not guarantee comprehensive coverage of a multidimensional clinical rubric. This contrast perfectly highlights the complementary nature of our framework: while \texttt{iS} is a highly effective selection mechanism for scalar verification tasks, open-ended domains requiring adherence to complex external criteria necessitate the active step-level steering provided by \texttt{iPF} and \texttt{dPF}. Finally, the baseline model retains a slight edge on Emergency Referrals, suggesting that for highly urgent and straightforward triage decisions, standard zero-shot sampling avoids the over-complication that can arise from advanced resampling techniques.

\clearpage

\section{Entropy Correlation}
\label{appx:entropy-correlation}

\begin{figure}[ht!]
    \centering
    \includegraphics[width=0.5\linewidth]{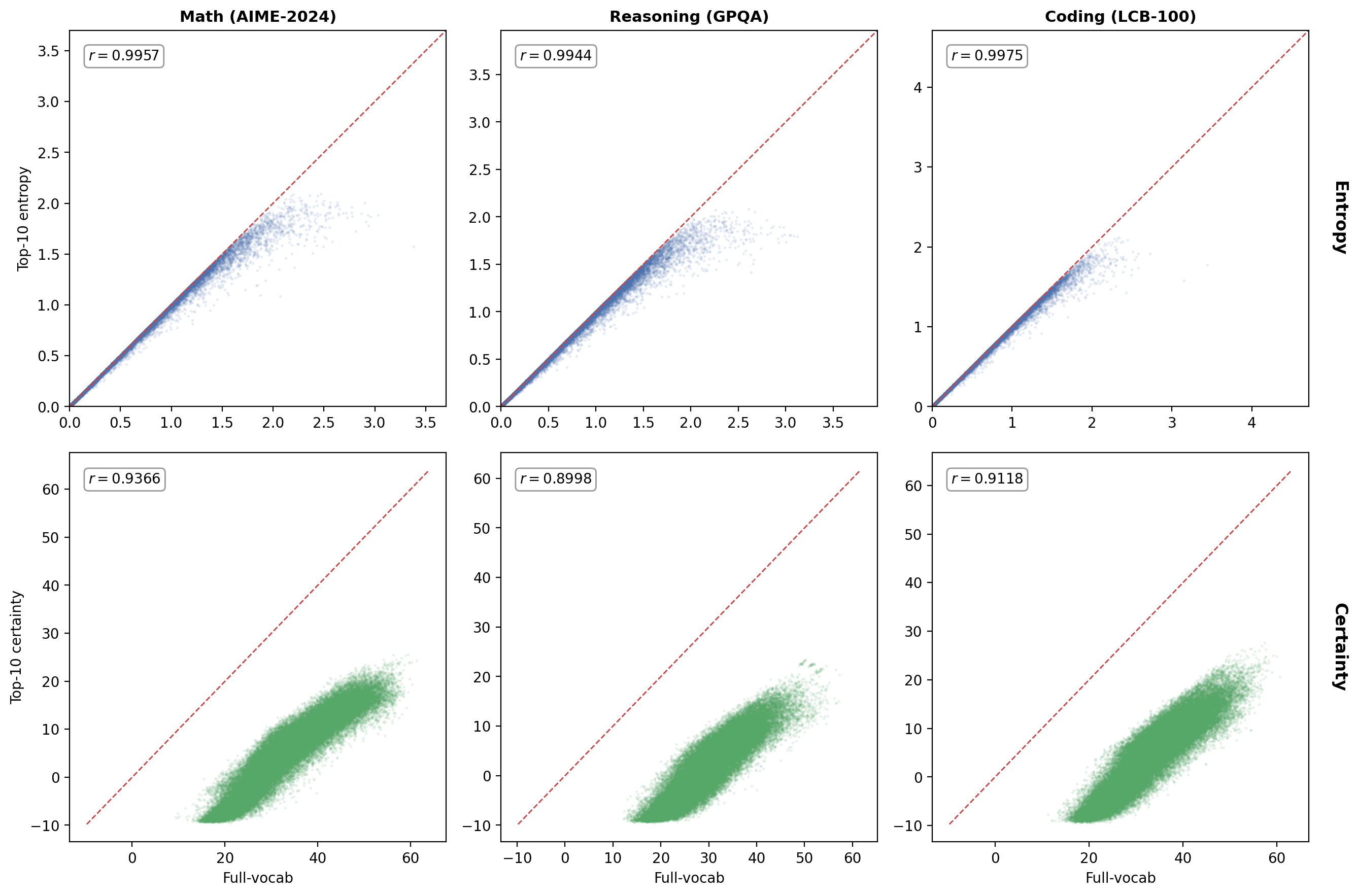}
    \caption{Scatter plot comparing top-$k$ partial log-probabilities against full-vocabulary log-probabilities for both entropy and certainty metrics. The near-perfect linear alignment for entropy highlights its robustness to vocabulary truncation.}
    \label{fig:entropy_certainty_scatter}
\end{figure}

\begin{figure}[ht!]
    \centering
    \includegraphics[width=0.45\linewidth]{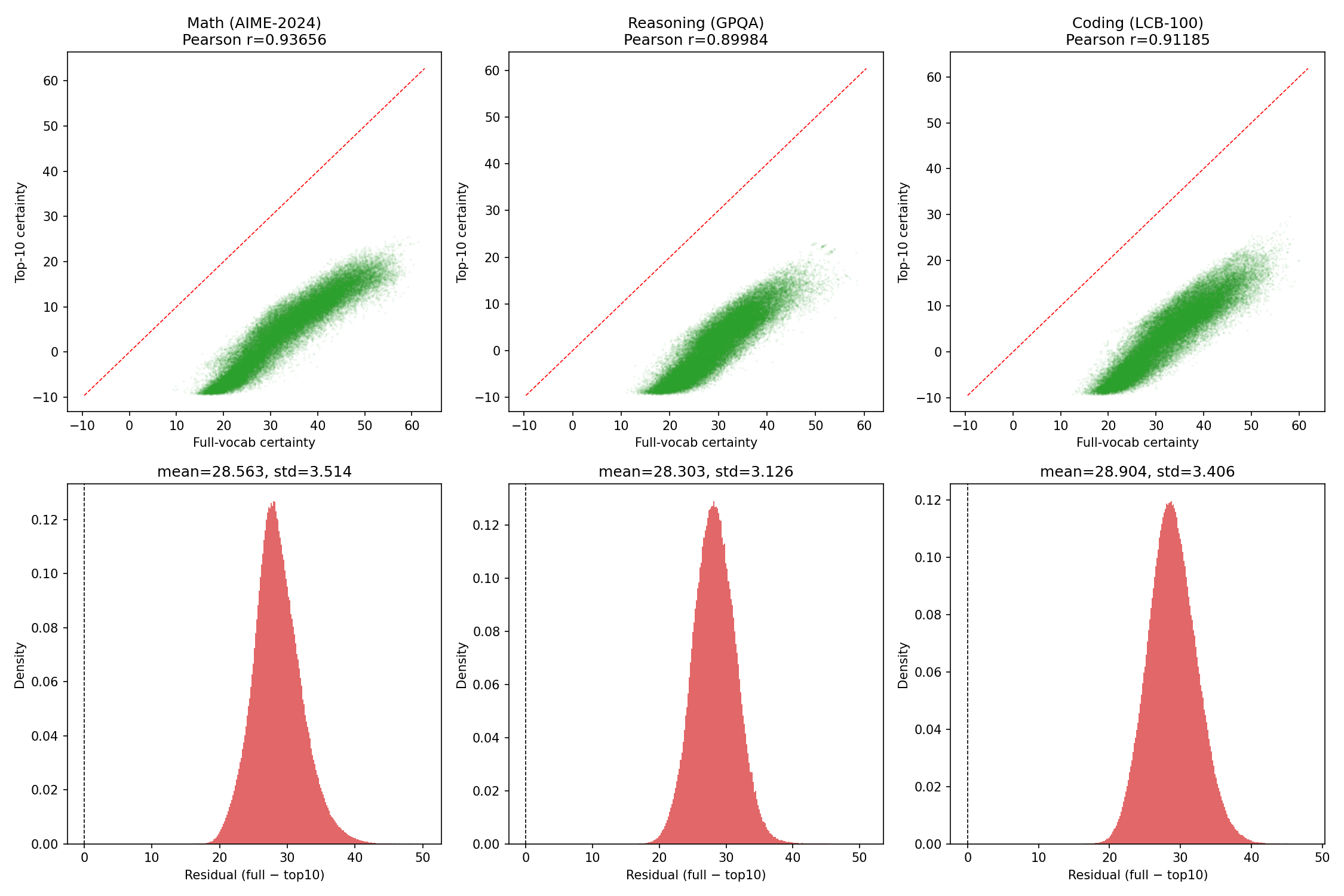}
    \includegraphics[width=0.45\linewidth]{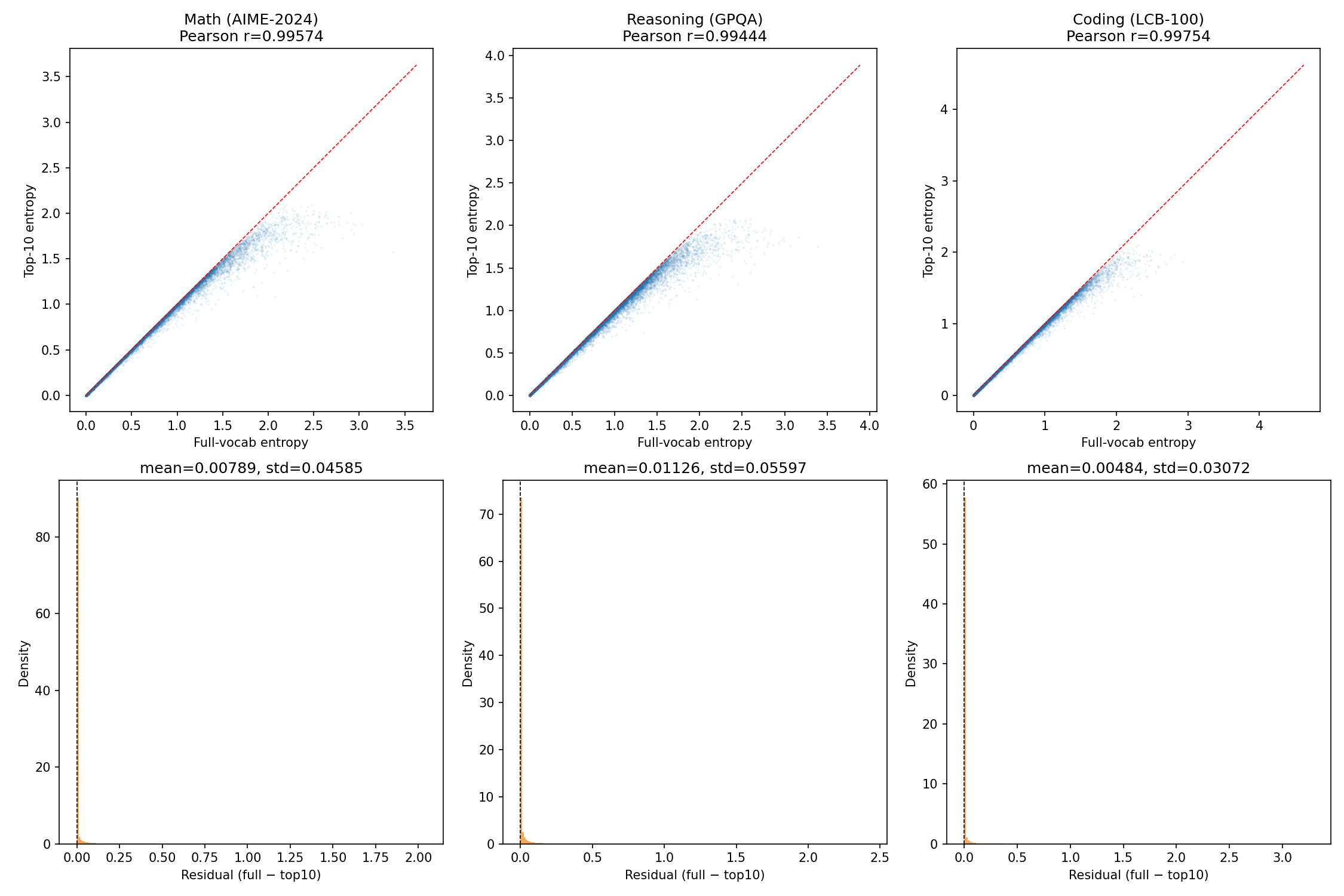}
    \caption{Detailed correlation and residual analysis for certainty (left) and entropy (right). While entropy maintains a near-zero residual, certainty exhibits a distinct, constant positive shift due to the mathematical formulation of the uniform baseline in the truncated KL divergence calculation.}
    \label{fig:entropy_certainty_residuals}
\end{figure}

To validate the computational efficiency of our intrinsic metrics, we analyzed the correlation between statistics computed over the full vocabulary versus a truncated top-10 logit distribution. As detailed in Table~\ref{tab:entropy-correlation}, the top-10 partial entropy exhibits a near-perfect monotonic relationship with the full-vocabulary entropy across mathematical, reasoning, and coding domains (Spearman $\rho = 0.9999$), accompanied by negligible residuals. While the certainty metric maintains a strong linear correlation (Pearson $r > 0.89$), it exhibits a large, constant residual shift (approximately $28.5$) because truncating the vocabulary fundamentally alters the uniform baseline used in the Kullback-Leibler divergence calculation. These findings, further visualized in Figures~\ref{fig:entropy_certainty_scatter} and \ref{fig:entropy_certainty_residuals}, confirm that top-$k$ entropy provides a highly accurate and computationally lightweight proxy for full-vocabulary uncertainty estimation, making it highly suitable for deployment in memory-constrained inference engines without requiring expensive full-distribution marginalization.

\begin{table}[ht!]
\centering
\caption{Correlation between full-vocabulary and top-10 entropy/certainty over the last 2048 tokens per response. Entropy uses Shannon $H(p) = -\sum_v p_v \log p_v$; top-10 computes a partial sum over the 10 highest-probability tokens (without renormalization). Certainty is measured as the Kullback-Leibler divergence from a uniform distribution, $\mathbb{KL}(\mathrm{U} \| p_{\theta})$.}
\label{tab:entropy-correlation}
\begin{tabular}{@{}llccc@{}}
\toprule
{Metric} & {Domain} & {Pearson $r$} & {Spearman $\rho$} & {Residual $\mu \pm \sigma$} \\
\midrule
\multirow{3}{*}{Entropy}
  & Math (AIME-2024)   & 0.9957 & 0.9999 & $0.008 \pm 0.046$ \\
  & Reasoning (GPQA)   & 0.9944 & 0.9999 & $0.011 \pm 0.056$ \\
  & Coding (LCB-100)   & 0.9975 & 0.9999 & $0.005 \pm 0.031$ \\
\midrule
\multirow{3}{*}{Certainty}
  & Math (AIME-2024)   & 0.9366 & 0.9444 & $28.56 \pm 3.51$ \\
  & Reasoning (GPQA)   & 0.8998 & 0.9044 & $28.30 \pm 3.13$ \\
  & Coding (LCB-100)   & 0.9118 & 0.9163 & $28.90 \pm 3.41$ \\
\bottomrule
\end{tabular}
\end{table}

\clearpage

\section{Qualitative Analysis of Token-Level Metrics}
\label{appx:qualitative}

To better understand the dynamics of our intrinsic metrics, we provide a qualitative visualization of per-token certainty and entropy across varying tail and smoothing window sizes. As the figures demonstrate, generating token-level metrics at a very small, localized scale (e.g., 10 to 50 tokens) primarily captures high-frequency syntactic noise, heavily reflecting immediate formatting choices or punctuation. However, as the tail and smoothing windows expand (scaling up to 5000 tokens), this localized variance is filtered out, revealing the macro-level trends in the model's generation confidence. These larger temporal views isolate the broader semantic certainty of the reasoning trajectory. Ultimately, these visualizations underscore the necessity of our adaptive tail formulation: relying on excessively small or rigid windows risks overfitting to localized entropy spikes, whereas appropriate smoothing over a dynamically scaled tail isolates the genuine discriminative signal necessary for robust inference scaling.

\begin{figure}[ht!]
    \centering
    \includegraphics[width=0.7\linewidth]{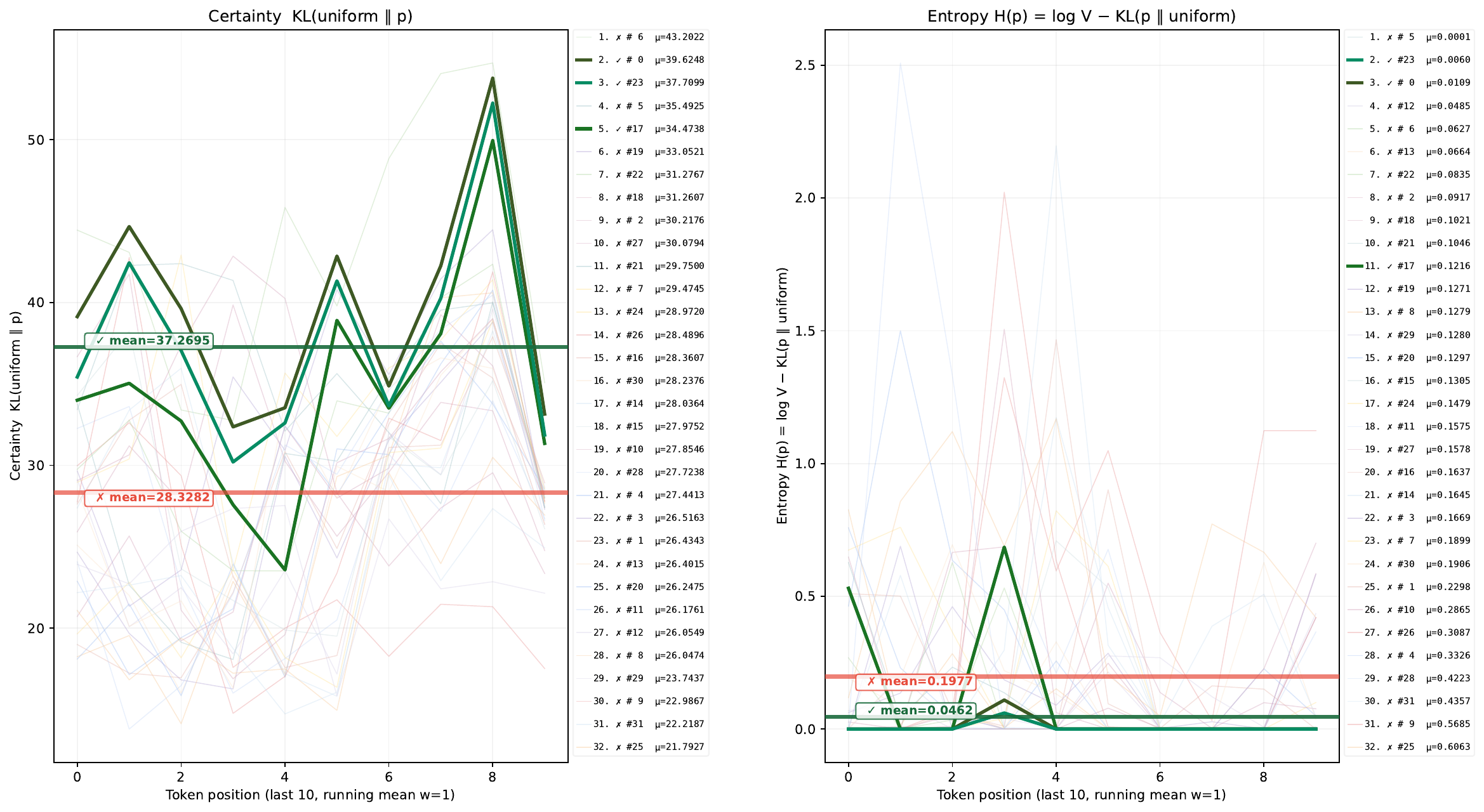}
    \includegraphics[width=0.7\linewidth]{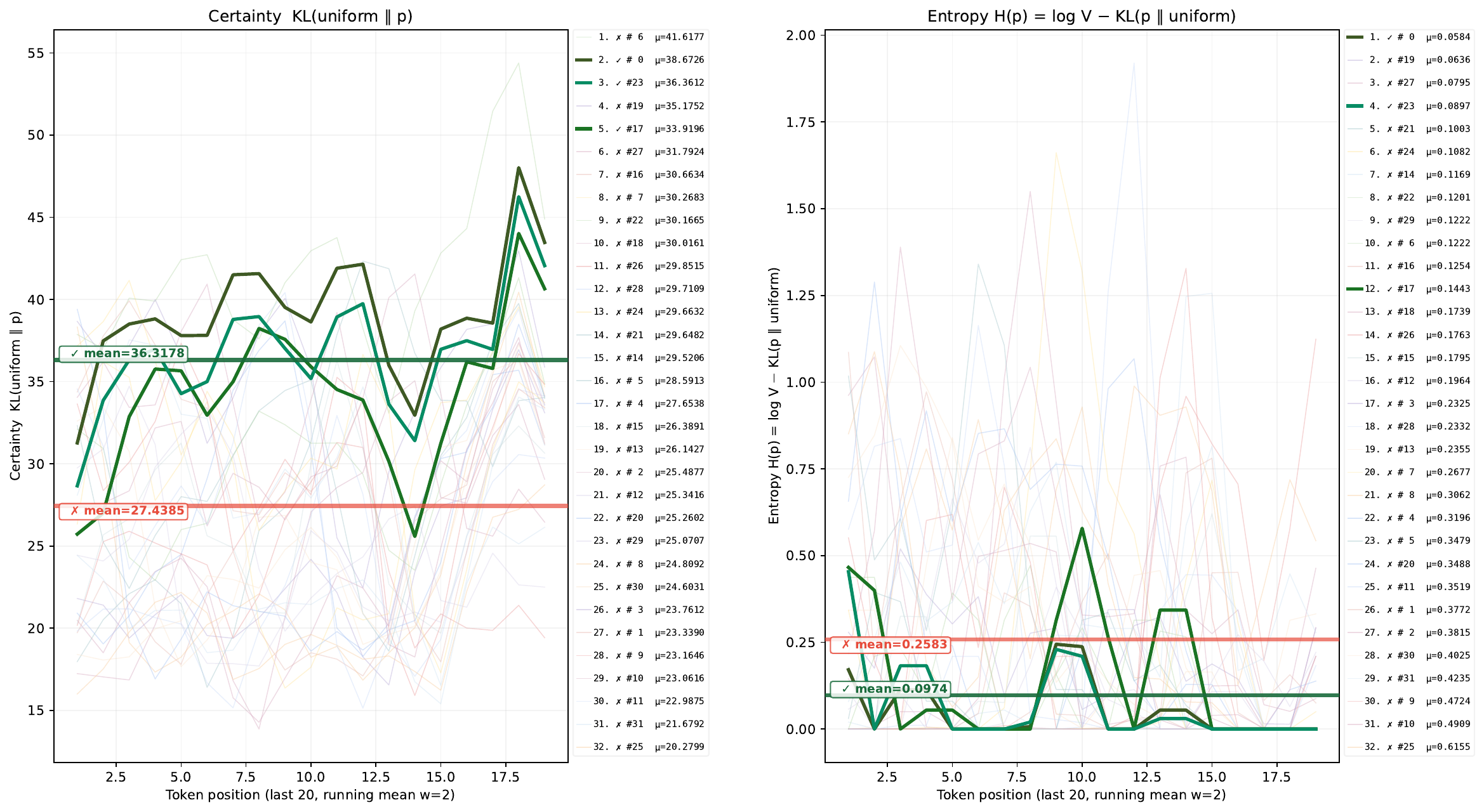}
    \includegraphics[width=0.7\linewidth]{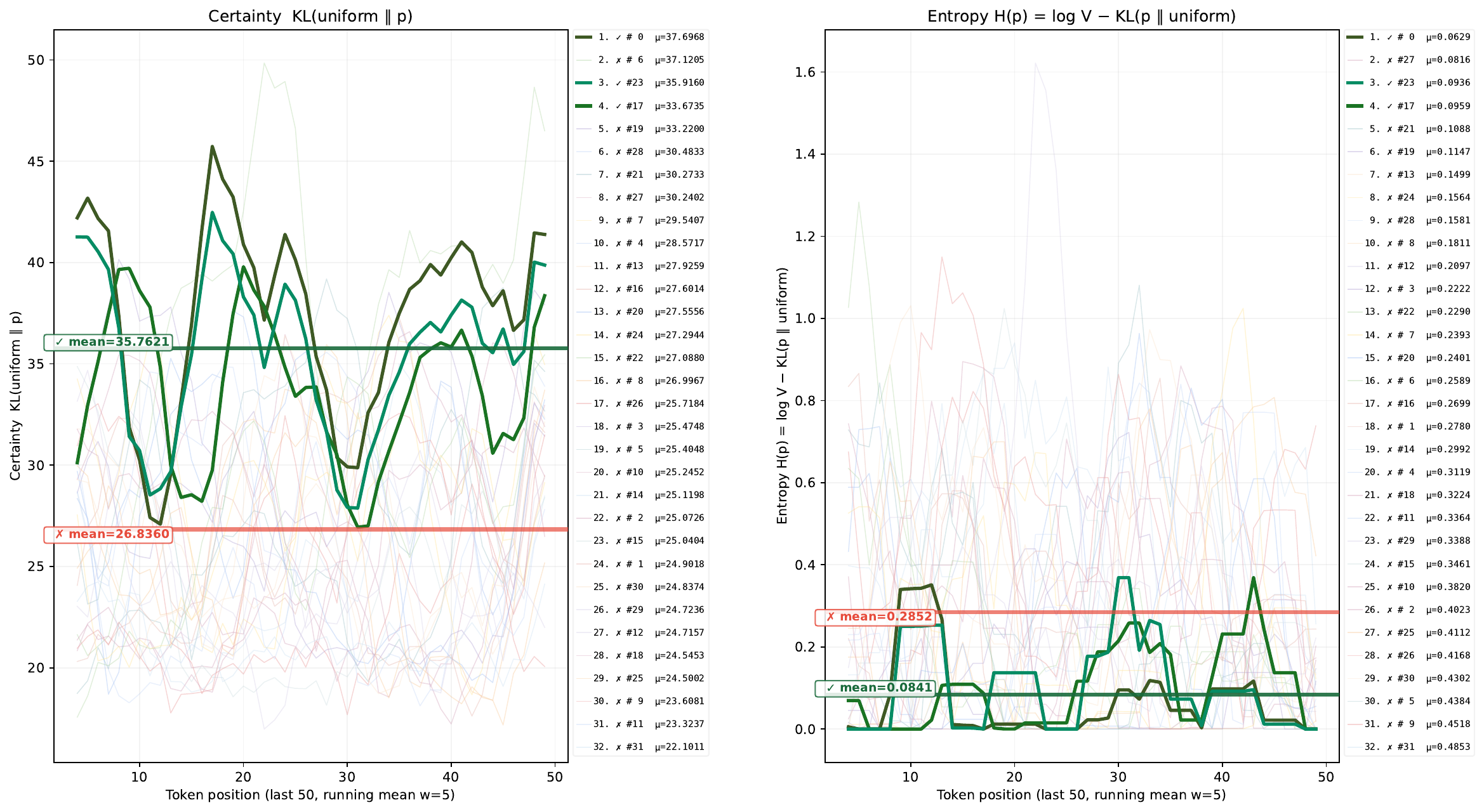}
    \caption{Per-token certainty vs.\ entropy for small tail windows (10, 20, and 50 tokens) with localized smoothing (windows of 1, 2, and 5). At this fine-grained resolution, the metrics predominantly capture high-frequency syntactic variations and immediate token-level uncertainty.}
     \label{fig:certainty-entropy-small}
\end{figure}

\begin{figure}[ht!]
    \centering
    \includegraphics[width=0.7\linewidth]{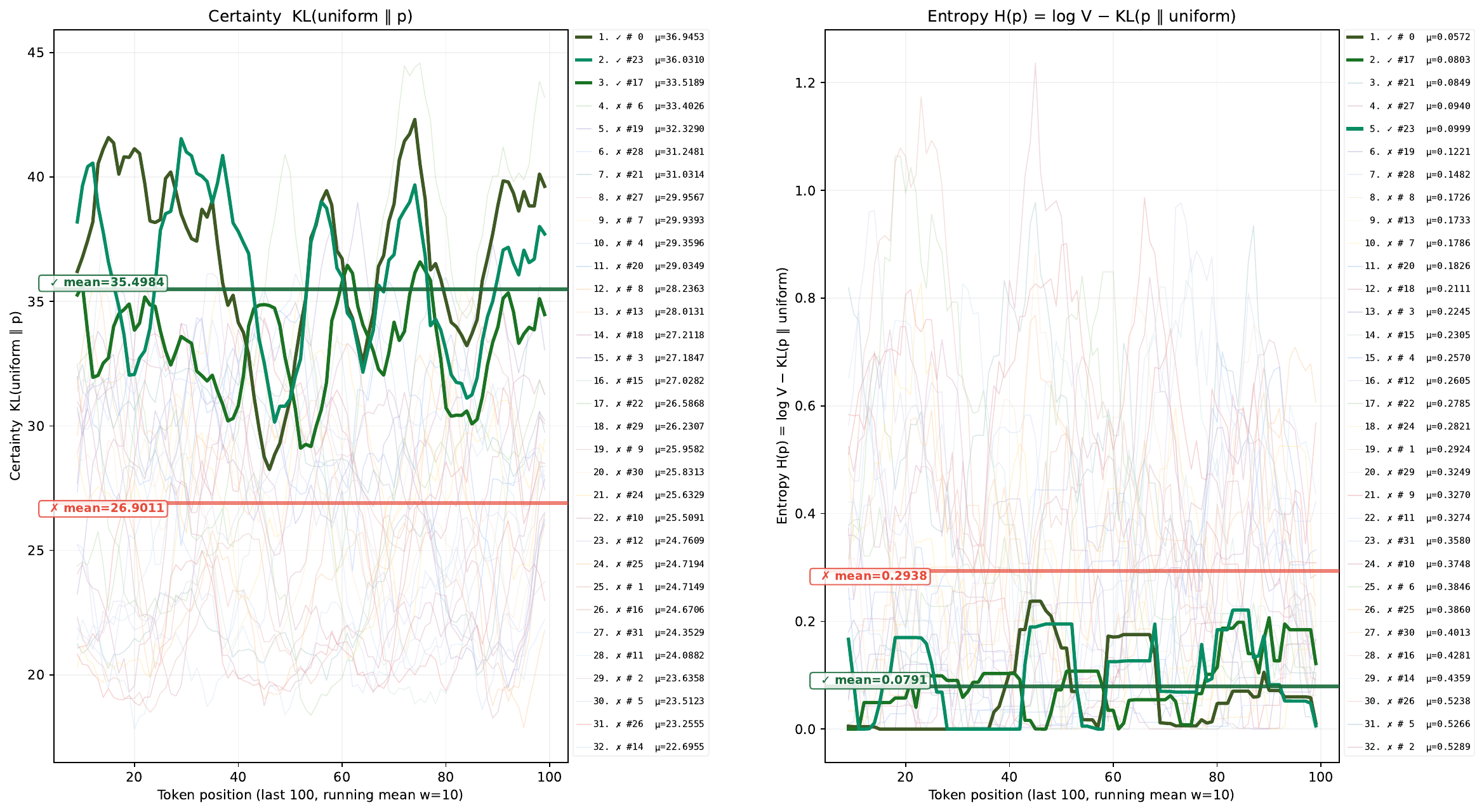}
    \includegraphics[width=0.7\linewidth]{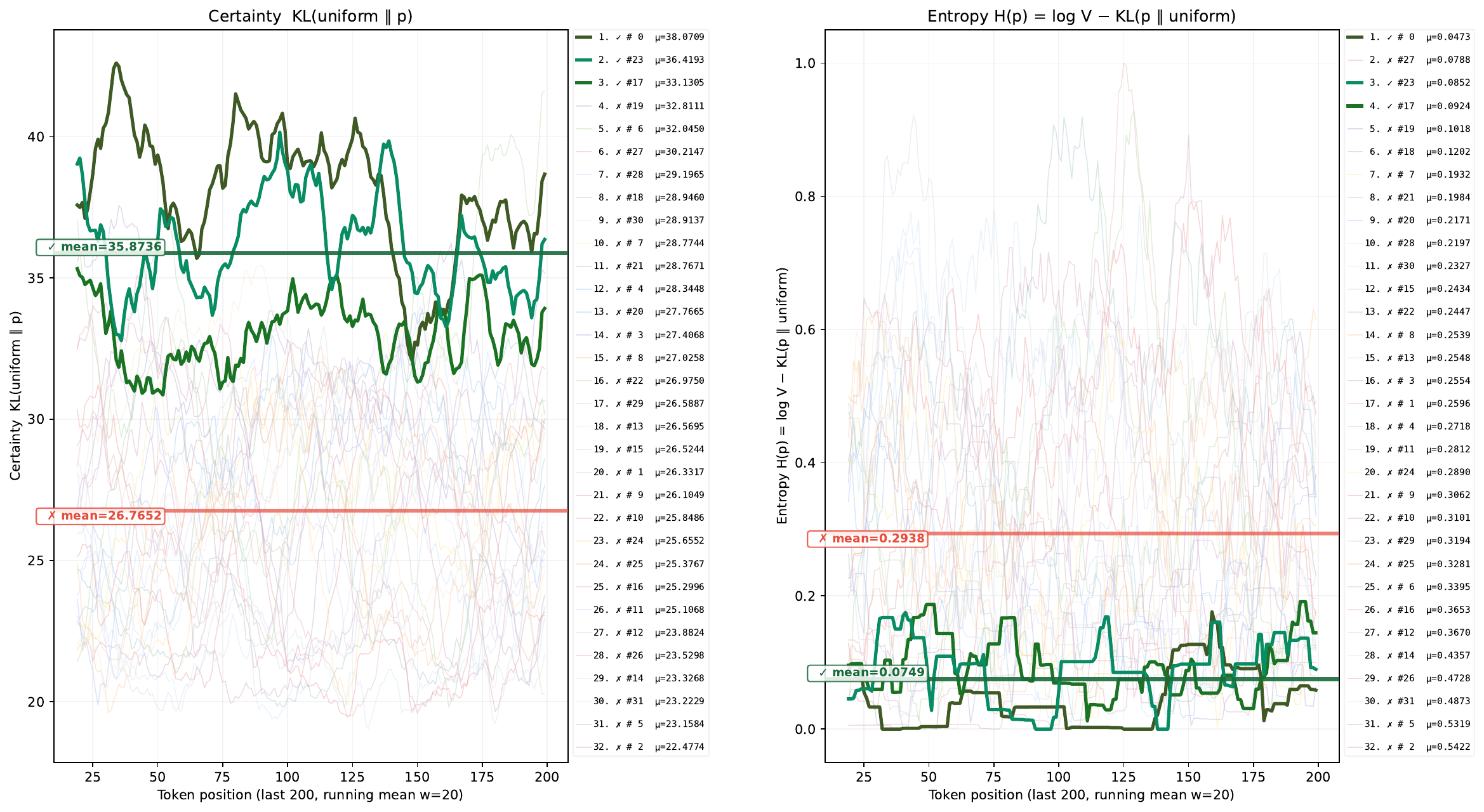}
    \includegraphics[width=0.7\linewidth]{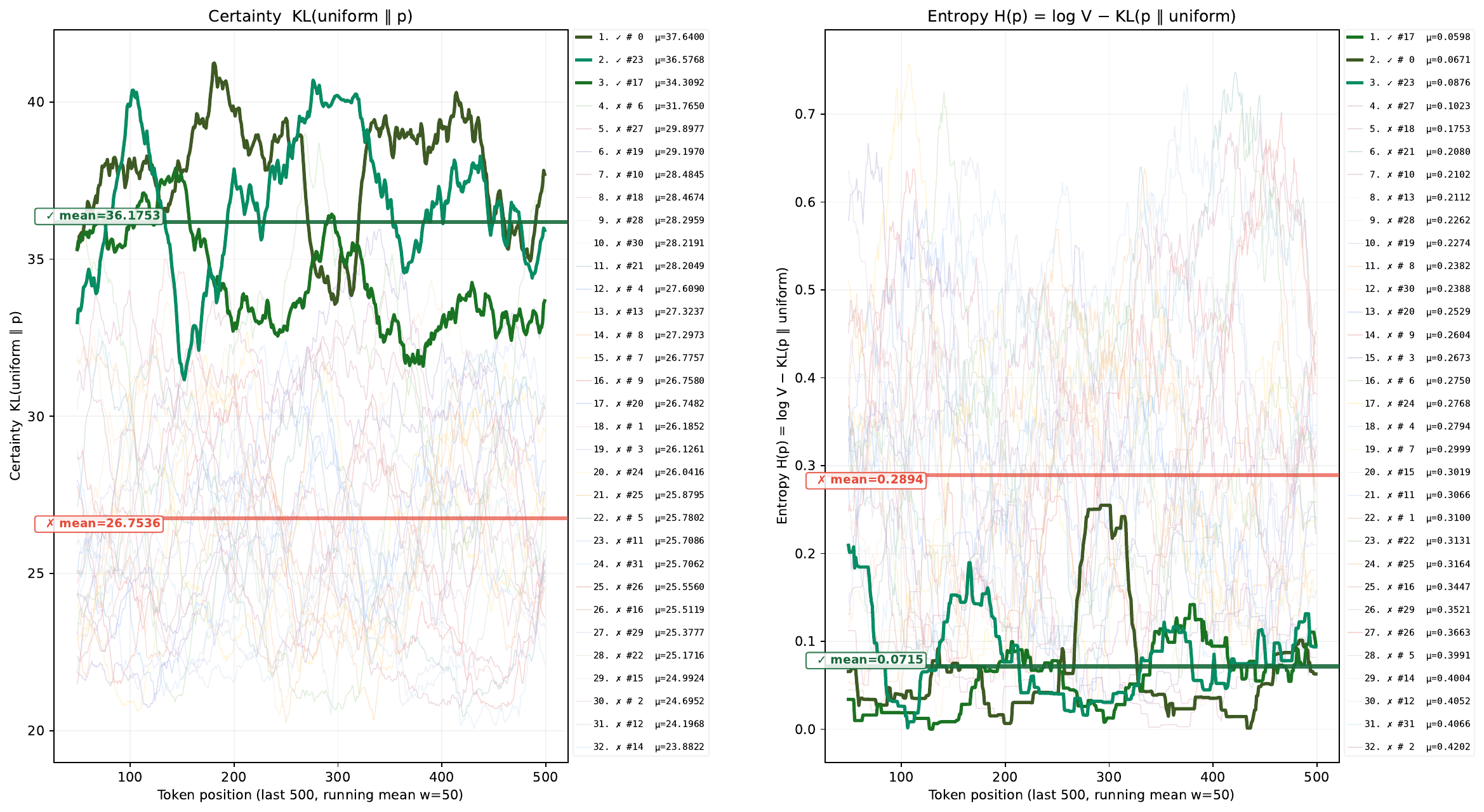}
    \caption{Per-token certainty vs.\ entropy for medium tail windows (100, 200, and 500 tokens) with moderate smoothing (windows of 10, 20, and 50). This intermediate scale begins to filter out localized punctuation noise, revealing structural confidence trends over individual reasoning steps.}
     \label{fig:certainty-entropy-medium}
\end{figure}

\begin{figure}[ht!]
    \centering
    \includegraphics[width=0.7\linewidth]{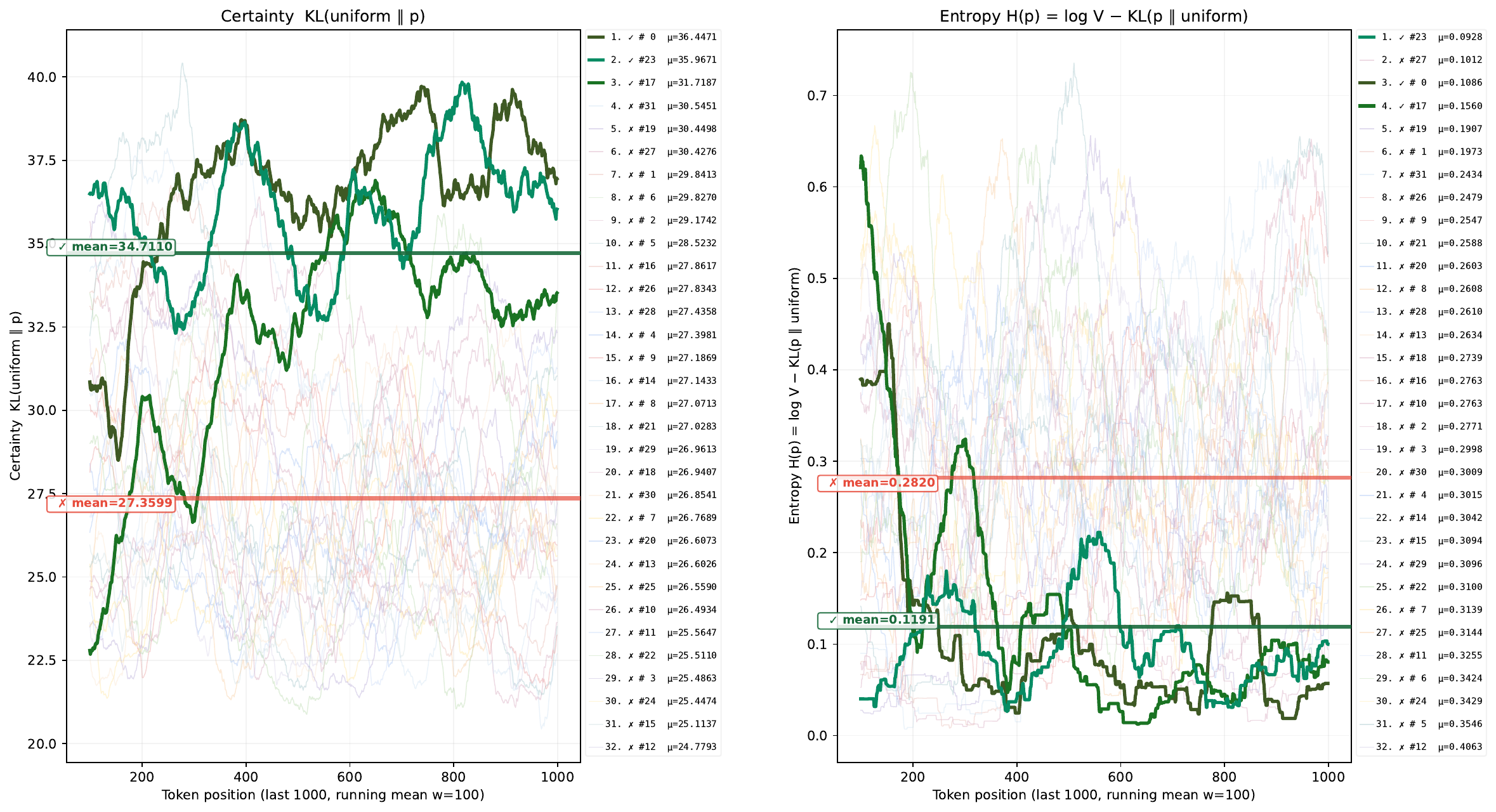}
    \includegraphics[width=0.7\linewidth]{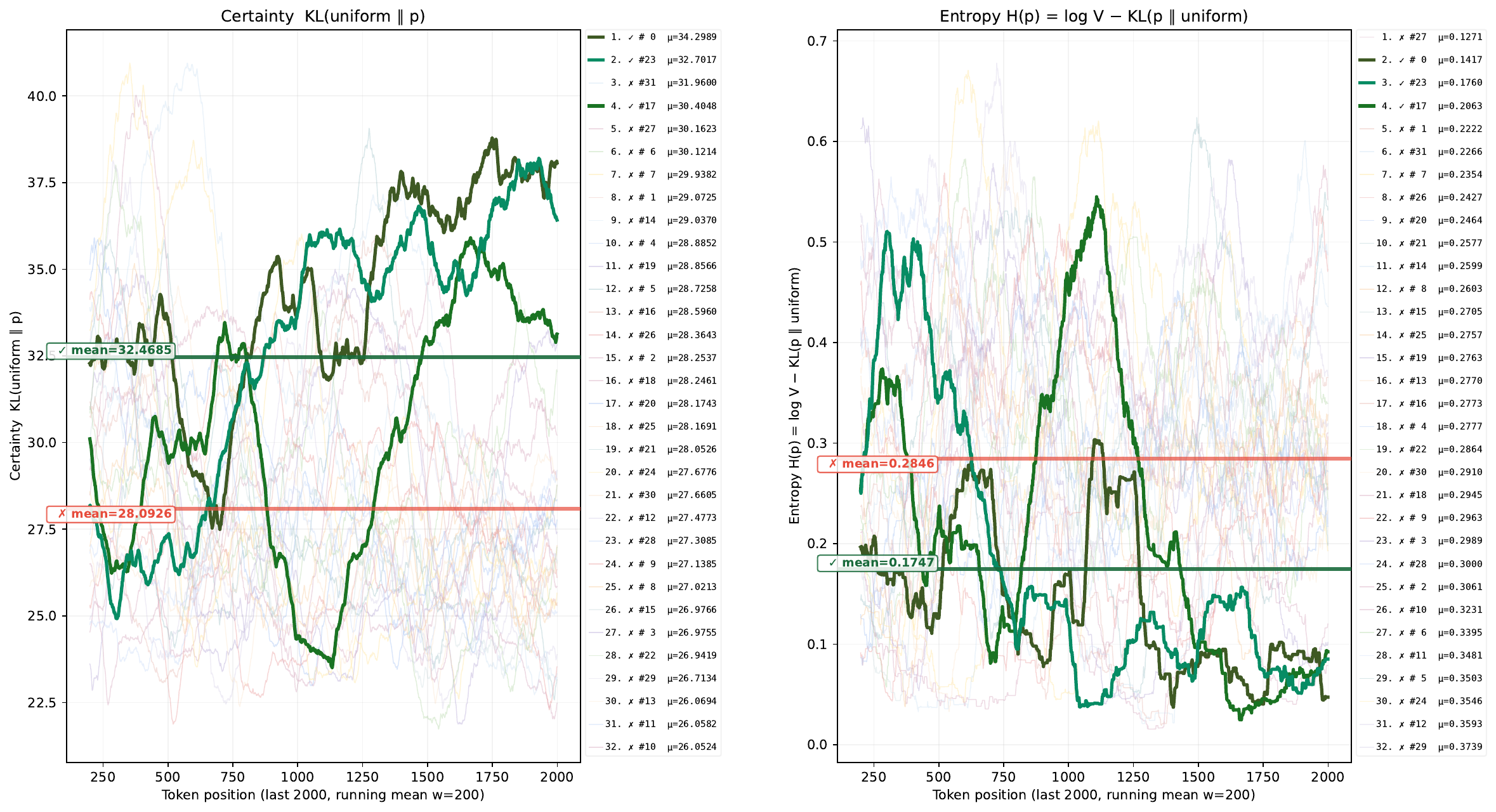}
    \includegraphics[width=0.7\linewidth]{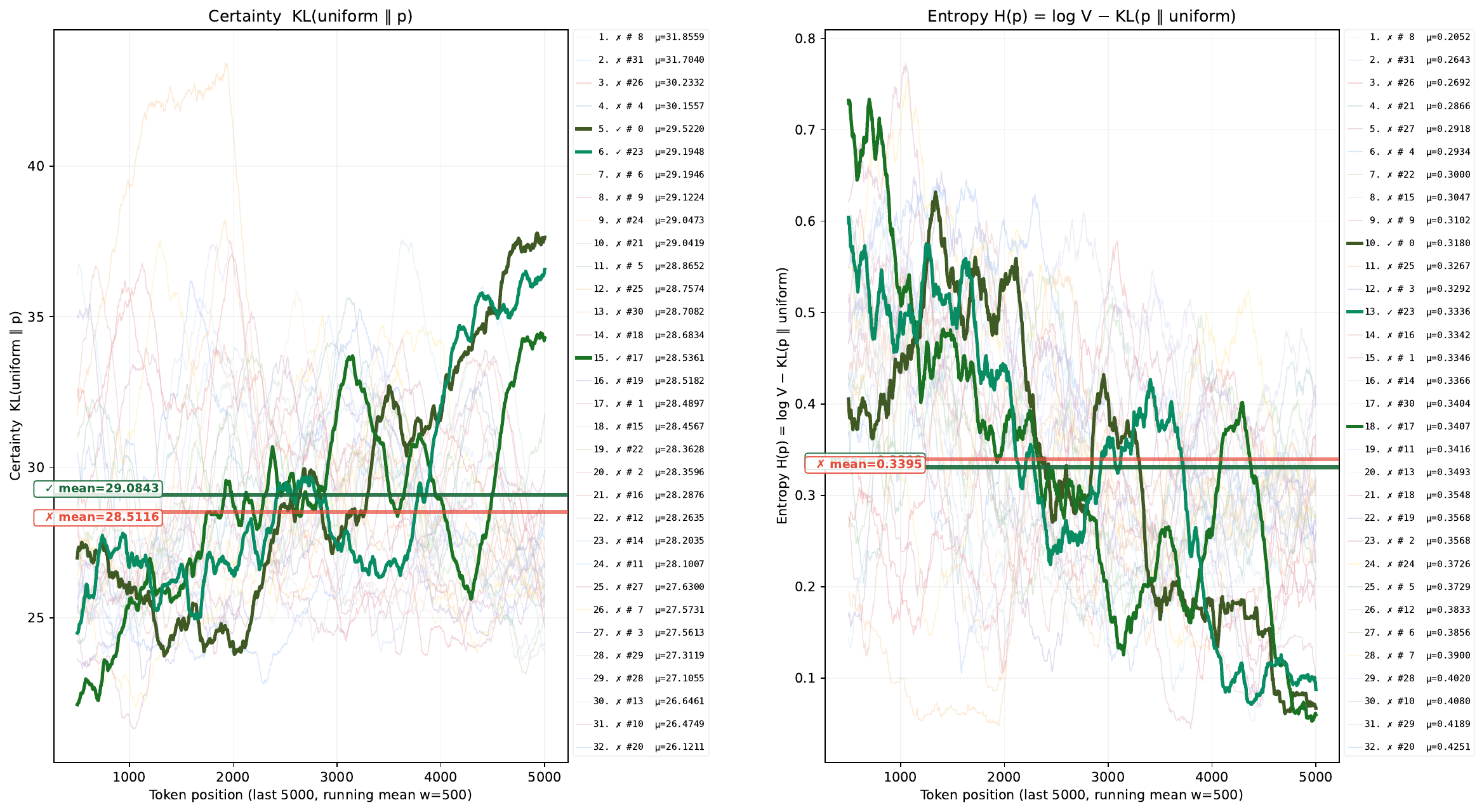}
    \caption{Per-token certainty vs.\ entropy for large tail windows (1000, 2000, and 5000 tokens) with broad smoothing (windows of 100, 200, and 500). At this macro scale, high-frequency noise is entirely smoothed out, clearly illustrating the model's overarching semantic confidence throughout the entire generated trajectory.}
     \label{fig:certainty-entropy-large}
\end{figure}

\clearpage

\section{Entropy and KL Guidance Visualizations}
\label{appx:visualizations}

To provide further intuition into the dynamics of our generation pipeline, we visualize the progression of Kullback-Leibler (KL) divergence across several challenging AIME 2024 problems. These granular trace plots illustrate how the model's internal predictive distributions evolve step-by-step during complex mathematical reasoning. By tracking the KL divergence throughout the sequence, we can observe critical transition points - such as when the model commits to a specific logical pathway, encounters structural uncertainty, or successfully aligns with the privileged guidance during particle distillation. These problem-specific profiles visually reinforce the highly dynamic nature of the problem-solving process and highlight exactly where step-level interventions are most impactful for navigating hard-to-verify domains.

\begin{figure}[ht!]
    \centering
    \includegraphics[width=0.8\linewidth]{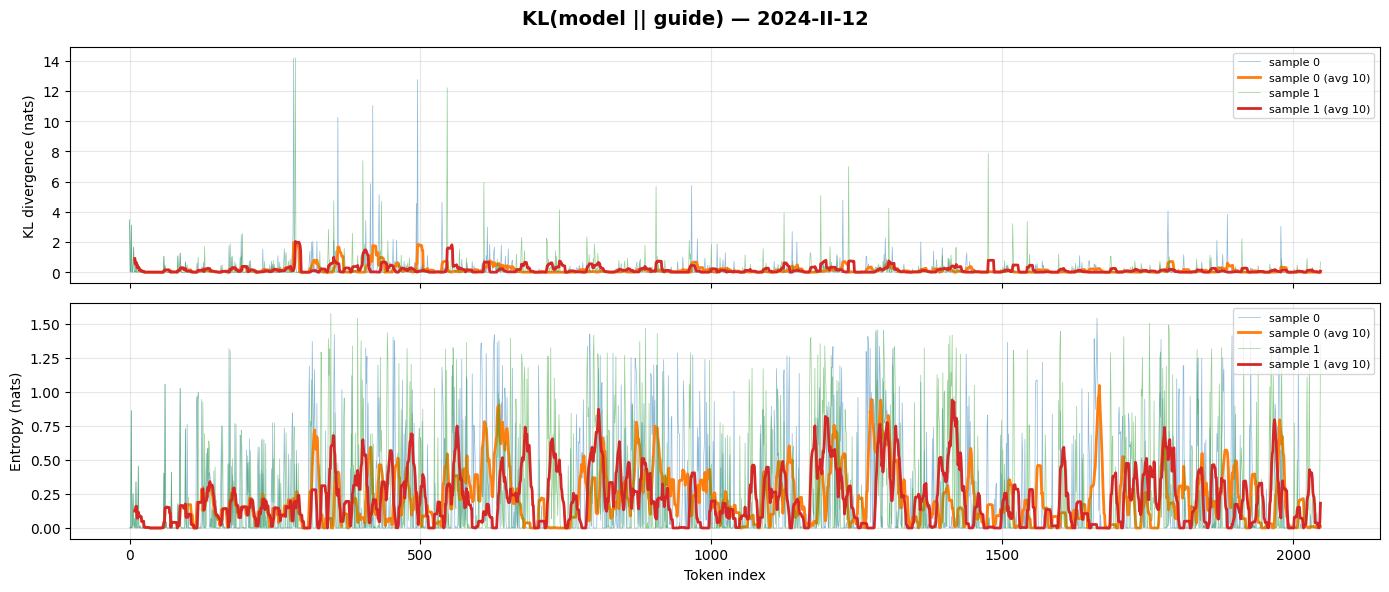}
    \includegraphics[width=0.8\linewidth]{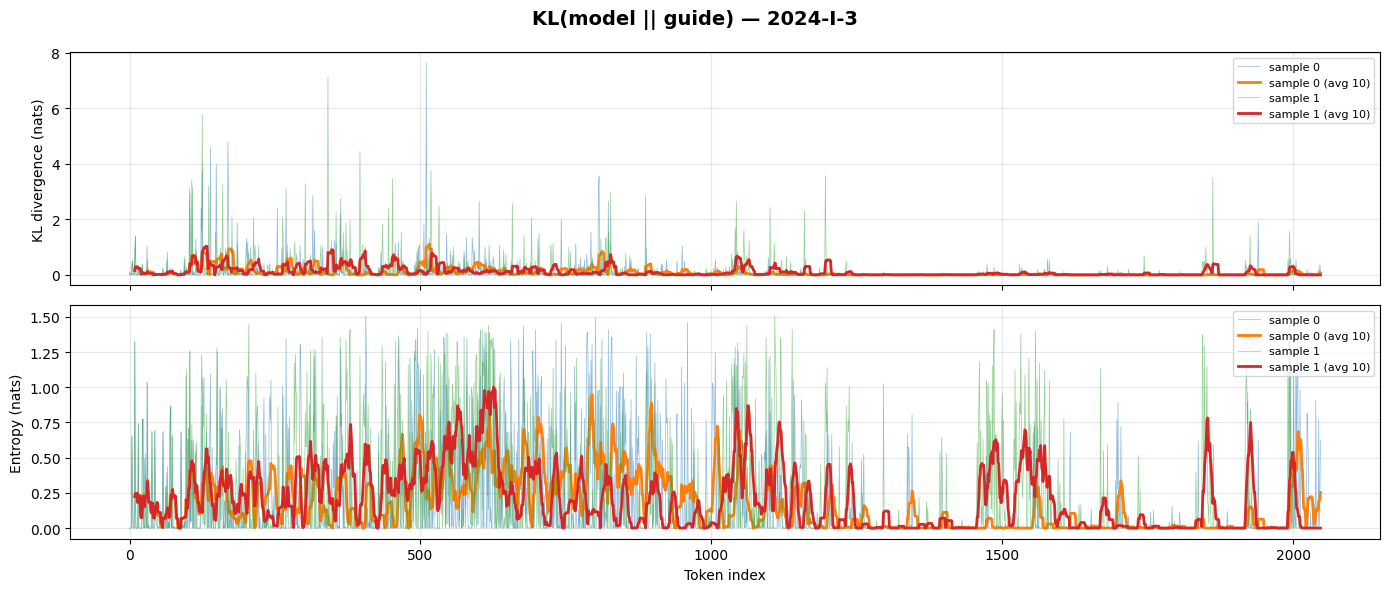}
    \caption{Temporal visualization of KL divergence for AIME 2024 problems II-12 (top) and I-3 (bottom). The traces reveal localized spikes in divergence corresponding to specific trajectory steps that require significant mathematical deduction or structural transitions.}
     \label{fig:certainty-entropy-kl-1}
\end{figure}

\begin{figure}[ht!]
    \centering
    \includegraphics[width=0.8\linewidth]{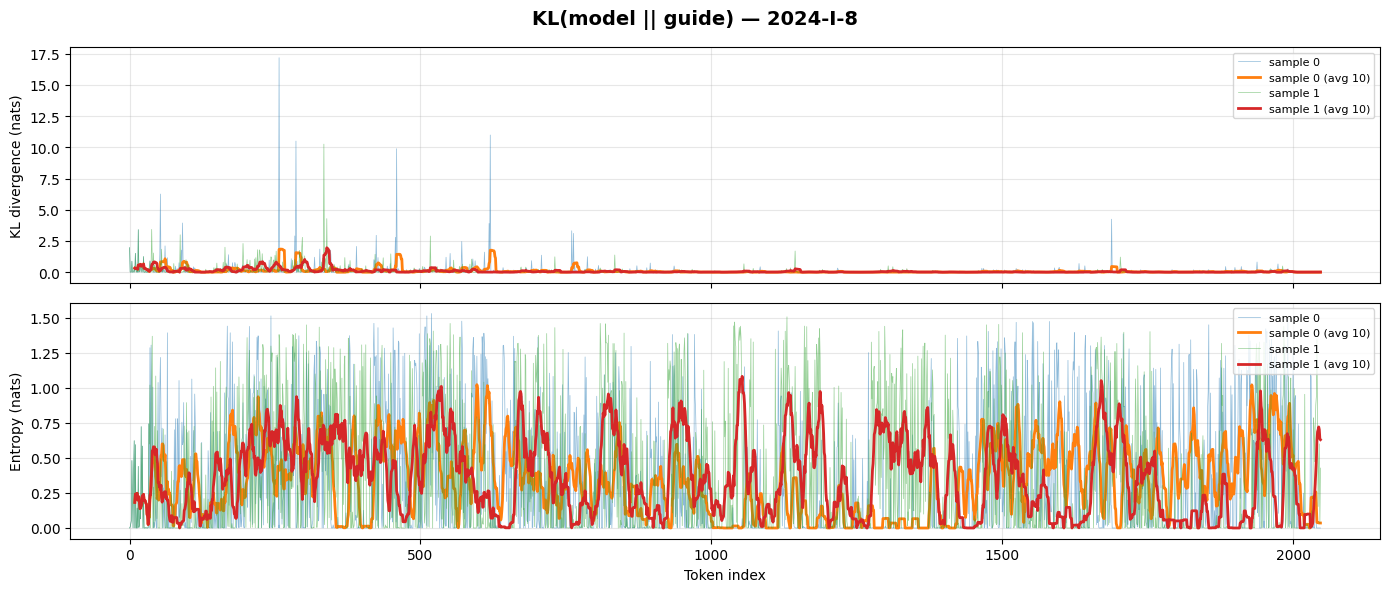}
    \includegraphics[width=0.8\linewidth]{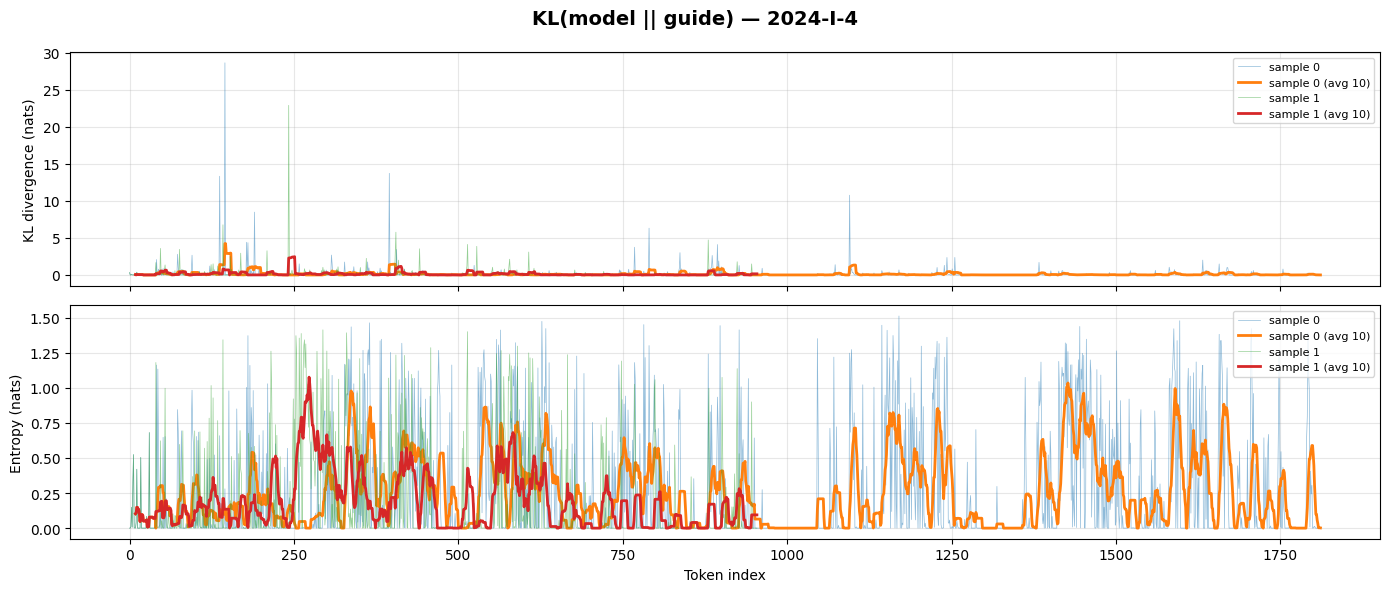}
    \includegraphics[width=0.8\linewidth]{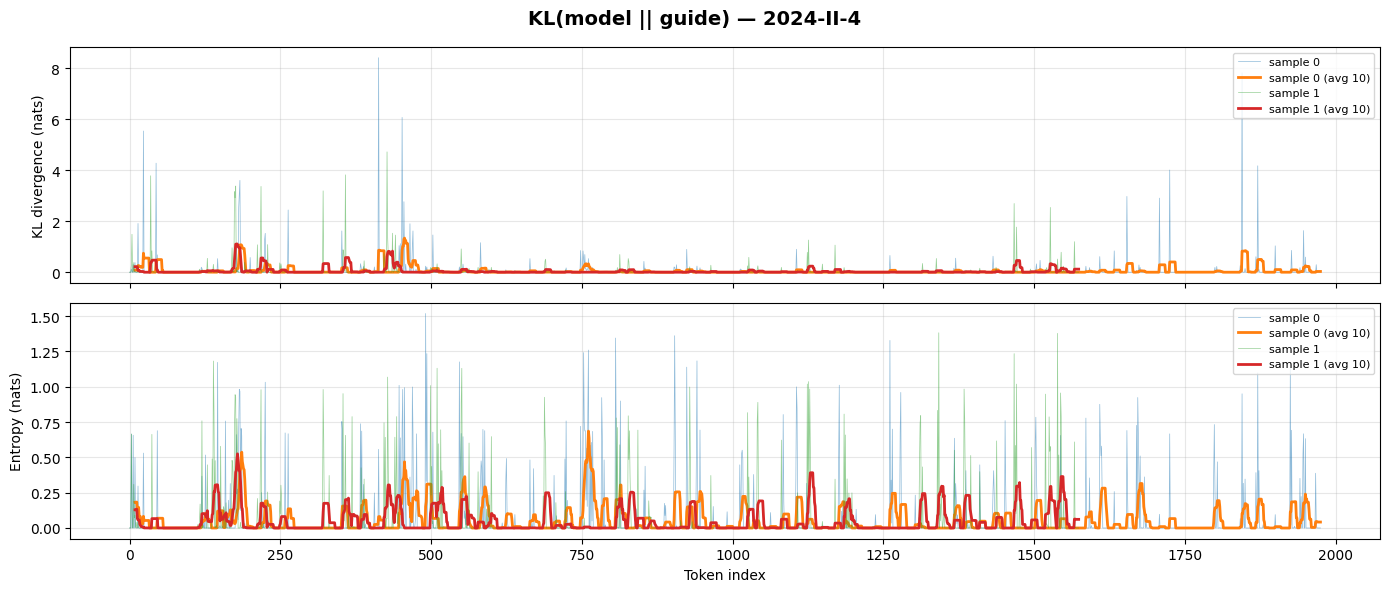}
    \caption{Temporal visualization of KL divergence for AIME 2024 problems I-8 (top), I-4 (middle), and II-4 (bottom). These problem-level profiles demonstrate how the model navigates prolonged reasoning phases, with sustained periods of stable divergence indicating confident algorithmic execution.}
     \label{fig:certainty-entropy-kl-2}
\end{figure}

\clearpage

\section{Examples}

\subsection{Hard Math Problems Examples}

\subsection*{Summary}

\begin{table}[h]
\centering
\caption{Parallel iid Sampling vs.\ iPF Sampling selection results across 7 examples.
Correct selections are \textbf{bolded}.}
\label{tab:summary}
\resizebox{\textwidth}{!}{%
\begin{tabular}{clllccccccc}
\toprule
ID & Dataset & Problem & GT
   & iid correct/16 & iPF correct/16
   & iid SC & iPF SC
   & iid iS & iPF iS \\
\midrule
1 & AIME 2025 & Sawtooth piecewise function          & 259
  & 2/16  & \textbf{11/16}
  & 23    & \textbf{259}
  & 275   & \textbf{259} \\
2 & AIME 2025 & Triangle circumcircle angles          & 336
  & 1/16  & \textbf{9/16}
  & 360   & \textbf{336}
  & 840   & \textbf{336} \\
3 & AIME 2026 & Binomial coefficient sums mod 502     & 39
  & 2/16  & \textbf{10/16}
  & 0     & \textbf{39}
  & 0     & \textbf{39} \\
4 & AIME 2026 & Same problem (seed 1)                 & 39
  & 2/16  & \textbf{14/16}
  & 501   & \textbf{39}
  & 501   & \textbf{39} \\
5 & AIME 2026 & Permutations with $\pi^6 = \mathrm{id}$ & 396
  & 4/16  & \textbf{15/16}
  & 1     & \textbf{396}
  & \textbf{396} & \textbf{396} \\
6 & AIME 2026 & Binomial sums (seed 3)                & 39
  & 2/16  & \textbf{13/16}
  & 0     & \textbf{39}
  & 500   & \textbf{39} \\
7 & AIME 2026 & Die sticker placement                 & 29
  & 1/16  & \textbf{7/16}
  & 13    & \textbf{29}
  & \textbf{29} & \textbf{29} \\
\bottomrule
\end{tabular}
}
\end{table}

\subsection*{The PF Mechanism}

\textbf{Parallel Sampling.} Answers are scattered. The correct answer appears only 1--4
times out of 16 i.i.d.\ samples. Self-consistency (SC) picks the wrong majority
(a frequent distractor), and entropy selects a response that appears confident
but is ultimately wrong.

\textbf{iPF Sampling.} The particle filter's (PF) resampling step amplifies the
correct answer from a minority to a majority. The correct answer appears 7--15
times out of 16 particles. Both SC and entropy now select correctly, because the
answer pool has converged.

The PF does not discover new solutions; rather, it identifies which partial
solutions are more promising---via entropy scoring during generation---and
allocates more particles to those trajectories.

\subsection*{Detailed Examples}

\subsubsection*{Example~1: AIME 2025~\#11 --- Sawtooth Piecewise Function
(Seed~1)}

\textbf{Problem:} A piecewise linear function $f(x)$ with period~4; find the
sum involving quadratic intersections.\\
\textbf{Ground truth:} 259

\textbf{Parallel Sampling} (2/16 correct):
\begin{verbatim}
Answer distribution: {'23': 2, '0': 1, '103': 1, '44': 1, '20': 1, '22': 1,
                       '275': 1, '259': 1, '343': 1, '4801': 1, ...}
SC picks:      '23'  (2 votes, wrong)  -- answers nearly uniform; SC is random
Entropy picks: '275' (idx=8, score=0.093) -- confident but wrong
\end{verbatim}

\textbf{iPF Sampling} (11/16 correct):
\begin{verbatim}
Answer distribution: {'259': 11, '23': 1, '17': 1, '114': 1, None: 1, ...}
SC picks:      '259' (11 votes, correct) -- PF converged heavily on correct answer
Entropy picks: '259' (idx=7, score=0.107) -- correct; also low entropy
\end{verbatim}

\textbf{Why it works.} In Parallel Sampling, the model occasionally solves the problem
(2/16), but 16 distinct distractors split the vote. The PF identifies the
correct approach early---via lower intermediate entropy---and resamples toward
it, producing 11 particles with the right answer. SC now has a clear majority.

\subsubsection*{Example~3: AIME 2026~\#13 --- Binomial Coefficient Sums
(Seed~0)}

\textbf{Problem:} Compute the sum of binomial coefficients $\binom{10000}{502n+r}$
and find the number of distinct values.\\
\textbf{Ground truth:} 39

\textbf{Parallel Sampling} (2/16 correct):
\begin{verbatim}
Answer distribution: {'0': 6, '501': 5, '39': 2, '1': 1, '2': 1, '251': 1}
SC picks:      '0' (6 votes, wrong)   -- dominant distractor
Entropy picks: '0' (idx=0, score=0.187) -- also picks the distractor
\end{verbatim}

\textbf{iPF Sampling} (10/16 correct):
\begin{verbatim}
Answer distribution: {'39': 10, '0': 4, '502': 1, '501': 1}
SC picks:      '39' (10 votes, correct) -- PF suppressed distractors
Entropy picks: '39' (idx=7, score=0.075) -- much lower entropy than Phase 1
\end{verbatim}

\textbf{Why it works.} In Parallel Sampling, the distractor \texttt{`0'} dominates,
because the model frequently---and incorrectly---concludes that all $S_r$ are
equal. The PF recognizes that particles converging on \texttt{`39'} exhibit
lower intermediate entropy (the correct reasoning path is more internally
consistent) and resamples toward them. By the end, 10/16 particles carry the
correct answer, overwhelming the \texttt{`0'} distractor.

\subsubsection*{Example~5: AIME 2026~\#7 --- Permutations with
$\pi^6 = \mathrm{id}$}

\textbf{Problem:} Count the permutations $\pi$ of $\{1,\ldots,6\}$ satisfying
$\pi^6 = \mathrm{id}$.\\
\textbf{Ground truth:} 396

\textbf{Parallel Sampling} (4/16 correct):
\begin{verbatim}
Answer distribution: {'1': 12, '396': 4}
SC picks:      '1'   (12 votes, wrong) -- strong distractor
Entropy picks: '396' (idx=15, score=0.009) -- CORRECT (very low entropy)
\end{verbatim}

\textbf{iPF Sampling} (15/16 correct):
\begin{verbatim}
Answer distribution: {'396': 15, None: 1}
SC picks:      '396' (15 votes, correct) -- near-total convergence
Entropy picks: '396' (idx=11, score=0.036) -- correct
\end{verbatim}

\textbf{Why it works.} This example is notable because Parallel Sampling entropy
\emph{also} selects correctly: the correct solution has an extremely low entropy
of $0.009$, compared to the distractor. However, SC fails in Parallel Sampling because
\texttt{`1'} receives 12 votes. iPF Sampling reverses this: the PF amplifies the
\texttt{`396'} trajectories to 15/16, so SC now agrees with entropy. This
illustrates a characteristic SC failure mode---a confidently wrong answer
overwhelming a rarer but more certain correct answer---that the PF resolves.

\subsubsection*{Example~7: AIME 2026~\#9 --- Die Sticker Placement}

\textbf{Problem:} Roll a die and place stickers labeled 1--6 on its faces; find
the numerator of the expected value.\\
\textbf{Ground truth:} 29

\textbf{Parallel Sampling} (1/16 correct):
\begin{verbatim}
Answer distribution: {'13': 7, '3': 2, '29': 1, '7': 1, '47': 1, ...}
SC picks:      '13' (7 votes, wrong) -- dominant distractor
Entropy picks: '29' (idx=1, score=0.152) -- CORRECT
\end{verbatim}

\textbf{iPF Sampling} (7/16 correct):
\begin{verbatim}
Answer distribution: {'29': 7, '304': 5, '13': 1, '149': 1, None: 2}
SC picks:      '29' (7 votes, correct) -- PF amplified the correct answer
Entropy picks: '29' (idx=12, score=0.092) -- correct; lower entropy than iid
\end{verbatim}

\textbf{Why it works.} As in Example~5, Parallel Sampling entropy already identifies
the correct answer, while SC fails due to a strong distractor. The PF amplifies
\texttt{`29'} from 1/16 to 7/16, establishing it as the SC majority. Note that
a new distractor \texttt{`304'} emerges in iPF Sampling with 5 votes: the PF does
not eliminate all wrong answers, but it shifts the balance sufficiently for SC
to succeed.

\subsection*{Key Observations}

\textbf{Amplification.}
In 5 of 7 cases, the correct answer already appears somewhere within the
parallel samples.
The particle filter does not unlock previously unsolvable problems---it makes
solvable problems \emph{reliably} solved, amplifying correct particles from
1--4 copies up to 7--15.

\textbf{Self-consistency is the primary beneficiary.}
iPF sampling with SC achieves the correct answer in all 7 examples.
The particle filter's main role is to correct SC failures: by amplifying the
correct answer to majority status, it resolves cases where the wrong answer
would otherwise dominate the vote.

\textbf{Entropy is an early signal.}
In 2 of 7 examples (5 and~7), parallel sampling entropy already identifies the
correct answer while SC fails---the correct solution carries a distinctively low
entropy score.
The particle filter then brings SC into agreement by amplifying that
low-entropy solution to majority status.

\clearpage
\subsection{HealthBench Examples}

\subsubsection*{Health Data Tasks: Outpatient Note Drafting ($+0.478$)}
\begin{itemize}
    \item \textbf{UID}: \texttt{ce231145-1911-4245-b990-85ca8fbdb7ef}
    \item \textbf{pass@1}: $0.214$ \textrightarrow{} \textbf{iPF}: $0.692$
    \item \textbf{Selection}: Particle 2 of 8, HF entropy 10\% tail median
    \item \textbf{Why it works}: The entropy-selected particle adheres to formal note formatting, provides a complete template with bracketed fields for unknown values, and includes a concise clinical summary. IID samples frequently fail to satisfy these structural requirements.
    \item \textbf{Criteria met}: Formal note format, complete template with brackets, routine check-up summary
\end{itemize}

\subsubsection*{Context Seeking: Sleep Aid Recommendation ($+0.424$)}
\begin{itemize}
    \item \textbf{UID}: \texttt{a20cbbf3-330b-46d9-a35b-92e89a022282}
    \item \textbf{pass@1}: $0.147$ \textrightarrow{} \textbf{iPF}: $0.571$
    \item \textbf{Selection}: Particle 2 of 8
    \item \textbf{Why it works}: The entropy-selected response recommends specific OTC options (e.g., diphenhydramine) with dosing information and side effects, and supplements these with sleep hygiene tips---covering the rubric's informational requirements that most IID samples omit.
    \item \textbf{Criteria met}: Safe OTC sleep aid suggestions with doses and side effects, sleep hygiene tips
\end{itemize}

\subsubsection*{Health Data Tasks: Email Rewriting ($+0.413$)}
\begin{itemize}
    \item \textbf{UID}: \texttt{fb585290-da30-4fd4-b2d6-31776aad9f1b}
    \item \textbf{pass@1}: $0.087$ \textrightarrow{} \textbf{iPF}: $0.500$
    \item \textbf{Selection}: Particle 3 of 8
    \item \textbf{Why it works}: This task requires rewriting a medical email with specific components, including antibiotic timing, dosing, and diagnosis. The entropy-selected particle includes all required components, whereas most IID samples omit the antibiotic timing or dosage details.
\end{itemize}

\subsubsection*{Postpartum Haemorrhage Risk Assessment ($+0.345$)}
\begin{itemize}
    \item \textbf{UID}: \texttt{d24b5c35-b04e-4cb4-98f7-d05891dc63d7}
    \item \textbf{pass@1}: $0.095$ \textrightarrow{} \textbf{dPF}: $0.439$
    \item \textbf{Selection}: Particle 0 of 16, KL 10\% tail median
    \item \textbf{Why it works}: The rubric requires a qualitative risk assessment aligned with clinical guidelines for postpartum haemorrhage management. The KL-selected particle provides a structured risk framework with explicit guideline references---a direct result of the rubric-conditioned guide steering generation toward guideline-aligned content.
    \item \textbf{Criteria met}: Qualitative risk assessment, factual accuracy, clear core answer
\end{itemize}

\subsubsection*{Lithium Drug Interactions ($+0.159$)}
\begin{itemize}
    \item \textbf{UID}: \texttt{bddd8168-f229-487f-bae7-215331e4330b}
    \item \textbf{pass@1}: $0.670$ \textrightarrow{} \textbf{dPF}: $0.829$
    \item \textbf{Selection}: Particle 8 of 16
    \item \textbf{Why it works}: This complex pharmacology question requires specific details about ACE inhibitor and diuretic interactions with lithium. The KL-selected response addresses pharmacokinetics, monitoring protocols, and toxicity signs---highly specific rubric criteria that the rubric-conditioned guide helps precisely target.
    \item \textbf{Criteria met}: Co-administration background, pharmacokinetic interaction details, monitoring protocol recommendation
\end{itemize}

\subsubsection*{Heart Failure and T2DM Management ($+0.147$)}
\begin{itemize}
    \item \textbf{UID}: \texttt{e89e00c9-29e9-420c-813d-e06ddc39d585}
    \item \textbf{pass@1}: $0.468$ \textrightarrow{} \textbf{dPF}: $0.615$
    \item \textbf{Selection}: Particle 11 of 16
    \item \textbf{Why it works}: This task requires a dual management strategy for heart failure and T2DM, including specific drug recommendations such as SGLT2 inhibitors. The KL-selected particle correctly classifies HFmrEF and recommends evidence-based, dual-benefit medications---criteria that the rubric guide directly conditions on.
    \item \textbf{Criteria met}: Dual management strategy, correct HFmrEF classification, SGLT2 inhibitor recommendation
\end{itemize}

\subsubsection*{Medical Coding Documentation --- Perfect Score ($+0.469$)}
\begin{itemize}
    \item \textbf{UID}: \texttt{d6280ff4-0f8d-4896-b2db-271c7cb8af20}
    \item \textbf{pass@1}: $0.531$ \textrightarrow{} \textbf{dPF}: $1.000$
    \item \textbf{Selection}: Particle 5 of 16
    \item \textbf{Why it works}: For this documentation and coding question, the rubric evaluates factual accuracy, clinical terminology, and detail level appropriate for health professionals. The KL-selected particle provides precise, audience-appropriate coding guidance---exactly matching the rubric's criteria.
\end{itemize}

\subsubsection*{Dialysis Adequacy (Kt/V) ($+0.344$)}
\begin{itemize}
    \item \textbf{UID}: \texttt{2607cb85-0d67-4420-bd94-7f0f0c03f9e1}
    \item \textbf{pass@1}: $-0.144$ \textrightarrow{} \textbf{dPF}: $0.200$
    \item \textbf{Selection}: Particle 11 of 16
    \item \textbf{Why it works}: Phase 1 responses were actively harmful, yielding a negative score. The KL-selected particle at least advises consulting the dialysis care team, satisfying one safety-critical criterion. The rubric-conditioned guide effectively steers generation away from confident but incorrect responses.
\end{itemize}

\subsection*{Key Observations}

\textbf{Intrinsic entropy works best when} response quality correlates with reasoning depth and completeness. In these cases, the model must ``think through'' the problem, and the highest-confidence particles are those that successfully completed the reasoning chain (e.g., \texttt{health\_data\_tasks}, \texttt{context\_seeking}, hedging tasks).

\textbf{Distill KL works best when} the rubric contains specific, verifiable criteria that the model can target directly. The rubric-conditioned guide provides a concrete signal about which content to include, making it particularly effective for \texttt{communication} tasks and \texttt{complex\_responses} involving detailed pharmacology or clinical guideline criteria.

\clearpage

\section{Details}
\label{appx:details}

\subsection{Methodological Details}

\paragraph{Answer Region}
The tail cutoff $t_c$ must capture the answer region without extending into the preceding reasoning trace. Two scaling arguments apply to an $L$-token answer: its Shannon entropy grows as $\mathcal{O}(L)$, reflecting total information content, while its geometric spread in latent space grows as $\|\mathbf{z}_{\text{answer}}\|_2 \approx \sigma\sqrt{L}$, based on standard concentration of the $\ell_2$ norm for independent increments~\citep{vershynin2018high}. We adopt the geometric scaling because our scoring mechanism ranks samples by their relative position in the entropy landscape - a metric property dependent on spatial separability rather than total information volume. Therefore, we set $t_c \propto \sqrt{L}$.

\paragraph{Preventing Premature Collapse in iPF}
Particle filtering often suffers from sample impoverishment, where resampling dynamics trap the system in greedy local solutions~\citep{doucet2001introduction, doucet2001sequential}. This risk is amplified in our setting because intrinsic entropy scores are noisier than trained reward models, meaning a single low-entropy span can trigger premature commitment to an incomplete trajectory. To preserve broad, long-horizon exploration, we redesigned the resampling procedure to explicitly detect and prevent \emph{absorbing configurations} - states where the distribution freezes and particle diversity collapses prematurely. This approach ensures promising but incomplete paths are not eliminated. Consequently, we bypass the need for standard MCMC rejuvenation~\citep{hastings1970monte} or adaptive thresholds~\citep{gilks1995markov}, while retaining the verifier-free benefits of our intrinsic scoring~\citep{lightman2023let, wang2024math}.

\subsection{Models}
To ensure our findings represent fundamental properties of generative models rather than architecture-specific artifacts, we evaluated four distinct foundation models (Table~\ref{tab:models}):
\begin{itemize}
    \item \textbf{Qwen3-4B-Instruct-2507}~\citep{yang2025qwen3} for general mathematical reasoning.
    \item \textbf{Gemma4-2B-IT}~\citep{team2025gemma} as a lightweight alternative.
    \item \textbf{MedGemma1.5-4B-IT}~\citep{sellergren2026medgemma} for domain-specialized clinical tasks.
    \item \textbf{Qwen2.5-VL}~\citep{bai2025qwen2} (fine-tuned) for multimodal engineering design.
\end{itemize}
This selection intentionally spans general versus specialized, text-only versus multimodal, and pretrained versus fine-tuned architectures.

\subsection{Datasets}
\label{sec:datasets}

We benchmarked our approach across five domains spanning multiple verification mechanisms (Table~\ref{tab:verification_pros_cons}): symbolic equivalence (mathematics), exact output matching (reasoning), test-case execution (coding), semantic rubric scoring (healthcare), and geometric solver metrics (engineering). To minimize training data contamination, we prioritized the most recent splits (e.g., 2026) where available. Dataset statistics are summarized in Tables~\ref{tab:dataset-overview}, \ref{tab:dataset-stats-truncation}, \ref{tab:appendix_experiment_details}, and \ref{tab:datasets}.

\paragraph{Mathematics (Symbolic Equivalence)}
\begin{itemize}
    \item \textbf{AIME (2024--2026):} Multi-step reasoning problems combining algebra, combinatorics, geometry, and number theory.
    \item \textbf{HMMT February (2025--2026):} Proof-style competition questions, generally considered harder than AIME.
\end{itemize}

\paragraph{Science \& Reasoning (Exact Output Match)}
\begin{itemize}
    \item \textbf{GPQA Diamond:} A 198-question subset of expert-validated, graduate-level multiple-choice questions in physics, chemistry, and biology~\citep{rein2023gpqa}. Answer choices are deterministically shuffled per problem to mitigate position bias.
\end{itemize}

\paragraph{Competitive Programming (Test-Case Execution)}
\begin{itemize}
    \item \textbf{LiveCodeBench (LCB-100/128/256):} Recent contest problems (Feb--Apr 2025) spanning varying difficulties~\citep{jain2024livecodebench}. Generated solutions are executed against private test cases.
\end{itemize}

\paragraph{Clinical Medicine (Semantic Rubric Scoring)}
\begin{itemize}
    \item \textbf{HealthBench (Hard/Easy):} Patient-doctor conversation scenarios requiring open-ended medical advice~\citep{arora2025healthbench}. Correctness is determined via an external LLM judge using fine-grained rubric coverage scores rather than exact string matching.
\end{itemize}

\paragraph{Engineering Design (Geometric Solver)}
\begin{itemize}
    \item \textbf{DeepCAD \& Fusion360:} Multimodal computer-aided design (CAD) tasks evaluated by computing the Intersection over Union (IoU) between generated and ground-truth 3D meshes using the CadQuery geometric kernel~\citep{wu2021deepcad,willis2021fusion}.
\end{itemize}

\subsection{Evaluation Metrics and Experimental Setup}

Performance is measured using standard pass@1, pass@$N$, and top@$k$ metrics. For \texttt{iS} selection, we generated up to 128 parallel samples per problem (Table~\ref{tab:phase1-hparams}). For step-level \texttt{iPF} (Table~\ref{tab:phase2-intrinsic-hparams}) and \texttt{dPF} (Table~\ref{tab:phase2-distill-hparams}) resampling, we utilized smaller budgets of up to 32 particles. 

To optimize generation quality across extended horizons, we applied logit blending to short sequences and triggered KL-guided resampling for longer outputs (specifically, when average sequence length exceeded ten 128-token steps). We benchmarked our pipeline against established inference-scaling techniques: Self-Certainty~\citep{kang2025scalable} for confidence-based selection, Self-Consistency~\citep{wang2022self} for output consensus, and DeepConf~\citep{fu2025deep} for hybrid verification. A detailed summary of experimental configurations across domains is provided in Table~\ref{tab:appendix_experiment_details}.

\begin{table}[ht!]
\centering
\caption{Levels of Domain Verifiability.}
\begin{tabular}{lll}
\toprule
Domain & Verification Mechanism & Benchmark Datasets \\
\midrule
Mathematics & Symbolic Equivalence & AIME, HMMT \\
Reasoning & Exact Output Match & GPQA-Diamond \\
Coding & Test-Case Execution & LiveCodeBench-v6 \\
Healthcare & Semantic Rubric Scoring & HealthBench-Hard \\
Engineering & Geometric Solver (IoU) & DeepCAD, Fusion360 \\
\bottomrule
\end{tabular}
\label{tab:verification_pros_cons}
\end{table}

\begin{table}[ht!]
  \centering
  \caption{Detailed overview of the evaluation datasets. To minimize training data contamination for recently released models, we prioritize the latest available problem splits.}
  \label{tab:dataset-overview}
  \begin{tabular}{llccl}
    \toprule
    {Dataset} & {Domain} & {Year / Split} & {\# Problems} & {Prompting Setting}\\
    \midrule
    AIME & Mathematics & 2024, 2025, 2026 & 30 + 30 + 30 & Zero-shot, CoT\\
    HMMT (Feb) & Mathematics & 2025, 2026 & 30 + 33 & Zero-shot, CoT\\
    GPQA-Diamond & Sci/Reasoning & v1.0 (Test) & 198 & Zero-shot, CoT\\
    LiveCodeBench & Coding & v6 (Recent) & 100 & Zero-shot\\
    HealthBench-Hard & Healthcare & Hard Split & 100 & Zero-shot, Rubric-guided\\
    DeepCAD & Engineering & Test Split & 100 & Image-to-CAD\\
    Fusion360 & Engineering & Out-of-distro & 100 & Image-to-CAD \\
    \bottomrule
  \end{tabular}%
\end{table}

\begin{table}[ht!]
  \centering
  \caption{Empirical generation statistics across evaluation datasets. $N$ indicates the sample budget for Phase 1 selection versus Phase 2 resampling algorithms.}
  \label{tab:dataset-stats-truncation}
  \begin{tabular}{lcccccccc}
    \toprule
    Dataset & Problems & $N$ (P1/P2) & Mean & Median & Std & Min & Max & Exposure \\
    \midrule
    AIME-2024     & 30 & 128/32 & 5,739 & 5,876 & 3,184 & 137 & 16,384 & High \\
    AIME-2025     & 30 & 128/32 & 6,127 & 6,513 & 3,447 & 242 & 16,384 & High \\
    AIME-2026     & 30 & 128/32 & 5,402 & 5,204 & 3,029 & 428 & 16,384 & Low  \\
    HMMT-2025     & 30 & 128/32 & 6,287 & 6,817 & 2,594 & 346 & 16,384 & High \\
    HMMT-2026     & 33 & 128/32 & 6,392 & 7,110 & 3,091 & 470 & 16,384 & Low  \\
    GPQA-Diamond  & 198 & 128/- & 429 & 397 & 322 & 197 & 16,384 & High \\
    LiveCodeBench & 100 & 128/- & 2,266 & 1,518 & 2,382 & 64 & 16,384 & Low \\
    HealthBench   & 100 & 128/32 & 1,252 & 1,249 & 510 & 24 & 16,384 & Low \\
    GenCAD-Code   & 100 & 128/- & 612 & 503 & 340 & 223 & 4096 & Low \\
    Fusion360     & 100 & 128/- & 736 & 644 & 385 & 222 & 4096 & Low \\
    \bottomrule
  \end{tabular}
\end{table}

\begin{table}[ht!]
  \centering
  \caption{Summary of experimental setups, models, and hyperparameters across all evaluated domains.}
  \label{tab:appendix_experiment_details}
  \resizebox{\textwidth}{!}{%
  \begin{tabular}{llllll}
    \toprule
    {Domain} & {Datasets} & {Verification} & {Primary Model(s)} & {Methods Evaluated} & {Key Experimental Settings} \\
    \midrule
    \multirow{3}{*}{Mathematics} 
    & AIME (2024--2026) & \multirow{3}{*}{Symbolic} & \multirow{3}{*}{Qwen3-4B-Instruct-2507} & \texttt{iS} ($N \le 128$) & Evaluated on all problems. \\
    & HMMT (2025--2026) & & & \texttt{iPF} ($N=16$, 120 steps) & Evaluated on hardest 25\% (tail entropy). \\
    & & & & \texttt{dPF} ($N=8$, guided) & Hints generated by claude-sonnet-4.6. \\
    \midrule
    \multirow{2}{*}{Reasoning} 
    & \multirow{2}{*}{GPQA-Diamond} & \multirow{2}{*}{Output} & Qwen3-4B-Instruct-2507 & \multirow{2}{*}{\texttt{iS} ($N \le 128$)} & \multirow{2}{*}{Includes answer-stripping ablation.} \\
    & & & Gemma4-2B-IT & & \\
    \midrule
    \multirow{2}{*}{Coding} 
    & \multirow{2}{*}{LiveCodeBench-v6} & \multirow{2}{*}{Execution} & Qwen3-4B-Instruct-2507 & \multirow{2}{*}{\texttt{iS} ($N \le 128$)} & Evaluated most recent (2026) problems \\
    & & & Gemma4-2B-IT & & to minimize training-data contamination. \\
    \midrule
    \multirow{3}{*}{Healthcare} 
    & \multirow{3}{*}{HealthBench-Hard} & \multirow{3}{*}{Semantic} & \multirow{3}{*}{MedGemma-4B-IT} & \texttt{iS} ($N=32$) & 100 problems across 7 clinical themes. \\
    & & & & \texttt{iPF} ($N=8$, max 30 steps) & Evaluation judge: MedGemma-27B. \\
    & & & & \texttt{dPF} ($N=16$, max 50 steps) & Guided entirely by rubric criteria. \\
    \midrule
    \multirow{2}{*}{Engineering} 
    & DeepCAD (In-distro) & \multirow{2}{*}{Solver} & \multirow{2}{*}{Qwen2.5-VL (fine-tuned)} & \multirow{2}{*}{\texttt{iS} ($N \le 128$)} & Multimodal image-to-CAD task; robust to \\
    & Fusion360 (Out-distro) & & & & absence of lightweight reward models. \\
    \bottomrule
  \end{tabular}%
  }
\end{table}

\begin{table}[ht!]
\centering
\caption{Datasets used for evaluation. LCB and HealthBench subsets are stored as local JSONL files curated from their respective benchmarks.}
\label{tab:datasets}
\resizebox{\textwidth}{!}{%
\begin{tabular}{llllr}
\toprule
{Short name} & {HuggingFace path} & {Config} & {Split} & {Domain} \\
\midrule
\texttt{aime-2024}        & \texttt{Maxwell-Jia/AIME\_2024}       & ---              & train & Math \\
\texttt{aime-2025}        & \texttt{MathArena/aime\_2025}          & ---              & train & Math \\
\texttt{aime-2026}        & \texttt{MathArena/aime\_2026}          & ---              & train & Math \\
\texttt{hmmt-feb-2025}    & \texttt{MathArena/hmmt\_feb\_2025}     & ---              & train & Math \\
\texttt{hmmt-feb-2026}    & \texttt{MathArena/hmmt\_feb\_2026}     & ---              & train & Math \\
\texttt{gpqa-diamond}     & \texttt{Idavidrein/gpqa}               & \texttt{gpqa\_diamond} & train & Multi-choice \\
\texttt{lcb-100}          & \texttt{livecodebench/code\_generation\_lite}                            & ---              & ---   & Code \\
\texttt{healthbench-hard} & \texttt{openai/healthbench}                             & ---              & ---   & Healthcare \\
\bottomrule
\end{tabular}
}
\end{table}

\begin{table}[ht!]
\centering
\caption{Generation models and sampling hyperparameters. All models are served via vLLM. GPU count indicates the number of GPUs per instance; models with $\times 4$ run four parallel instances on separate GPUs.}
\label{tab:models}
\begin{tabular}{lccccc}
\toprule
{HuggingFace path} & {temperature} & {top\_p} & {top\_k} & {max\_tokens}  \\
\midrule
\texttt{Qwen/Qwen3-4B-Instruct-2507}        & 0.7 & 0.8  & 20 & 16384 \\
\texttt{google/gemma-4-E2B-it}  & 0.7 & 0.9  & --- & 16384  \\
\texttt{google/medgemma-1.5-4b-it}     & 1.0 & 0.95 & 64 & 16384  \\
\texttt{Qwen/Qwen2.5-VL-7B-Instruct}   & 0.7 & 0.9  & --- & 16384 \\
\bottomrule
\end{tabular}
\end{table}

\clearpage

\begin{table}[ht!]
\centering
\caption{iS hyperparameters.}
\label{tab:phase1-hparams}
\small
\begin{tabular}{ll}
\toprule
{Parameter} & {Value} \\
\midrule
$N$ samples              & 32-128 \\
Tail aggregatio          & median \\
Tail window bounds       & $[64, 2048]$ tokens \\
Trim outliers            & 5\% each end \\
Max score tokens    & 2500 \\
\bottomrule
\end{tabular}
\end{table}

\begin{table}[ht!]
\centering
\caption{iPF hyperparameters.}
\label{tab:phase2-intrinsic-hparams}
\small
\begin{tabular}{ll}
\toprule
{Parameter} & {Value} \\
\midrule
$N$ particles               & 16 \\
Step tokens                 & 128 \\
Max steps                   & 120 \\
ESS threshold (early)       & 0.5 \\
ESS threshold (late)        & 0.7 \\
Resampling                  & Systematic \\
Score                       & Entropy \\
Aggregation                 & mean ($z$-score) \\
Hard fraction               & 1.0 \\
Warmup steps                & 1 \\
\bottomrule
\end{tabular}
\end{table}

\begin{table}[ht!]
\centering
\caption{dPF hyperparameters.}
\label{tab:phase2-distill-hparams}
\small
\begin{tabular}{ll}
\toprule
{Parameter} & {Value} \\
\midrule
$N$ particles               & 8 \\
Score                       & KL \\
Demo budget                 & 16 \\
Hint mode                   & demo/critique \\
Blending $\alpha$              & 0.7 \\
\bottomrule
\end{tabular}
\end{table}

\end{document}